\documentclass[11pt]{article}

\usepackage{booktabs}

\usepackage{calc}
\usepackage{textcomp}
\usepackage[english]{babel}
\usepackage{amssymb,amsmath,array}
\usepackage[dvips]{graphicx}
\usepackage{subfigure}
\usepackage[latin1]{inputenc}
\usepackage{verbatim}
\usepackage{fancybox}
\usepackage[novbox]{pdfsync}
\usepackage{fancyvrb}
\usepackage{todonotes}
\usepackage{float}
\usepackage{pbox}
\usepackage{enumitem}
\usepackage{setspace}
\usepackage{lipsum}
\usepackage{threeparttable}
\usepackage{rotating}
\usepackage{color,hyperref} 
\definecolor{darkblue}{rgb}{0.0,0.0,0.3}
\hypersetup{colorlinks,breaklinks,
            linkcolor=darkblue,urlcolor=darkblue,
            anchorcolor=darkblue,citecolor=darkblue}

\usepackage{tikz}
\usetikzlibrary{arrows,shapes,automata,backgrounds,petri,fit,positioning,snakes}

\usepackage{booktabs}

\usepackage{titlesec}

\usepackage[margin=1in]{geometry}

\title{A Historical Analysis of the Field of OR/MS\\using Topic Models}
\author{Christopher J. Gatti,$^a$ James D. Brooks,$^b$\\and Sarah G. Nurre $^c$}
\date{}

\begin{document}
\maketitle

\vspace{-2em}
{\singlespace
{\centering
{\small
$^a$ Cirque du Soleil\\8400 2e Avenue\\Montr\'{e}al, QC H1Z 4M6, Canada\\{\tt gatti.cj@gmail.com}\\
\vspace{1em} $^b$ GE Global Research\\1 Research Circle\\Niskayuna, NY 12309, USA \\{\tt brooksjd1@gmail.com}\\
\vspace{1em} $^c$ Department of Industrial Engineering\\University of Arkansas\\4207 Bell Engineering Center\\Fayetteville, AR 72701, USA\\{\tt snurre@uark.edu}

}
}
}

\vfill
\singlespace
\noindent Corresponding author:\\
\hspace*{0.5cm} Christopher J. Gatti\\
\hspace*{0.5cm} Cirque du Soleil\\
\hspace*{0.5cm} 8400 2e Avenue\\
\hspace*{0.5cm} Montr\'{e}al, QC H1Z 4M6, Canada\\
\hspace*{0.5cm} {\tt gatti.cj@gmail.com}\\
\hspace*{0.5cm} 734-730-3190

\doublespace

\clearpage

\begin{abstract}
This study investigates the content of the published scientific literature in the fields of operations research and management science (OR/MS) since the early 1950s. Our study is based on 80,757 published journal abstracts from 37 of the leading OR/MS journals. We have developed a topic model, using Latent Dirichlet Allocation (LDA), and extend this analysis to reveal the temporal dynamics of the field, journals, and topics. Our analysis shows the generality or specificity of each of the journals, and we identify groups of journals with similar content, which are both consistent and inconsistent with intuition. We also show how journals have become more or less unique in their scope. A more detailed analysis of each journals' topics over time shows significant temporal dynamics, especially for journals with niche content. This study presents an observational, yet objective, view of the published literature from OR/MS that would be of interest to authors, editors, journals, and publishers. Furthermore, this work can be used by new entrants to the fields of OR/MS to understand the content landscape, as a starting point for discussions and inquiry of the field at large, or as a model for other fields to perform similar analyses.
\end{abstract}

\clearpage

\begin{table}[!htbp]
\centering
\footnotesize{
\begin{tabular}{ c c p{9.5cm}}
\hline
Symbol & Value & Description \\ \hline
$w_d$ &  & (LDA) Number of words in document $d$ \\
$\zeta$ &  & (LDA) Parameter defining Poisson distribution of $w^{(d)}$ \\
$\theta$ &  & (LDA) Topic mixture $\sim$Dirichlet($\alpha$) \\
$\alpha$ & 0.1 & (LDA) Dirichlet prior defining the topic distribution \\
$w_{di}$ &  & (LDA) Word $i$ from document $d$ \\
$z_{di}$ &  & (LDA) Topic assignment for word $i$ from document $d$ \\
$\beta$ & estimated & (LDA) Word probabilities, conditioned on the topic $z_n$ \\
$D$ & 80,757 & Number of articles/documents \\
$K$ & 40 & Number of topics in LDA model \\
$Y$ & 61 & Publishing time span in years (1952-2012) \\
$V$ & 24,504 & Unique words in LDA model vocabulary \\
$J$ &  & Set of all journals \\
$N$ & 37 & Number of journals \\
$D^{(j)}$ &  & Set of articles/documents from journal $j$ \\
$D^{(j)}_y$ &  & Set of articles/documents from journal $j$, year $y$ \\
$d^{(j)}$ & & Article/document $d$ from journal $j$ (implicitly, a document $d$ is associated with a year $y$) \\
$n^{(j)}$ &  & Number of articles/documents by journal $j$ \\
$\gamma_{d}$ &  & Topic distribution for article/document $d$ \\
$\gamma_{dk}$ &  & Topic proportion for article/document $d$ from topic $k$ \\
$\bar{\gamma}^{(j)}$ &  & Topic distribution for journal $j$ \\
$\bar{\gamma}^{(j)}_k$ &  & Topic proportion for journal $j$ for topic $k$ \\
$\bar{\gamma}^{(j)}_{y}$ &  & Topic distribution for journal $j$ and year $y$ \\
$\bar{\gamma}^{(j)}_{yk}$ &  & Topic proportion for journal $j$, year $y$, topic $k$ \\
$\hat{\gamma}_{y}$ &  & Topic distribution for year $y$ \\
$\hat{\gamma}_{yk}$ &  & Topic proportion for year $y$ and topic $k$ \\
$\rho^{(j)}_k$ &  & Range of the topic distribution for journal $j$, topic $k$ \\
$\phi^{(j)}_{min/max}$ &  & Minimum/maximum of $\rho^{(j)}_k$ \\
$\psi^{(j)}_{dec/inc}$ &  & Decrease/increase of topic proportion for journal $j$ \\
$\tau^{(j)}_{min/max}$ &  & Minimum/maximum topic dynamics for journal $j$ \\
$\delta^{(j)}_{yk}$ &  & Difference in topic proportions for journal $j$, topic $k$ and years $y$ and $y+1$\\ \hline
\end{tabular}
}
\begin{minipage}[b]{5in}
\caption[Variable description]{Description of variables and parameters.}
\label{tab:variables}
\end{minipage}
\end{table}

\clearpage

\section{Introduction}

Historical studies allow us to see where we've been, where we currently are, and they also allow us to speculate on where we might be headed in the future. Knowledge of interesting events of the past further provokes questions of their causes, which motivates additional investigation. An assessment of the present, by taking a global perspective, is useful to understand the current state of affairs and to reflect on where we lie in relationship to the past, allowing us to adapt to the environment or circumstances.

The published scientific literature is one avenue by which one can seek to understand the history and current state of an academic field. Such an analysis has the ability to lead to questions and insights regarding the motivations for scientific work, the development of ideas and scientific progress, the current problems and questions that are relevant to society, and the work that is important to funding agencies. While there is growing literature using quantitative methods that study publication citations (e.g., scientometrics, infometrics, and bibliometrics \cite{Borner2003, Chen2006, He2009, Mann2006, Neff2009, Wu2010}), these approaches cannot easily provide insights to these questions. Thus, to gain an understanding of a scientific field, we are interested in a content analytic approach---a method of analyzing unstructured textual information and extracting trends, meaning, and stories behind the data \cite{Berelson1952, Stemler2001, Krippendorff2013}.

Historical studies of entire scientific fields are not commonly performed because of the immense amount of time and effort required in obtaining, processing, and analyzing the data. Many field-wide studies are often based on citation analysis and derived measures. For example, in the analysis of subfields of mathematics \cite{Smolinsky2012}, library and information science dissertations \cite{Sugimoto2011}, and computational linguistics \cite{Hall2008}. Such studies are aimed at understanding citation-related phenomena, and thus are limited in what they can reveal from a contextual standpoint of the field. Conventional methods of textual summarization, such as content analysis \cite{Krippendorff2013}, are not suited for large, field-wide studies because of the amount of manual effort that would be required. Contemporary machine learning methods, such as topic models, have the ability to automatically summarize and extract meaning from textual data \cite{Blei2003}. Topic models find the underlying latent topics, or groups of related words, that are commonly expressed in similar documents. This approach is not often used in large-scale analyses, though it has had a place in more focused analyses, such as in studying the content of individual journals including the Proceedings of the National Academy \cite{Airoldi2010, Griffiths2004} and Science \cite{Blei2006, Blei2006a}, and for individual conferences (e.g., Neural Information Processing Systems \cite{Wang2006, Pruteanu-Malinici2010}, and SIGMOD \cite{Mei2007}). Others have taken a broader view of the scientific literature by clustering all scientific journals indexed in the Science Citation Index Journal Citation Reports (2006) according to their $h$-similarity \cite{Schubert2010}.

There has been limited similar work done for the fields of operations research (OR) and management science (MS). One of the earliest studies, though, took a content analytic approach of classifying articles from the journals Operations Research, Management Science, and Interfaces from 1962 and 1992 in which the authors found a publishing productivity imbalance between theory and application \cite{Reisman1994}. Another small comparative study of arguably the top two OR/MS journals, Operations Research and Management Science \cite{Eto2002}, used two volumes from each journal and compared author and citation patterns. Lozano and Salmer\'on used a Data Envelopment Analysis approach to explore the refereeing and publication process of OR/MS journals and also analyzed the relationship between article length and publication impact \cite{Lozano2005}. However, despite the limited number of works in this area, it is still believed that historical studies of the field of OR are essential to the progression and advancement of the field \cite{Kirby2000}.

The purpose of the present work is to understand the historical and current trends in the scientific content of the journals in the fields of operations research and management science. We are interested in determining the prevalence and the dynamics of topics in major journals and of these fields at large. That is, we seek to determine which topics have grown and faded over the years, and how the content within and between journals is similar or unique. The main contributions of this work are to: 1) present a comprehensive analysis of the content of the fields of OS/MS and of 37 journals, 2) provide authors, journals, editors, and publishers with objective information about the state of these fields and journals, and 3) to encourage discussion about the state of these fields and journals. Additionally, this work can help new entrants to the OR/MS fields understand the historical and current content of the published literature. This work also serves as a framework for performing an extensive and temporal content analysis of a body of literature, as this methodology may be of interest to other fields as well.

This work begins with our methodology in Section \ref{sec:methods}, which presents our underlying model, Latent Dirichlet Allocation, the data used in this work, as well as our analysis procedures. Section \ref{sec:results} presents an in-depth analysis of the topic model developed, including the distribution of topics for the field as a whole and for individual journals, in addition to numerous other metrics that capture topic characteristics and dynamics. Section \ref{sec:discussion} provides a practical context for all of our results, including how this work may be useful to different groups of individuals. This section also illuminates some of the interesting questions that our work generates, which are potential avenues for future work.

\section{Methodology}
\label{sec:methods}

In this section, we first provide an overview of Latent Dirichlet Allocation (LDA), the topic modeling approach used in this work. We present enough to give the reader a basic understanding of this approach, however, the interested reader is directed to the seminal work by Blei and colleagues \cite{Blei2003}. This section then describes the data used in this work as well as our preprocessing approach.

In the modeling of a set of documents, we use a particular vocabulary, as follows. We define a \emph{corpus} to be the set of all documents in our analysis, where $D$ is the total number of documents. A \emph{document} is a single abstract from any journal of interest from their respective publication history. Each document $d$ is comprised of a set of $w_d$ \emph{words}, and the unique words from all of the documents (over all journals) make up the \emph{dictionary} or \emph{vocabulary} (of size $V$) of the model. Furthermore, each document is associated with a journal and the year in which it was published, and thus there are additional levels of document groupings. A \emph{topic} is considered to set of terms that are similar and coherent and often appear in the same context.

\subsection{Latent Dirichlet Allocation}

\begin{figure}[!ht]
\centering
\includegraphics[width=0.95\linewidth]{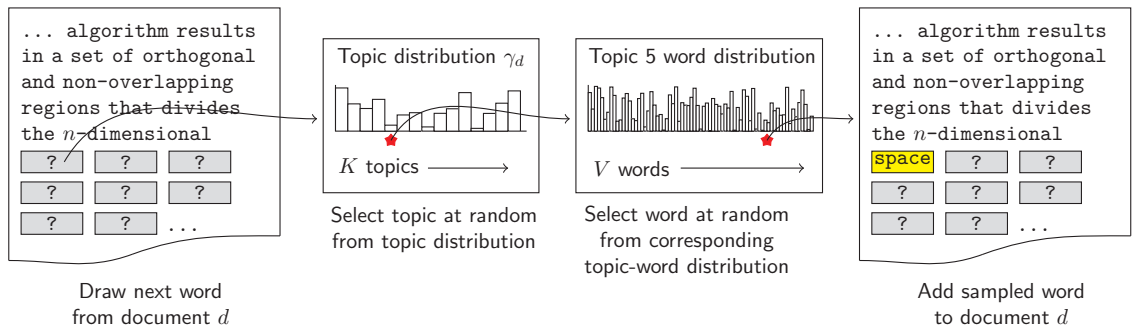}
\begin{minipage}[b]{5in}
\caption{Pictorial description of the generative process by which LDA assumes documents are generated. The document $d$ is assumed to have a topic distribution $\gamma_d$, that is specific to that document. Each word in the document is assumed to come from one of the $K$ latent topics, where it is selected at random from the $k^{\textup{th}}$ topic-word distribution. Note that the $K$ topic-word distributions are fixed across all documents.}
\label{fig:ldamodel}
\end{minipage}
\end{figure}

The LDA model is based on a probabilistic model of a corpus. Each document can be thought of as a composition, or a distribution, of latent topics, where each topic has an underlying distribution of the words in the dictionary. Figure \ref{fig:ldamodel} attempts to show the process by which documents are assumed to be formed. Each document  as a whole has an underlying topic distribution $\gamma_d$, and these topic distributions are unique to each individual document. Each word in the document is then assumed to be drawn from one of the $K$ topic distributions. The LDA algorithm assumes this process when determining the $D$ document-topic distributions ($\gamma$) and the $K$ topic-word distributions ($\beta$).

LDA is only one type of topic model, and other topic model variations or related textual modeling methods could have been used. Correlated topic models (CTM) \cite{Blei2007} allow for topics to be correlated, though this approach is significantly more computationally demanding than LDA. Dynamic topic models (DTM) \cite{Blei2006} exploit the temporal structure of a set of documents and create topics that essentially evolve over time, and this type of model seems like a natural approach for the analysis presented herein. However, as with CTM, DTM is very computationally demanding and was not feasible given the size of our data set, though the resulting model would surely be of interest if sufficient computational resources were available. Latent semantic analysis (LSA) or latent semantic indexing (LSI) \cite{Deerswester1990, Landauer1998} attempts to represent documents in a low-dimensional space that reflects the semantic relation among documents, however, the resulting model is often difficult to interpret. Each of these methods has pros and cons, and our approach of using LDA and a post-modeling aggregation procedure was the most ideal approach given our data set, computational feasibility, and analysis goals, and this approach (or similar approaches) have been used elsewhere in the literature \cite{Hall2008}.

\subsection{Modeling Methodology}

\begin{figure}[!ht]
\centering
\includegraphics[width=0.95\linewidth]{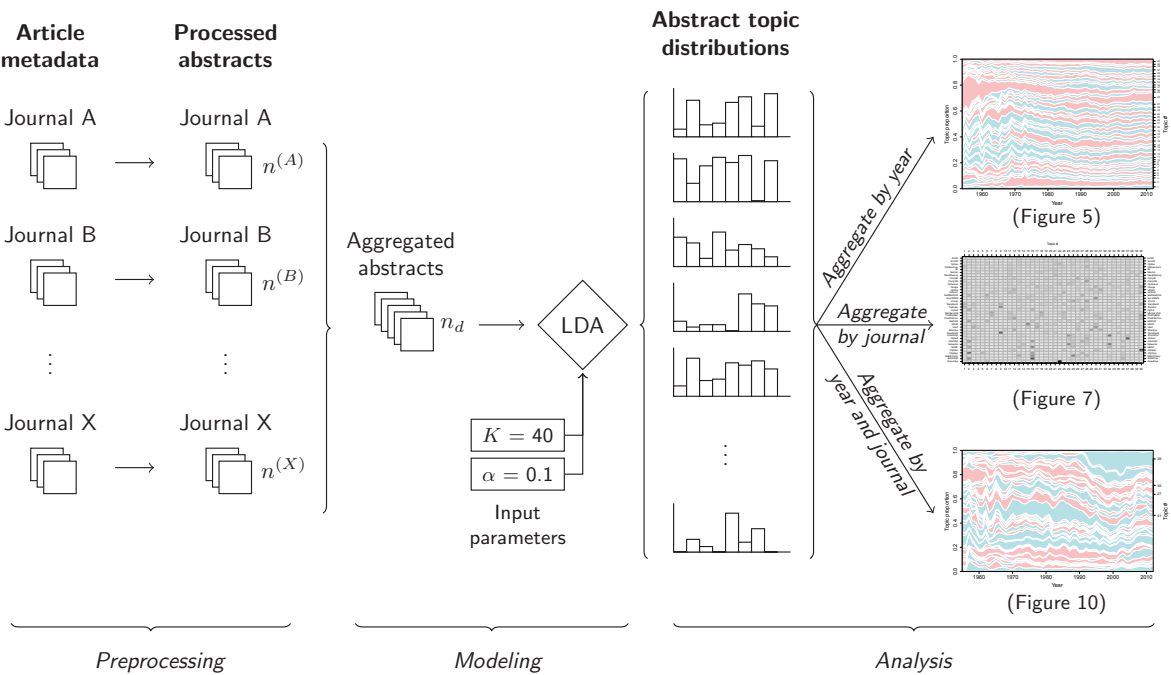}
\begin{minipage}[b]{5in}
\caption{Modeling methodology. Abstracts are preprocessed and input to the LDA model, from which document-topic distributions are obtained. These document-topic distributions are then aggregated in three different ways: by year, by journal, and by journal and year.}
\label{fig:methods}
\end{minipage}
\end{figure}

Figure \ref{fig:methods} provides an overview of the modeling methodology used in this work, where the details of each step will be described later in the text. Article metadata (including article abstracts) was procured from a select set of OR/MS journals. The abstracts were preprocessed to obtain abstracts that are abridged from their original form and are suitable to LDA modeling, and these abstracts were then aggregated to form the corpus that was input to a single LDA model. The primary output of the LDA model are the $D$ document-topic distributions (topic distributions for each individual document). These distributions are then aggregated in three different ways for analysis: aggregation by year (over all journals); aggregation by journal (over all years); and aggregation by year and journal. Each of these analysis are fully explained in the Results section.

\subsection{Data Selection}

The journals included in our analysis consist of the primary journals for the fields of operations research and management science. These 37 journals were selected from similar analyses \cite{Reisman1994, Stonebreaker2012} or from Journal Citation Reports \cite{jcr2012} and Scientific Journal Reports (SJR: {\tt www.scimagojr.com}). Table \ref{tab:journals} presents the journals that are included in our analysis, their abbreviations used in this work, the total number of articles for each journal, and the years from which articles were obtained for each journal. Note that articles from Handbooks in Operations Research and Management Science were recategorized as Surveys in Operations Research and Management Science because this journal changed its name during its publication history. The journal names listed are the most recent journal names, however, we obtained and included as much data as possible regardless of whether the name of the journal was changed during its publication history.

Only metadata from articles that had abstracts completely in English were included in this study. Note that due to our data selection and processing procedures, some of the journals included have either a break (discontinuity) in their publication history or have a truncated publication history from their actual publication history. Journals that have breaks in their publication history based on our data selection procedures are indicated by an asterisk (*) in Table \ref{tab:journals}.

\begin{table}[!htbp]
\centering
\footnotesize{
\begin{threeparttable}
\begin{tabular}{ p{9.5cm} l c c }
\hline
Journal & Abbreviation & Article & Years \\
& & count \\ \hline
Annals of Operations Research & AnnOR & 3397 & 1984--2012 \\
Computers \& Industrial Engineering & CompIE & 4501 & 1976--2012 \\
Computers \& Operations Research & CompOR & 3859 & 1974--2012 \\
European Journal of Operational Research & EJOR & 12021 & 1977--2012 \\
IIE Transactions & IIE & 2764 & 1969--2012 \\
INFORMS Journal on Computing & IJComp & 920 & 1989--2012 \\
Interfaces & Interf & 2335 & 1970--2012 \\
International Journal of Production Economics & IJPEcon & 452 & 1991--2012 \\
International Journal of Production Research & IJPRes & 7138 & 1961--2012$^*$ \\
Journal of Combinatorial Optimization & JCombOpt & 728 & 1997--2012 \\
Journal of Global Optimization & JGlobOpt & 1689 & 1991--2012 \\
Journal of Manufacturing Systems & JManSys & 385 & 1982--2006 \\
Journal of Operations Management & JOpMan & 971 & 1980--2012$^*$ \\
Journal of Optimization Theory and Applications & JOptApp & 5284 & 1967--2012 \\
Journal of Scheduling & JSched & 365 & 2003--2012 \\
Management Science & ManSci & 6453 & 1954--2012 \\
Manufacturing \& Service Operations Management & ManServOM & 399 & 1999--2012 \\
Mathematical Methods of Operations Research & MathMethOR & 522 & 1996--2012$^*$ \\
Mathematical Programming & MathP & 2448 & 1971--2012$^*$ \\
Mathematical Programming Computation & MathPComp & 43 & 2009--2012 \\
Mathematics of Operations Research & MathOR & 1749 & 1976--2012 \\
Military Operations Research & MilOR & 305 & 1994--2012$^*$ \\
Naval Research Logistics & NavalResLog & 2643 & 1954--2012$^*$ \\
Networks & Networks & 1606 & 1971--2012 \\
Omega & Omega & 2221 & 1973--2012 \\
Operations Research & OpRes & 4951 & 1954--2012 \\
Operations Research Letters & OpResLet & 2300 & 1981--2012 \\
Optimization Letters & OptLet & 454 & 2007--2012 \\
OR Spectrum & ORSpectrum & 380 & 1981--2012$^*$ \\
Production and Operations Management & ProdOpMan & 744 & 1992--2012 \\
Production Planning \& Control & ProdPlanCon & 1350 & 1990--2012 \\
Queueing Systems & QueueSys & 1237 & 1986--2012 \\
SIAM Journal on Optimization & SIAMJOpt & 1258 & 1991--2012 \\
Surveys in Operations Research and Management Science & SurvORMS & 228 & 1989--2012$^*$ \\
Transportation Research Part B & TransResB & 663 & 1979--2012 \\
Transportation Research Part E & TransResE & 701 & 1997--2012 \\
Transportation Science & TransSci & 1293 & 1967--2012 \\ \hline
Total & & 80,757 \\ \hline
\end{tabular}
\begin{tablenotes}
       \item{$^*$ Indicates that there was a break in publication between the given years.}
     \end{tablenotes}
  \end{threeparttable}
}
\begin{minipage}[b]{5in}
\caption[Journal stats]{Summary of the journals used herein, with the total number of articles and the years of publication for each journal. Note that some journal names have changed during their publication history, and the current journal names are shown in this table.}
\label{tab:journals}
\end{minipage}
\end{table}

Article metadata was obtained for all articles of these journals throughout their publication history, and this information included: title, authors, abstract, keywords, and bibliographic information. Permission was obtained from each journal or publisher prior to obtaining the article information, and the article metadata was obtained using automated web scraping scripts. The primary metadata of interest for this analysis was the article abstract, as this likely contains a high density of words appropriate for inferring the topics of a particular article \cite{Griffiths2004}. Indeed, if this were not so, the abstract would have failed at its purpose. Further, a similar modeling approach, correlated topic models (CTMs) \cite{Blei2006a}, was found to be relatively good at inferring other words (and thus a topic) in a document when fit with only a subset of the document \cite{Blei2007}. We argue that the abstract is the best possible subset of that size. 

Non-content articles were first removed from our data set because these generally do not have scientific content and are instead general communications. Examples of these non-content articles include those that present the authors (i.e., `About the Authors'), editorials, meeting announcements, contents list, etc. After removing these non-content articles, our data set consisted of $D = 80,757$ articles that span $Y = 61$ years (1952--2012). It should be noted that some articles that are listed as 2012 were not actually in print at this time and they were instead electronically published and were in press. It was not possible to discern which articles were in print versus electronically published given the vast number of articles, however, this caveat does not have a large affect on our analysis.

\begin{figure}[!ht]
\centering
\subfigure[Journal and article counts per year.]{\includegraphics[width=0.7\linewidth]{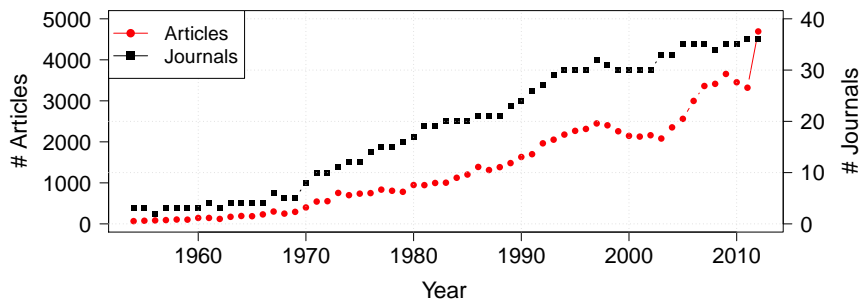} }
\subfigure[Average number of articles per journal per year.]{\includegraphics[width=0.7\linewidth]{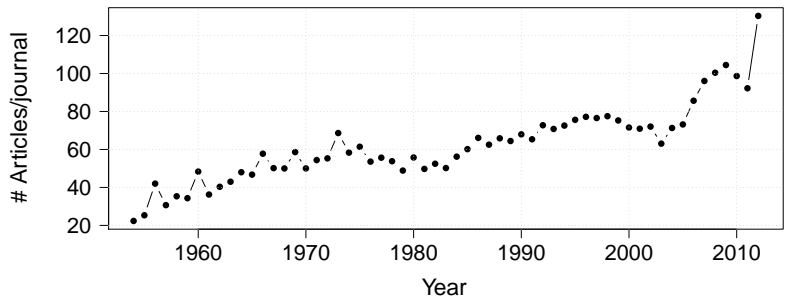} }
\begin{minipage}[b]{5in}
\caption{Growth of the number of articles and journals over time (a) and the average number of articles per journal over time (b). The spike in the number of articles in year 2012 is due to electronic pre-prints being included this year.}
\label{fig:journal_article_count}
\end{minipage}
\end{figure}

Figure \ref{fig:journal_article_count} shows the growth of the number of articles, the number of journals over time, and the average number of articles per journal from 1952 until 2012, all of which are generally increasing over time. The large spike in the number of articles in the latest years is due to electronic pre-prints being included in the articles from these years. Also note that the list of journals included in our analysis is not a comprehensive list of journals from the field and many more journals have started publishing in the fields of interest since 2000. However, the number of articles in the journals of interest is growing faster than the number of journals (Fig. \ref{fig:journal_article_count}b).

Textual preprocessing included removing all stopwords (as per the SMART information retrieval system \cite{Salton71}), removing all hyphens (thus treating hyphenated words as individual words), removing all whitespaces, removing terms with less than 3 occurrences in the entire data set, and removing documents with less than 10 words \cite{Blei2007}. These procedures resulted in a corpus vocabulary of $V = 24,504$ unique words.


\subsection{Model Specification}

The {\tt R} package {\tt topicmodels} \cite{Grun2011} was used to create an LDA topic model of the article abstract data set using the variational-expectation maximization (VEM) method. The number of topics $K$ was set to 40, $\alpha$ (topic distribution Dirichlet parameter) was set to 0.1, and all values of $\beta$ (term distribution of the topics) were estimated. We chose this number of topics largely based on subjective analysis of the content of each topic, but also based on the numbers of topics used in similar analyses \cite{Blei2007, Airoldi2010}. Default settings were used for the tolerances and number of iterations for VEM. The model was created on a Mac Mini server with a quad-core Intel i7 processor with 8GB of RAM, and took approximately 284 hours to run ($\sim 5$ days).

\subsection{Analysis}

This section presents a variety of metrics that are used to understand the topic distributions at a variety of levels, including: overall (static) topic distributions, journal similarity, journal uniqueness, journal entropy, temporal topic distributions, and a series of metrics that quantify the dynamics of temporal topic distributions. Overall topic distributions (aggregated over all years) are computed for each journal, where these distributions are static and cover all years of publication for each respective journal. For each journal, this distribution provides a general picture of the topic distribution over the course of the journal's publication history. Overall topic distributions, were summarized by creating a composite topic distribution from all articles from the same journal over all years. For each journal $j \in J$, where $J$ is the set of all 37 journals in our analysis, the composite topic distributions were computed by:
\begin{equation}
\bar{\gamma}^{(j)}_{k} = \frac{\sum \limits_{d \in j} \gamma_{dk}}{n^{(j)}}
\label{eqn:gamma_journaltopic}
\end{equation}

\noindent where $\gamma_{dk}$ is the $k^{th}$ topic proportion for document/article $d$, and $n^{(j)}$ is the total number of documents from journal $j$. Note that the bar symbol on $\bar{\gamma}^{(j)}_{k}$ indicates that $\gamma$ is a topic distribution for a single journal. In other words, for each journal, the document-topic distributions were summed across topics and then normalized by the total number of articles for that specific journal such that the resulting vector is a proper probability distribution. Note that this journal-topic distribution does not consider time as articles from all years are grouped together, though we also compute composite topic distributions by year for each journal in a similar way in an additional analysis.

Journals were then assessed by their similarity using hierarchical clustering \cite{Jain1988} and by their uniqueness compared to all other journals. Journal similarity is aimed at determining which journals have similar content, whereas journal uniqueness provides an indication of the broadness of a journal's scope. For hierarchical clustering, pairwise distances between the composite journal distributions were computed using the Euclidean distance metric, and clusters were agglomerated using the average-link criteria. Journal uniqueness was computed based on the Hellinger distance of the composite topic distributions between each journal \cite{Cha2007}. For two journals $P$ and $Q$, the Hellinger distance between the topic distributions is computed as:
\begin{equation}
H(P,Q) = \frac{1}{\sqrt{2}} \sqrt{ \sum \limits^{k}_{i=1} \left ( \sqrt{\bar{\gamma}^{(P)}_{i}} - \sqrt{\bar{\gamma}^{(Q)}_{i}} \right )^2 }
\label{eqn:hellinger}
\end{equation}

\noindent where a smaller Hellinger distance indicates that two journal topic distributions are more similar. A pairwise Hellinger distance was computed between all journal-topic distributions resulting in a symmetric $N \times N$ matrix $H$, where $N = 37$ is the number of journals. Journal uniqueness $u_j$ for journal $j$ was computed as:
\begin{equation}
u_j = \sum \limits^{N}_{i=1} H(i,j)
\label{eqn:uniqueness}
\end{equation}

\noindent where, similar to the Hellinger distance, a smaller uniqueness value indicates that a journal is more similar (or less unique) to all other journals. Journal entropy $e_j$ is an opposite measure to uniqueness, though not exactly, which is used to measure the uniformity of topic distributions:
\begin{equation}
e_j = - \sum \limits^K_{k=1} \bar{\gamma}^{(j)}_k \ ln \left ( \bar{\gamma}^{(j)}_k \right )
\end{equation}

The dynamics of the topics of the field at large were analyzed by computing topic distributions over time for all journals lumped together and for all articles that were published in each respective year. These field-wide distributions provide a sense of how the field has changed over the years and how topics have progressed or regressed with time. For each year $y \in Y$, the composite topic-year proportion was computed by:
\begin{equation}
\hat{\gamma}_{yk} = \frac{\sum \limits_{d \in y} \gamma_{dk}}{n_y}
\label{eqn:gamma_yeartopic}
\end{equation}

\noindent where $y$ is the set of articles published in a particular year, $\gamma_{dk}$ are their corresponding topic distributions, and $n_y$ is the number of articles published in year $y$. The hat symbol in $\hat{\gamma}_{yk}$ indicates that articles from all journals are aggregated for the topic $k$ and year $y$ of interest. In other words, the composite topic distribution for a particular year $y$ was computed by averaging the topic distributions of all articles published in that year.

The dynamics of each of the journals' topic distributions over time could also be used to understand the trajectories of the topics each of the individual journals published. We first define the individual topic proportions over time for each journal, which is similar to Equation (\ref{eqn:gamma_yeartopic}), but that is specific to each journal $j$. The proportion of topic $k$ for journal $j$ in year $y$ is computed by:
\begin{equation}
\bar{\gamma}^{(j)}_{yk} = \frac{\sum \limits_{d \in j \ \cup \ d \in y} \gamma^{(j)}_{dk}}{n^{(j)}_y}
\end{equation}

\noindent where $n^{(j)}_y$ is the number of articles for journal $j$ in year $y$. Again, the bar symbol in $\bar{\gamma}^{(j)}_{yk}$ indicates that articles are aggregated for a particular journal $j$, and for the topic $k$ and year $y$ of interest.

Additionally, for all journals, we quantify some of the dynamics of the topic proportions over time using the following eight metrics. Furthermore, rather than reporting the quantitative dynamics of the topic proportions, we report the topic numbers that have the respective dynamics.

\begin{itemize}
\item[]{\textbf{Minimum/maximum topic ranges}: The topics with the minimum and maximum ranges are those that have the smallest and largest, respectively, topic proportion ranges (i.e., maximum proportion minus minimum proportion, per topic) over time. These measures indicate the topics have been either insignificant or prominent, respectively, in a journal's publication history. The minimum and maximum topic proportion ranges are defined by:
\begin{align}
\rho^{(j)}_k & = \underset{y}{\textup{range}} \left ( \bar{\gamma}^{(j)}_{yk} \right ) \nonumber \\
\phi^{(j)}_{\min} & = \min_{k} \left ( \rho^{(j)}_k \right ) \nonumber \\
\phi^{(j)}_{\max} & = \max_{k} \left ( \rho^{(j)}_k \right ) \nonumber
\end{align}

\noindent which computes the range of each topic and is assessed over all years $y$.}

\item[]{\textbf{Decreasing/increasing topic changes}: The topics with the largest decreasing and the largest increasing changes are those that have changed the least and most, respectively, from the first publication time point of each journal until the latest publication time point. These measures are used to identify the topics that have either receded or grown the most over a journal's publication history, which may potentially indicate a change in the direction of a journal's scope. The smallest and largest topic proportion changes are defined by:
\begin{align}
\psi^{(j)}_{\textup{dec}} & =  \min_{k} \left ( \bar{\gamma}^{(j)}_{Y k} - \bar{\gamma}^{(j)}_{1 k} \right ) \quad \forall \quad k \ \ s.t. \ \ \bar{\gamma}^{(j)}_{Y k} - \bar{\gamma}^{(j)}_{1 k} < 0 \nonumber \\
\psi^{(j)}_{\textup{inc}} & = \max_{k} \left ( \bar{\gamma}^{(j)}_{Y k} - \bar{\gamma}^{(j)}_{1 k} \right ) \quad \forall \quad k \ \ s.t. \ \ \bar{\gamma}^{(j)}_{Y k} - \bar{\gamma}^{(j)}_{1 k} > 0 \nonumber
\end{align}

\noindent where $\bar{\gamma}^{(j)}_{1 k}$ and $\bar{\gamma}^{(j)}_{Y k}$ are the first and latest, respectively, proportions of topic $k$ for journal $j$.}

\item[]{\textbf{Least/most dynamic topics}: The topics that are the least (minimum) and the most (maximum) dynamic over time are those for which the standard deviation of the difference in proportions of the adjacent time points are least and greatest, respectively. This metric attempts to identify topics which are the most consistently and erratically, respectively, represented throughout the course of a journal's publication history. These metrics are defined by:
\begin{align}
\delta^{(j)}_{yk} & = \underset{y}{\textup{diff}} \ \bar{\gamma}^{(j)}_{yk} \nonumber \\
\tau^{(j)}_{\min} & = \min_{k} \left ( \underset{y}{\textup{sd}} \left ( \delta^{(j)}_{yk} \right ) \right ) \nonumber \\
\tau^{(j)}_{\max} & = \max_{k} \left ( \underset{y}{\textup{sd}} \left ( \delta^{(j)}_{yk} \right ) \right ) \nonumber
\end{align}

\noindent where $\delta^{(j)}_{yk}$ is the difference in topic proportions for topic $k$ between adjacent years, and $\underset{y}{\textup{sd}} \left ( \delta^{(j)}_{yk} \right )$ is the standard deviation of the differences across years $y$ for the same topic $k$.}

\item[]{\textbf{Greatest topic proportions}: We also report the topics (and the years in parentheses) which had the greatest overall topic proportion at any time during the publication history ($p_{\max}$) and at the most recent year (mry) ($p^{\textup{mry}}_{\max}$). These measures and topics indicate the most prominent topic over each journal's publication history and the most prominent topic at the latest available time point.}
\end{itemize}

\section{Results}
\label{sec:results}

We use wordclouds to present the 40 topics identified by the topic model. This type of \emph{raw} topic presentation is used to allow the reader to interpret the content of the topics themselves. The most common approach to describing the latent topics, however, is through displaying a list of a fixed number of top words (i.e., those with highest probability) for each topic (e.g., 5-10) \cite{Taddy2012}. In other cases, authors simply provide names by inspection \cite{Hall2008}, and only one other method of automatically naming topics is known \cite{Mei2007}. There is no current classification scheme for the OR/MS fields, however, this work could provide an initial classification scheme for this literature, as latent topic models have been found to correlate reasonably well to existing classification schemes \cite{Airoldi2010}.

\begin{figure}[!h]
\centering
\subfigure[Topic \#7]{\includegraphics[trim=1.2cm 1.2cm 1.2cm 1.2cm, clip=true, width=0.4\textwidth]{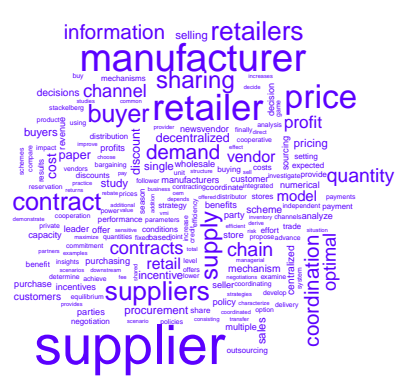} }
\subfigure[Topic \#17]{\includegraphics[trim=1.2cm 1.2cm 1.2cm 1.2cm, clip=true, width=0.4\textwidth]{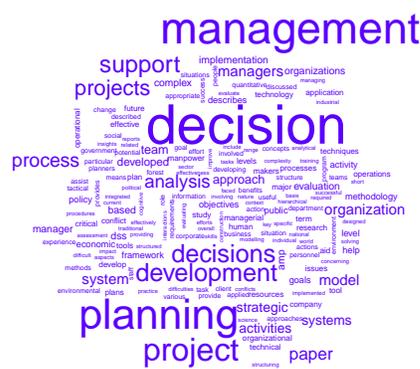} }
\subfigure[Topic \#28]{\includegraphics[trim=1.2cm 1.2cm 1.2cm 1.2cm, clip=true, width=0.4\textwidth]{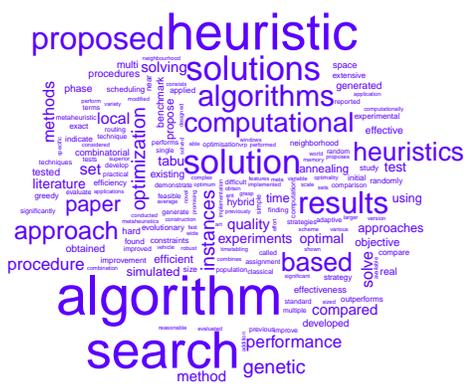} }
\subfigure[Topic \#39]{\includegraphics[trim=1.2cm 1.2cm 1.2cm 1.2cm, clip=true, width=0.4\textwidth]{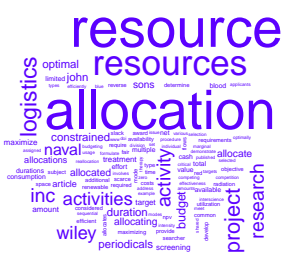} }
\caption{Wordclouds for topics 7, 17, 28, and 39.}
\label{fig:wordclouds0}
\end{figure}

Figure \ref{fig:wordclouds0} shows examples of the topic wordclouds for topics 7, 17, 28, and 39, and all wordclouds are provided in Appendix A. Each figure shows the prominent terms for a single topic, where the size of the terms are scaled based on the resulting term probabilities from the topic model. In these example topics that are presented in Figure \ref{fig:wordclouds0}, there seems to be clear and cohesive concepts that are consistent with themes in OR/MS and provide initial validity for the model.

\subsection{Topic dynamics over time}

\begin{figure}[!ht]
\centering
\includegraphics[width=\linewidth]{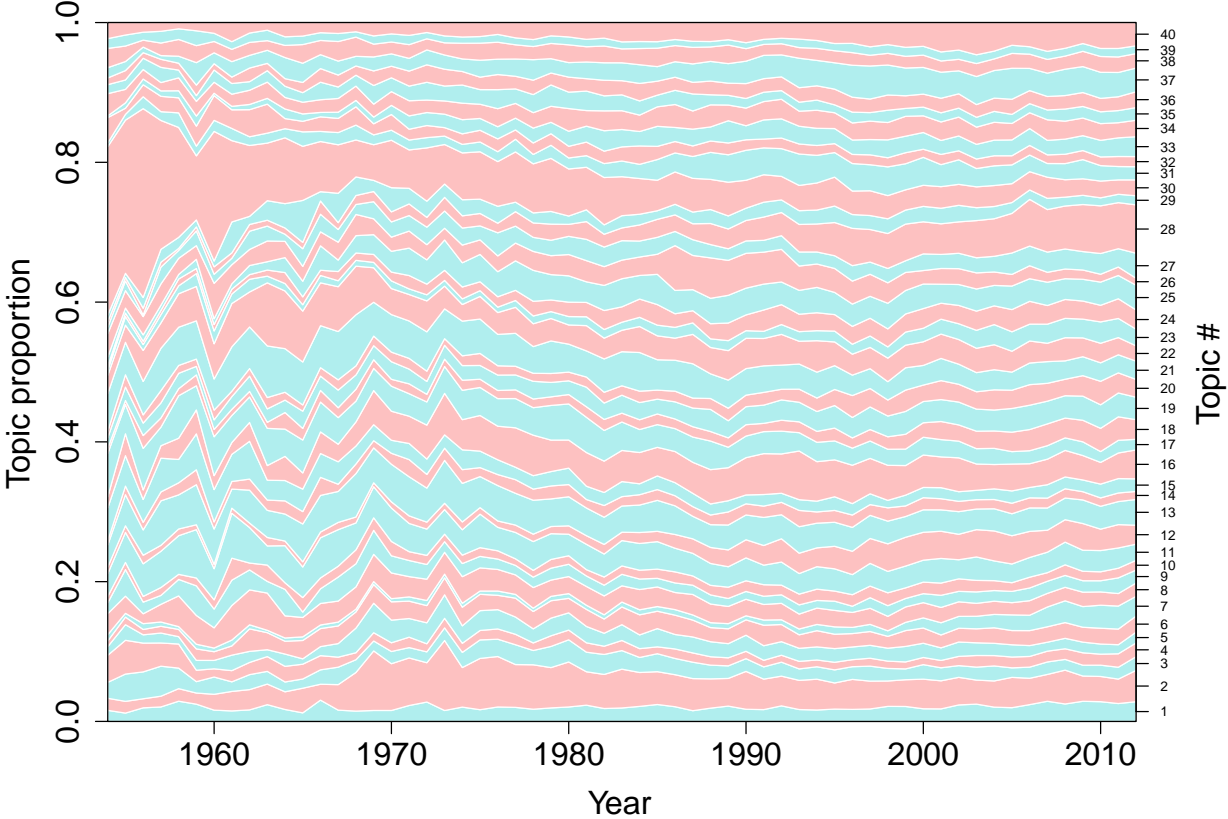}
\begin{minipage}[b]{5in}
\caption{Topic distributions over time for the field at large (i.e., all journals together). Topic numbers are shown on the right side of this figure.}
\label{fig:topicsovertime}
\end{minipage}
\end{figure}

Figure \ref{fig:topicsovertime} shows the temporal dynamics of the topic distributions for all of the journals considered herein during the entire publication history for the data we have obtained. This figure shows how relative topic popularity in the field has increased or decreased over time. It should be noted that the dynamics of the topics are somewhat influenced by the total number of articles published in each year, and this is most evident in the early time points where there are large changes in topic proportions. Nonetheless, we do see that at the early time points, two topics (topics 4 and 30) were clearly the most prevalent, and these topics seems to be related to auctions (topic 4) and very general terms of the field (topic 30). At the latest time point, topics 28 and 2 are the most prevalent, which are related to algorithms (topic 28) and proofs (topic 2), although there is not a single topic that stands out and dominates the field. Instead, the field is nearly evenly represented by all topics at the most recent time point.

\begin{figure}[!ht]
\centering
\includegraphics[width=0.5\linewidth]{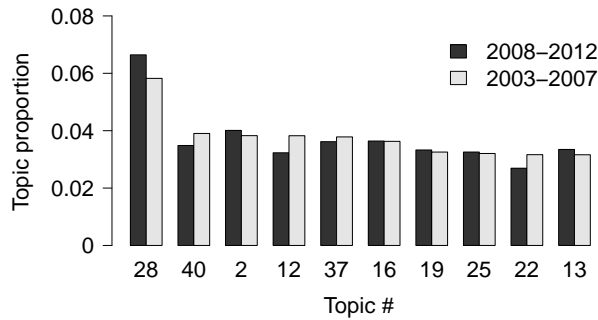}
\begin{minipage}[b]{5in}
\caption{Top 10 topic proportions of all journals for the years 2008--2012 compared to those of the years 2003--2007. Topics are sorted by their topic proportion in the years 2008--2012.}
\label{fig:topicsprop}
\end{minipage}
\end{figure}

The top 10 topics, based on their average proportion during the time period 2008--2012, are shown in Figure \ref{fig:topicsprop}. Additionally, the topic proportions for these topics are compared to those during 2003--2007 in order to highlight the most recent publication years, as well as to determine how these topics have increased or decreased in popularity. We find that topic 28 has remained the most prominent topic in the past 10 years by a fair margin and its topic proportion has increased from 2003--2007 to 2008--2012 as well. However, we see relatively minor changes in the other topic proportions for these topics.

\subsection{Composite topic distributions}

\begin{sidewaysfigure}
\centering
\subfigure{\includegraphics[width=\linewidth]{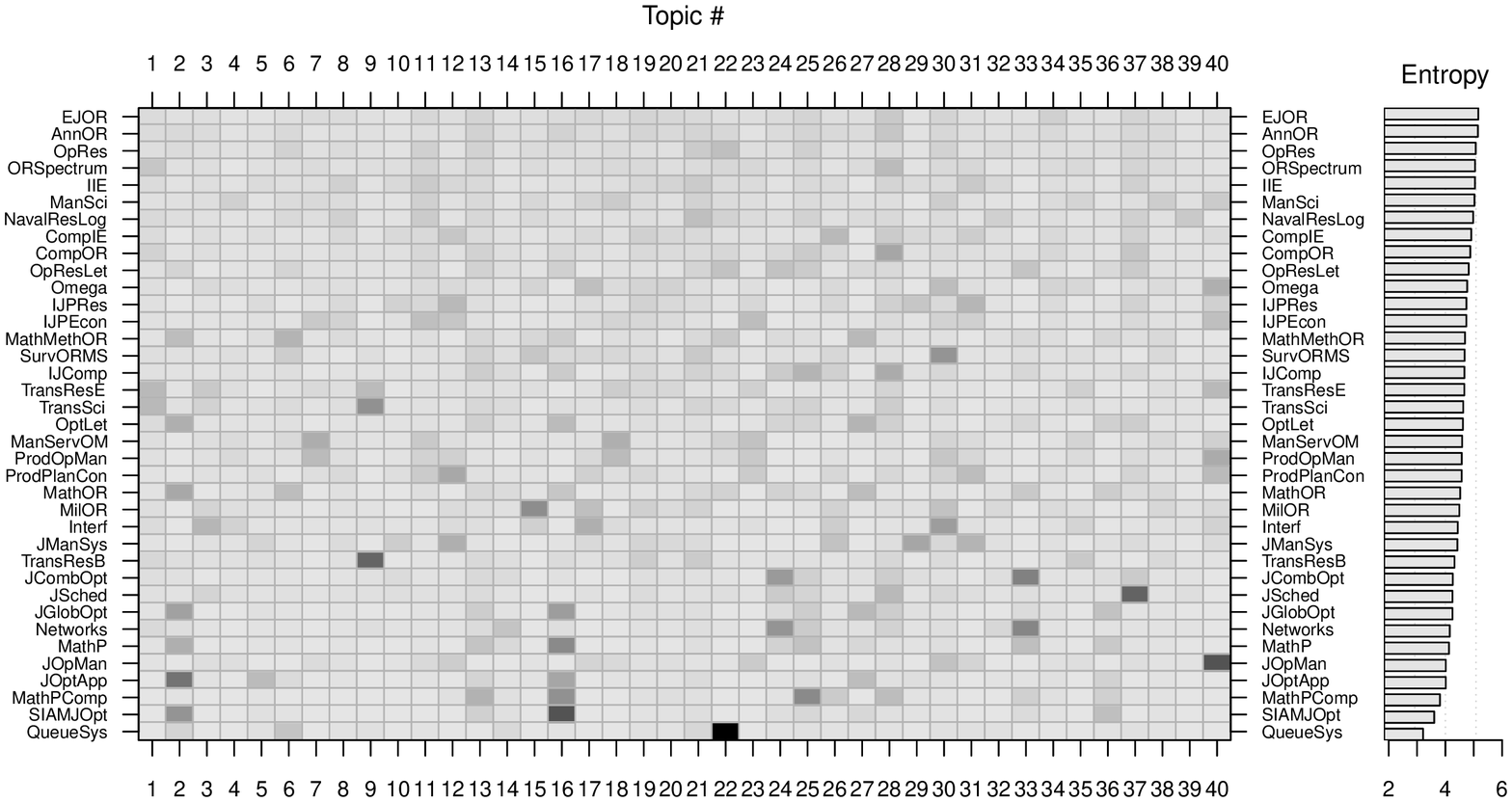}}
\subfigure{\includegraphics[width=0.4\linewidth]{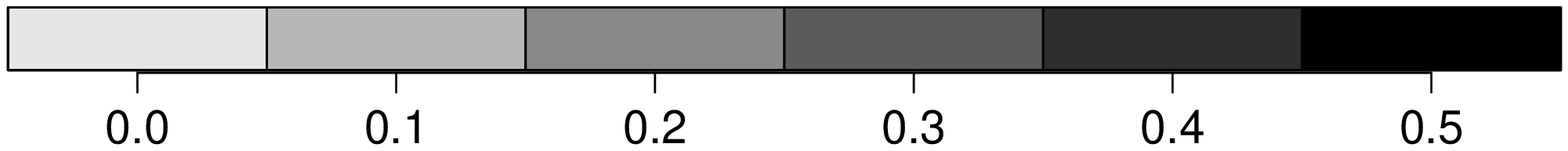}} \\
\begin{minipage}[b]{5in}
\caption{Topic distributions for all journals. Topic numbers correspond to the topics show in Appendix A.}
\label{fig:journaltopicdist}
\end{minipage}
\end{sidewaysfigure}

Figure \ref{fig:journaltopicdist} shows the composite topic distributions for each journal as a heatmap where each row represents a single journal and the darkness of the colored boxes is scaled to the proportion of the topic (indicated by the color bar) as represented in each respective journal. Additionally, the barplot on the right shows the entropy for each journal's topic distribution, and the journals for the heatmap and the entropy barplot are sorted according to their entropy. This figure is particularly useful for identifying which journals have relatively narrow content and which have broad content, such that journals with narrow scopes have a single or a few dominant topic proportions and a lower entropy, whereas journals with broad and general scopes have consistently small topic proportions across most topics and have a larger entropy. For example, {\it Queueing Systems} has a narrow scope that is represented largely by topic 22, whereas {\it Annals of Operations Research} publishes articles with a large variety of topics.

\subsection{Journal similarity}

\begin{figure}[t]
\centering
\includegraphics[width=0.85\linewidth]{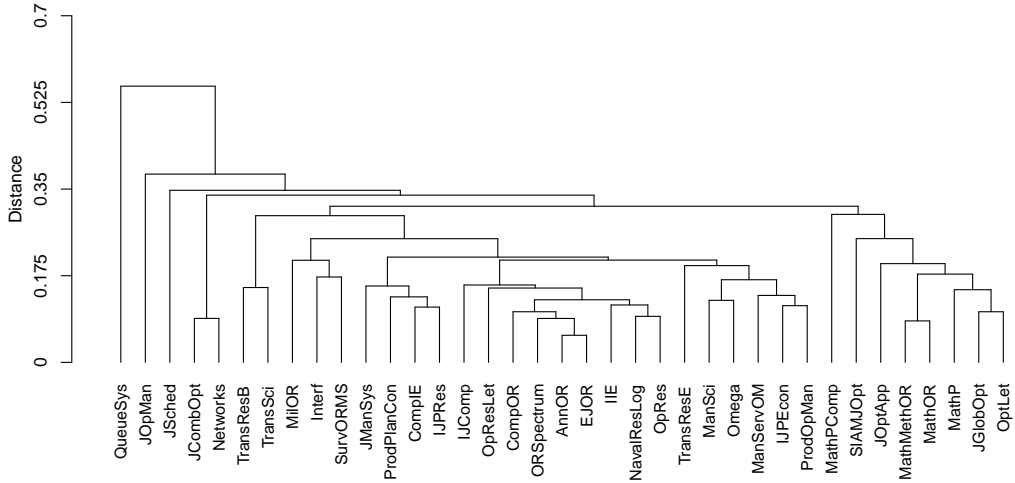}
\begin{minipage}[b]{5in}
\caption{Hierarchical clustering of journals based on their composite topic distributions over all years. Cluster distances for hierarchical clustering are indicated by the vertical axis of this figure.}
\label{fig:clustering}
\end{minipage}
\end{figure}

\begin{figure}[H]
\centering
\includegraphics[width=0.75\linewidth]{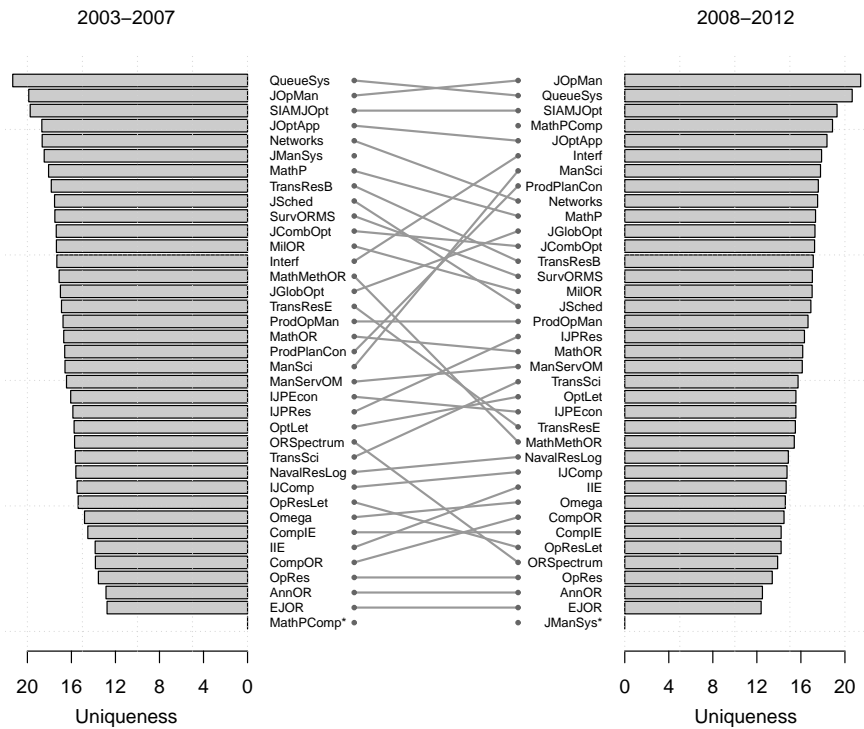}
\begin{minipage}[b]{5in}
\caption{Journal uniqueness of journals based on their composite topic distributions for the years 2003--2007 and for 2008--2012. Journals are sorted by uniqueness for each time periods, and are connected between time periods to easily see changes in uniqueness values. Note that {\it Mathematical Programming Computation} (MathPComp) and {\it Journal of Manufacturing Systems} (JManSys) had no uniqueness value for 2003--2007 and 2008--2012, respectively, and that these cases are noted with an asterisk.}
\label{fig:uniqueness}
\end{minipage}
\end{figure}

Figures \ref{fig:clustering} and \ref{fig:uniqueness} show hierarchical clustering and uniqueness, respectively, for all journals. Journal similarity and journal uniqueness, as presented here, say different things about the relationship of journals to each other and to the group of journals as a whole, though they present somewhat (but not exactly) complementary characteristics. At the lowest/finest level of agglomeration (smaller cluster distances), hierarchical clustering shows the relative similarity between pairs or small groups of journals, whereas at the highest level of agglomeration (larger cluster distances), hierarchical clustering shows how journals relate to the group of journals as a whole. Journal uniqueness, on the other hand, shows how each journal's content differs from all of the other journals, and thus there is no pairwise comparison.

Hierarchical clustering shows that many journal pairs and groups have relatively low cluster distances (as indicated by the vertical axis) that are intuitively consistent such that they should have similar content, which support the validity of our model, for example: {\it Math Methods of Operations Research} and {\it Mathematics of Operations Research}, {\it Transportation Research Part B} and {\it Transportation Science}, {\it Annals of Operations Research} and {\it European Journal of Operations Research}, and {\it Management Science} and {\it Omega}. This figure also shows which journals or groups of journals have relatively unique content, as their distance from other journals increases. However, we find that there are pairs or groups of journals that have surprisingly similar content, including: the {\it Journal of Combinatorial Optimization} and {\it Networks,} {\it Naval Research Logistics} and {\it Operations Research}, and {\it Military Operations Research} with respect to {\it Interfaces} and {\it Surveys in Operations Research} and {\it Management Science}. Additionally, we find other notable artefacts as well, such that {\it Transportation Research Part E} is quite distant from the other transportation-related journals, and that the {\it Journal of Operations Management} and the {\it Journal of Combinatorial Optimization} are not grouped with the other optimization journals.

Journal uniqueness (Figure \ref{fig:uniqueness}) is shown for each journal for 2003--2007 and for 2008--2012, where the journals are sorted by their uniqueness values for each of the time periods, and where each journal is `connected' between the two time periods to show the relative change in value and position. Journals that have small uniqueness values (e.g., {\it Annals of Operations Research} and {\it European Journal of Operational Research}), are generally regarded to have a relatively wide scope, whereas journals with large uniqueness values (e.g., {\it Queueing Systems} and {\it SIAM Journal of Optimization}) generally have a narrower scope, and we see that these scope-generalizations appear in the uniqueness values. The consistencies of these figures with our intuition is encouraging and adds credence to the topic model.

This figure also highlights the changes in each journal's uniqueness during the selected time periods. The most unique journals, {\it Queueing Systems}, the {\it Journal of Operations Management}, and {\it SIAM Journal of Optimization} all remain very unique over the last 10 years. However, we do see significant changes in uniqueness. Some of the journals that have become more unique in their scope include {\it Interfaces}, {\it Management Science}, {\it Production Planning \& Control}, and those that have become less unique include {\it Mathematical Methods of Operations Research}, {\it Transportation Research Part E}, and {\it OR Spectrum}, amongst others.

\subsection{Journal topic dynamics over time}

\begin{figure}[!ht]
\centering
\subfigure[Computers \& Industrial Engineering.]{\includegraphics[width=0.48\textwidth]{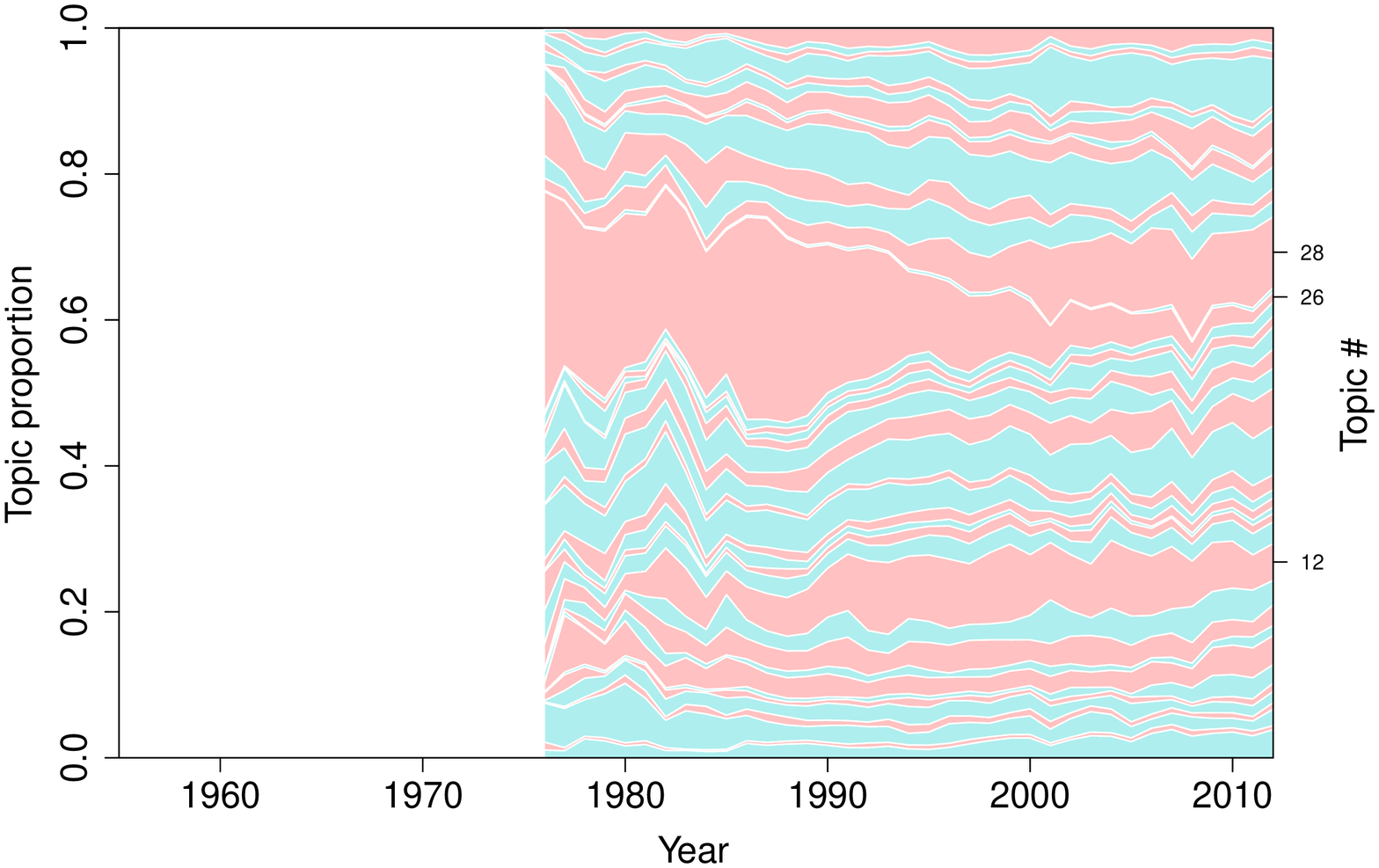} }
\subfigure[Computers \& Operations Research.]{\includegraphics[width=0.48\textwidth]{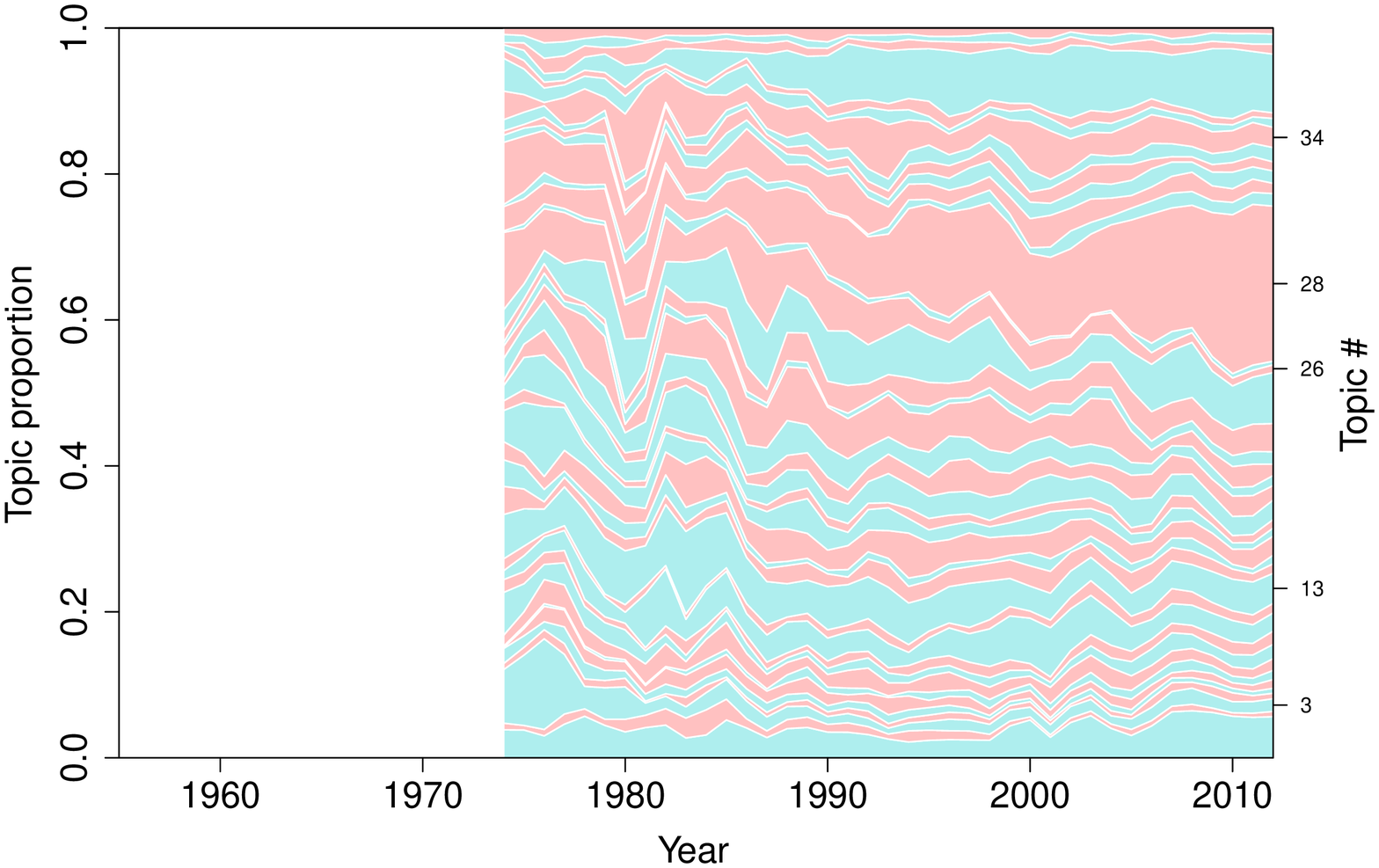} }
\subfigure[Interfaces.]{\includegraphics[width=0.48\textwidth]{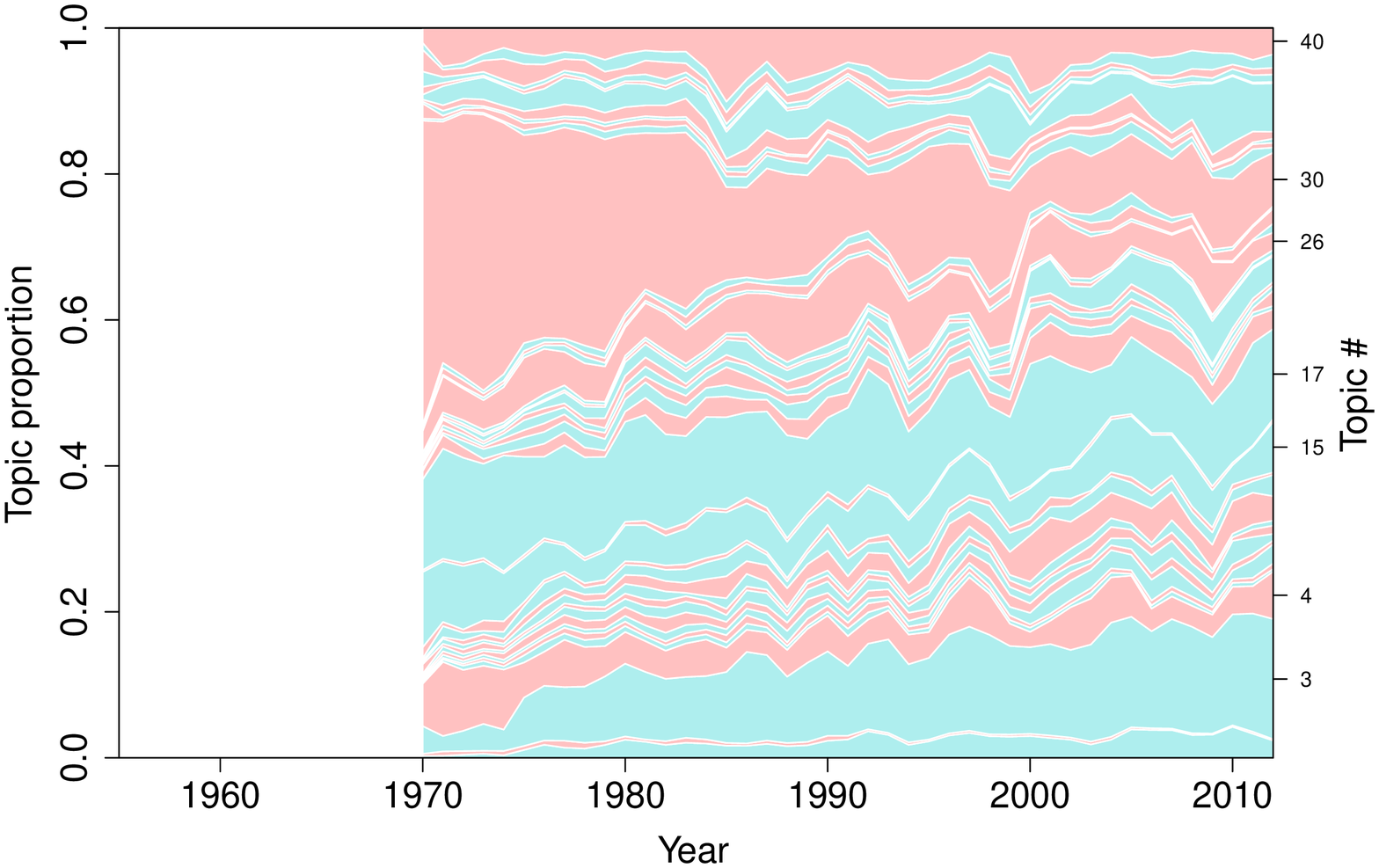} }
\subfigure[International Journal of Production Research.]{\includegraphics[width=0.48\textwidth]{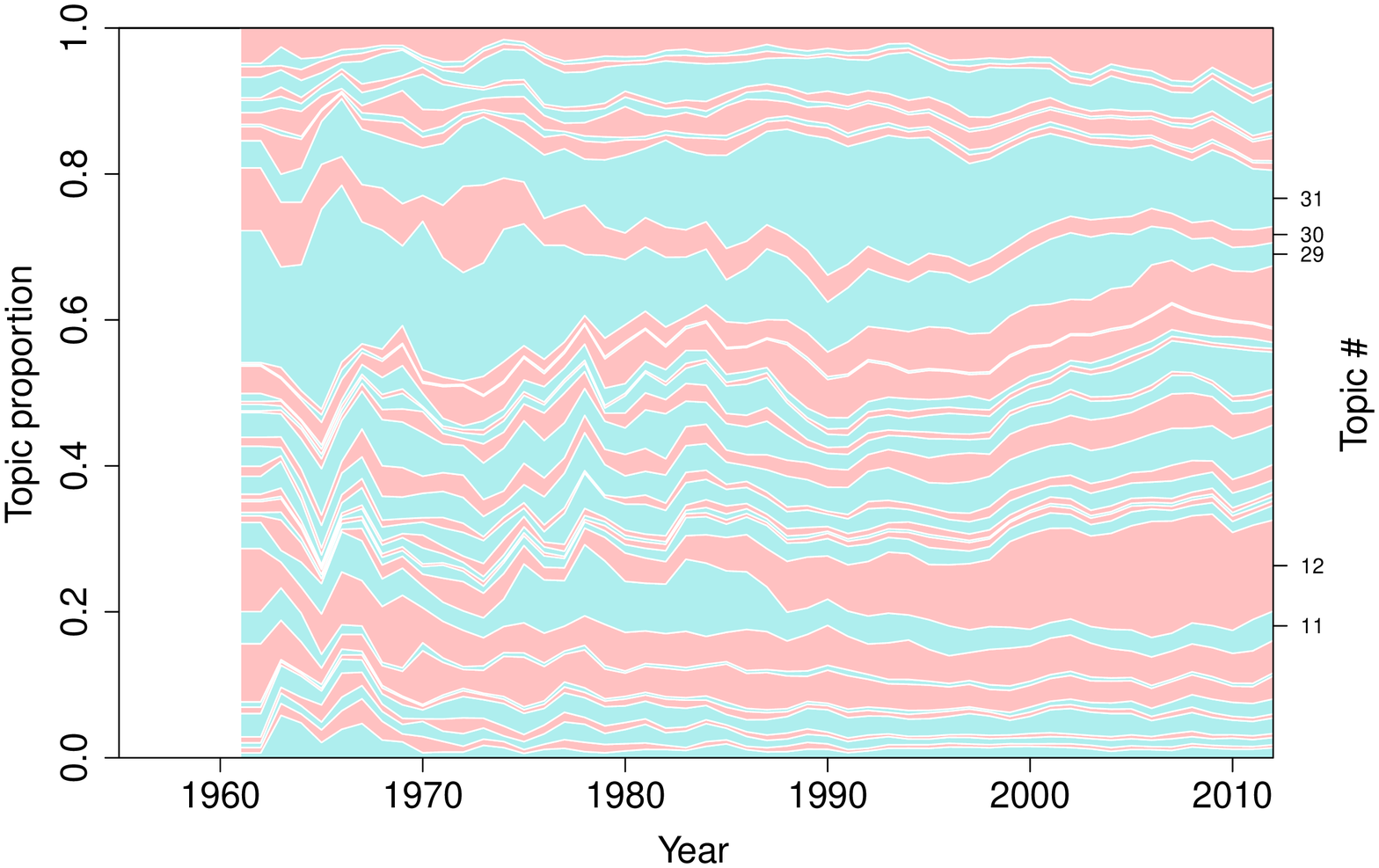} }
\subfigure[Journal of Global Optimization.]{\includegraphics[width=0.48\textwidth]{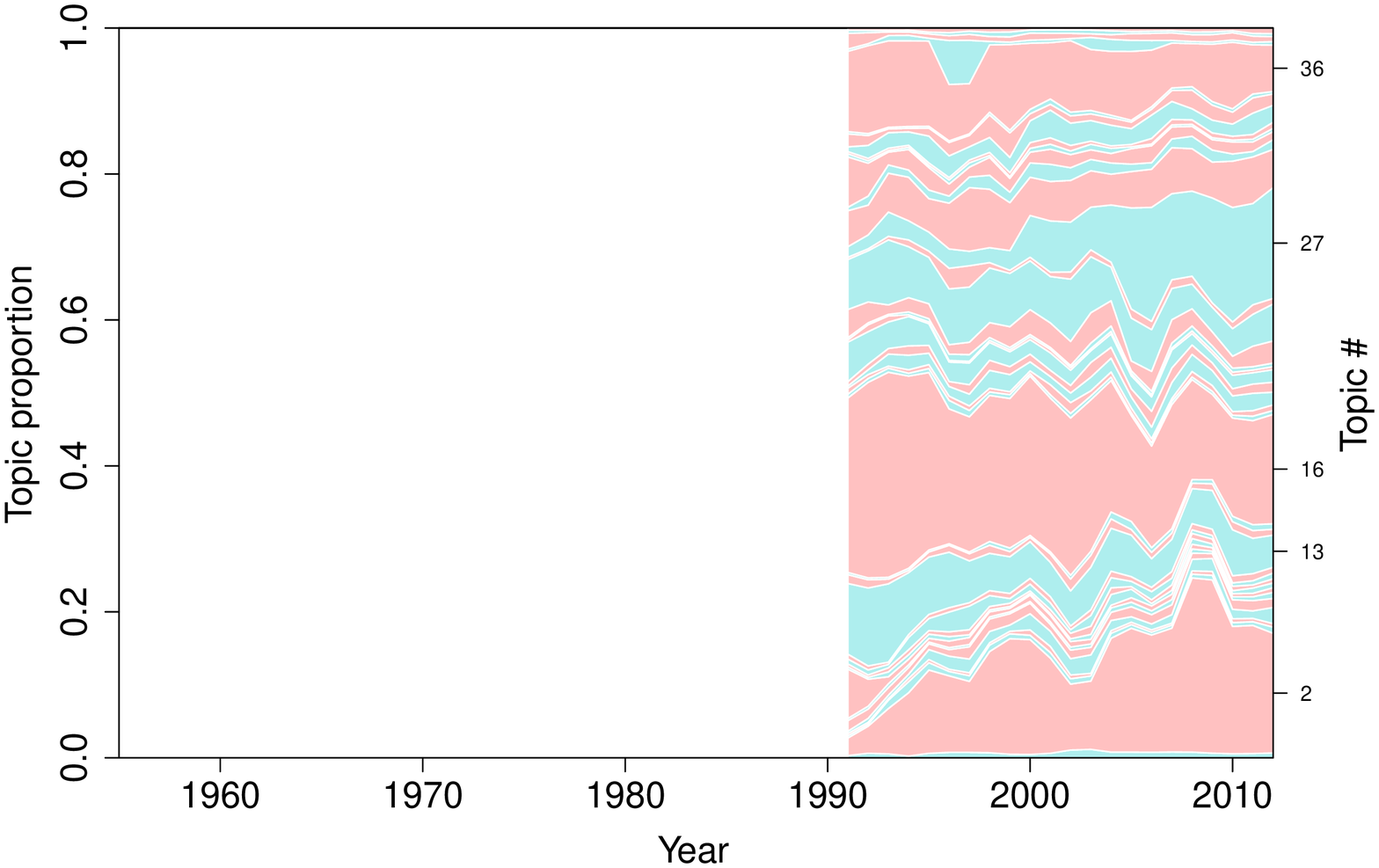} }
\subfigure[Journal of Manufacturing Systems.]{\includegraphics[width=0.48\textwidth]{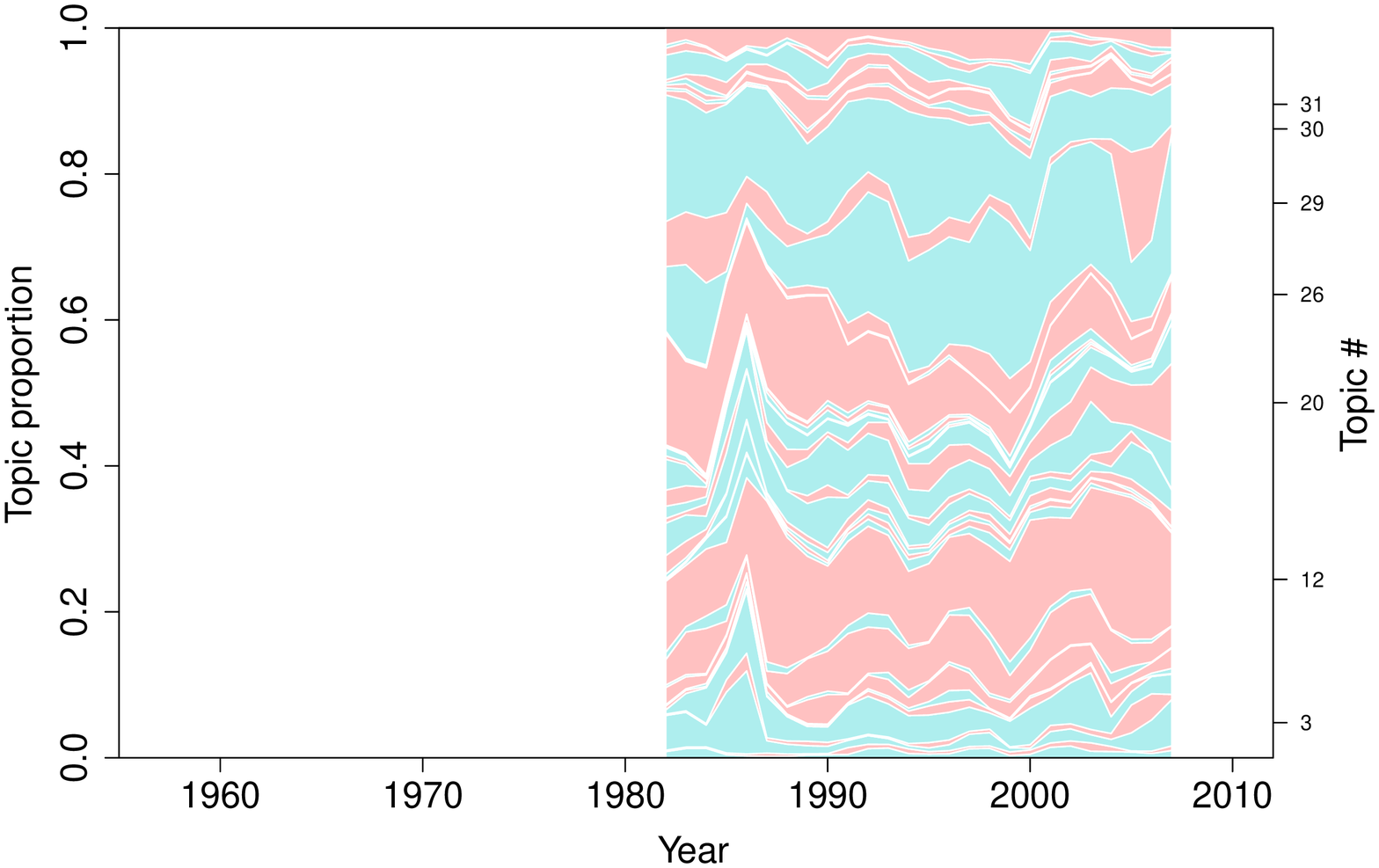} }
\begin{minipage}[b]{5in}
\caption{Journal topic distributions over time for {\it Computers \& Industrial Engineering}, {\it Computers \& Operations Research}, {\it Interfaces}, {\it International Journal of Production Research}, {\it Journal of Global Optimization}, and {\it Journal of Manufacturing Systems}. Topics that have a proportion of at least 0.1 at some point during the journal's history are labeled on the right side of each plot.}
\label{fig:jtopic_dist_overtime1}
\end{minipage}
\end{figure}

\begin{figure}[!ht]
\centering
\subfigure[Journal of Operations Management.]{\includegraphics[width=0.48\textwidth]{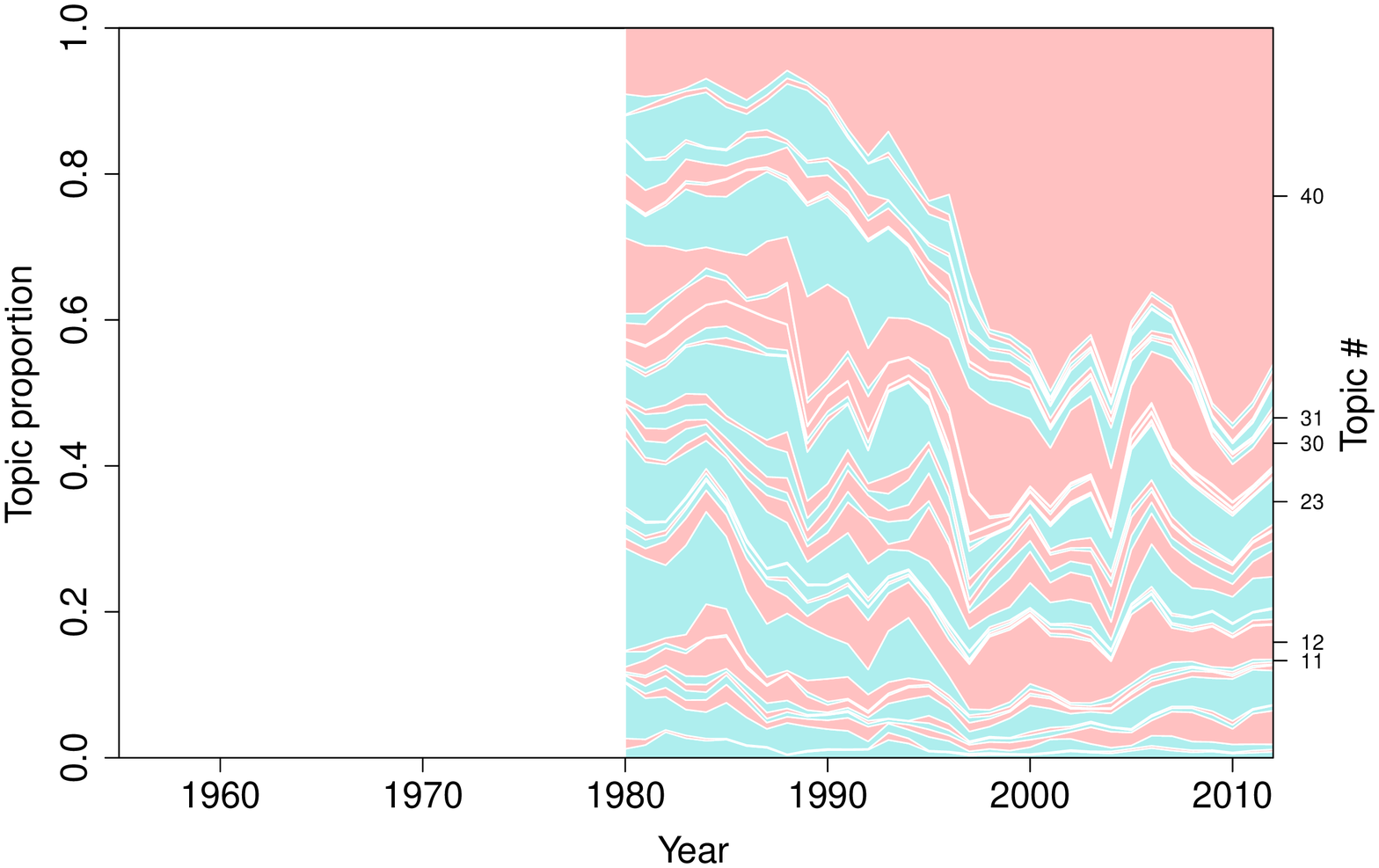} }
\subfigure[Management Science.]{\includegraphics[width=0.48\textwidth]{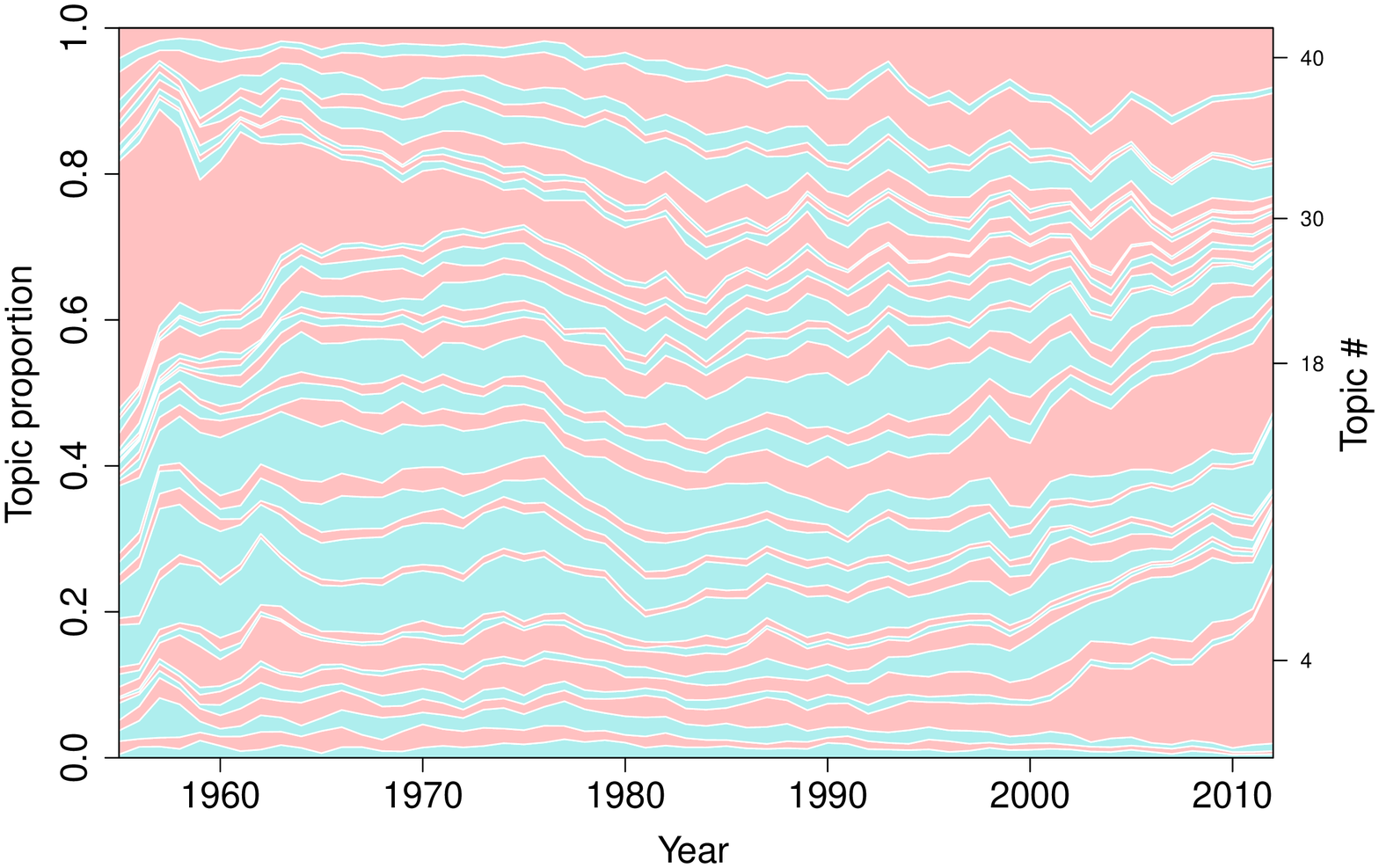} }
\subfigure[Naval Research Logistics.]{\includegraphics[width=0.48\textwidth]{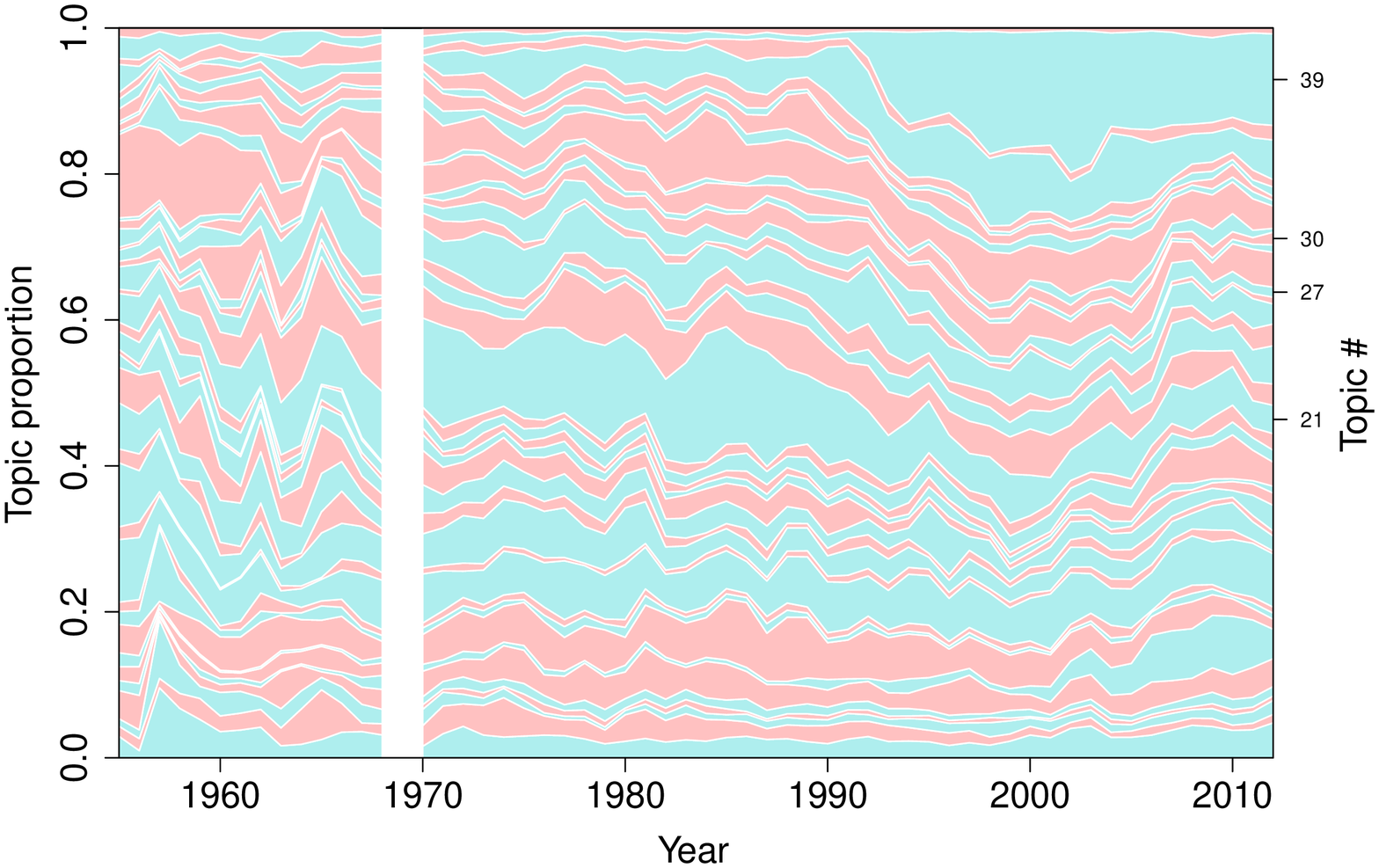} }
\subfigure[Production and Operations Management.]{\includegraphics[width=0.48\textwidth]{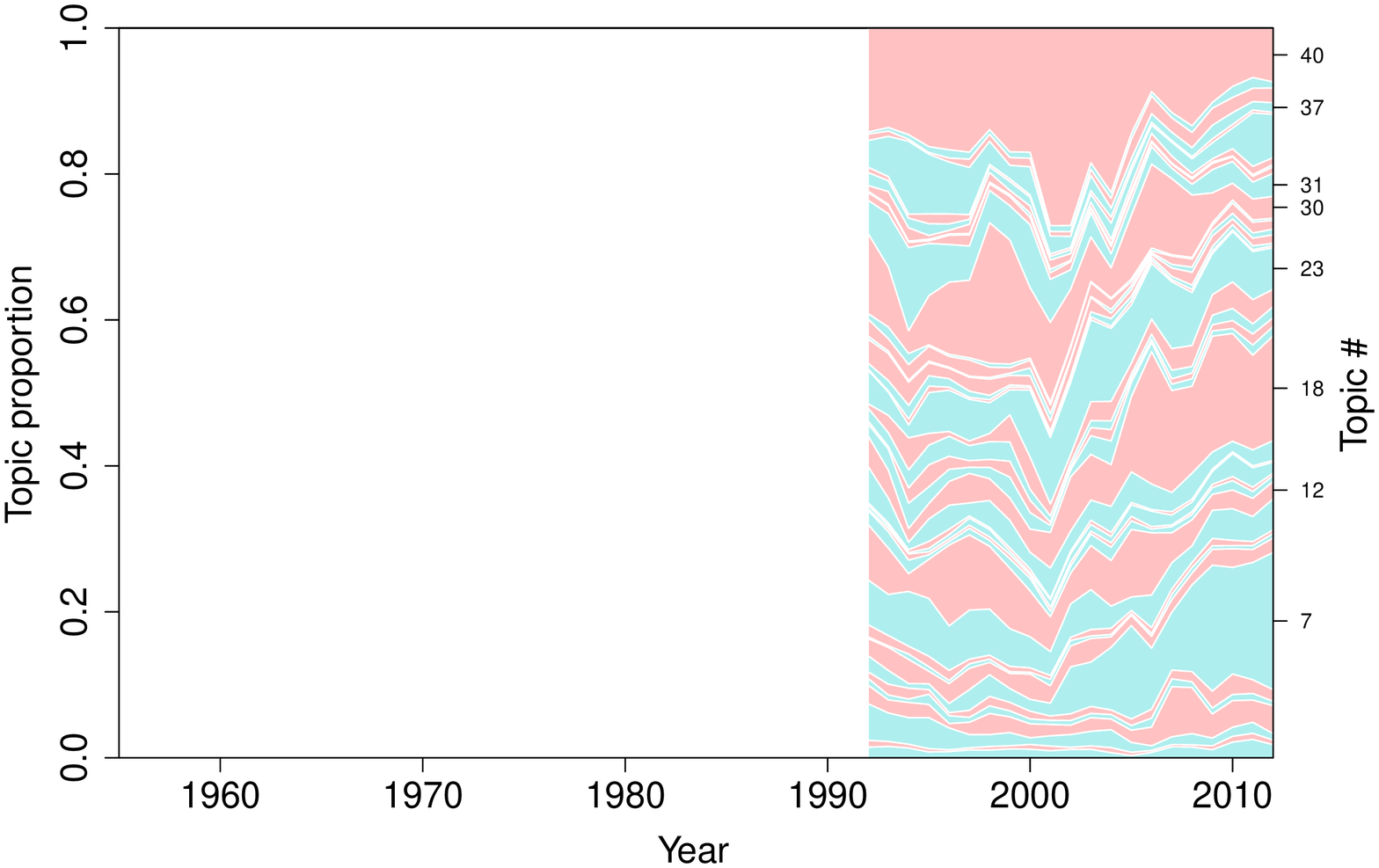} }
\subfigure[SIAM Journal on Optimization.]{\includegraphics[width=0.48\textwidth]{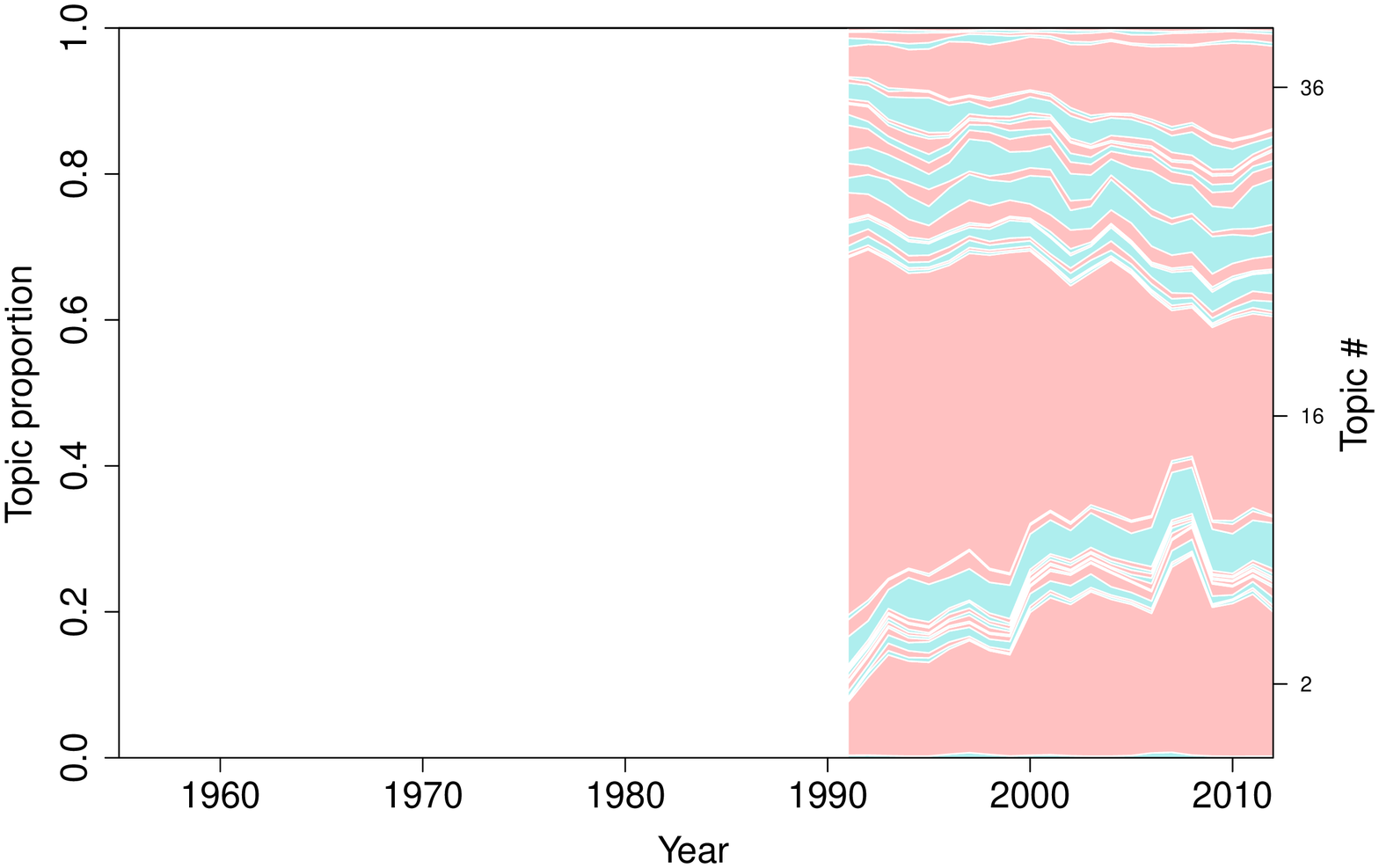} }
\subfigure[Transportation Science.]{\includegraphics[width=0.48\textwidth]{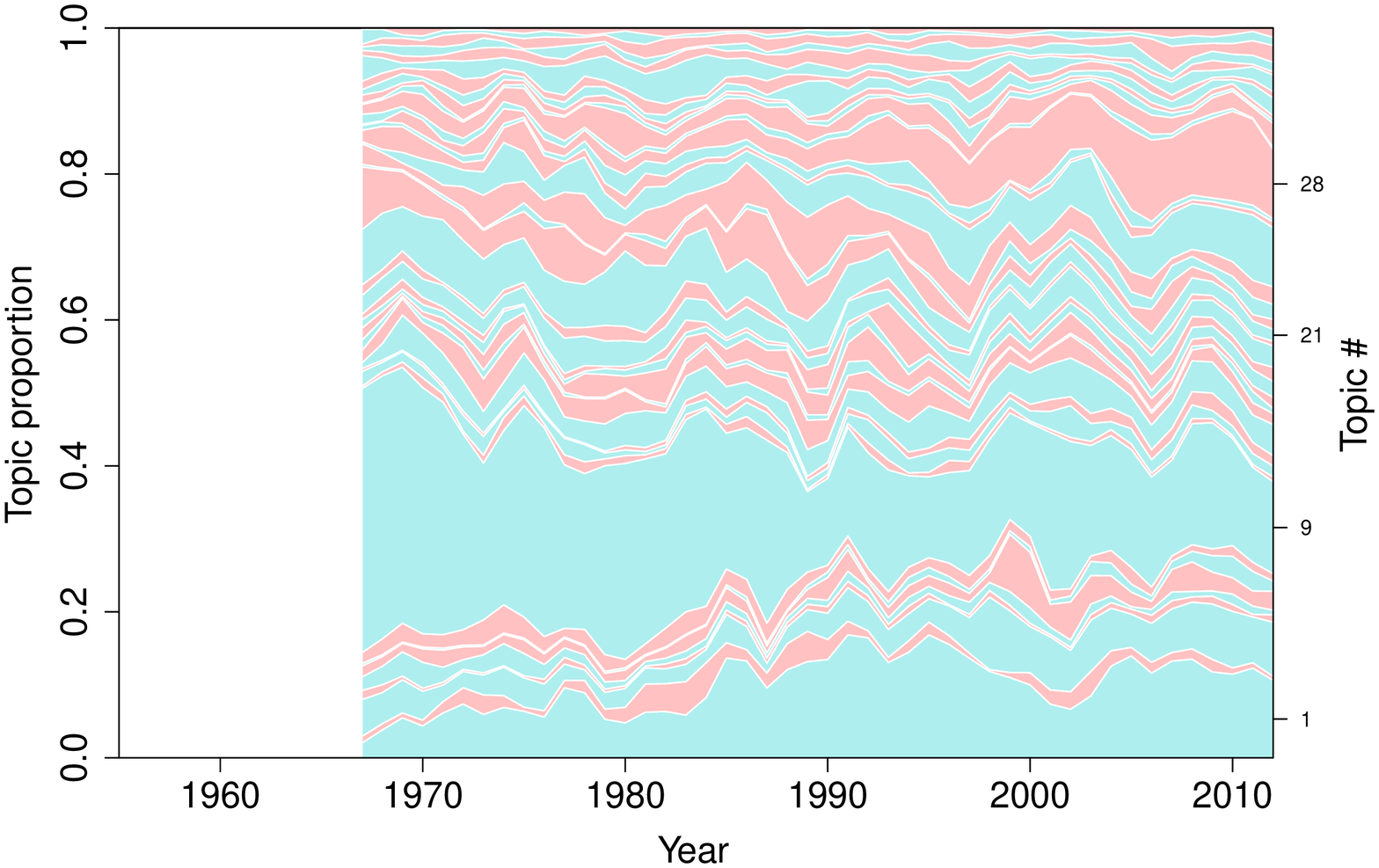} }
\begin{minipage}[b]{5in}
\caption{Journal topic distributions over time for {\it Journal of Operations Management}, {\it Management Science}, {\it Naval Research Logistics}, {\it Production and Operations Management}, {\it SIAM Journal on Optimization}, and {\it Transportation Science}. Topics that have a proportion of at least 0.1 at some point during the journal's history are labeled on the right side of each plot.}
\label{fig:jtopic_dist_overtime2}
\end{minipage}
\end{figure}

Figures \ref{fig:jtopic_dist_overtime1} and \ref{fig:jtopic_dist_overtime2} present temporal topic distributions for 12 journals that have some topic proportions that have either considerably decreased or increased over time. The remainder of the journals' temporal topic distributions are provided in Appendix B. In these figures, topics are numbered 1 through 40 from the bottom to the top, and topics that have a proportion of at least 0.1 at some point during the journal's history are labeled on the right side of each plot. These figures show how some topics that were very well-represented early on in a journal's history may have decreased over time (e.g., topic 26 for {\it Computers \& Industrial Engineering}), as well as topics that have grown considerably over time (e.g., topic 40 for {\it Journal of Operations Management}). These figures also show how a journal's scope may be consistently dominated by a single topic (e.g., topic 16 for {\it SIAM Journal of Optimization}).

Table \ref{tab:jtopicstats} shows the topic numbers for the metrics related to the dynamics of the topic proportions over time. For these computations, topic 30 was not considered because this was deemed to be a general catch-all topic. When considering the maximum topic proportions ($p_{\max}$ and $p^{\textup{mry}}_{\max}$), many journals seem to have consistent primary topics over time up to the most recent time point. When considering all journals as a whole, topic 2 was a dominant topic early on (recall that topic 30 was not considered), and has been the most dynamic topic through time. Topic 28 has had the greatest increase in topic proportion and has the greatest topic proportion in the year 2012.

We attempt to summarize the information in Table \ref{tab:jtopicstats} by tabulating the most frequently occurring topic numbers for each of the topic dynamic metrics, and these tabulations are shown in Table \ref{tab:jtopicstats2}. For example, this table says that there are 4 journals from Table \ref{tab:jtopicstats} where topic 16 has the greatest topic range ($\phi^{(j)}_{\max}$), and all other metrics and counts are interpreted in a similar manner.

\begin{table}[!h]
\centering
{\footnotesize
\begin{tabular}{ l c c c c c }
\hline
Journal & $\phi^{(j)}_{\min / \max}$ & $\psi^{(j)}_{\textup{dec} / \textup{inc}}$ & $\tau^{(j)}_{\min / \max}$ & $p_{\max}$ (yr) & $p^{\textup{mry}}_{\max}$ (yr) \\ \hline
AnnOR & 10 / 21 & 17 / 28 & 10 / 22 & 21 (1987) & 28 (2012) \\
CompIE & 27 / 26 & 26 / 28 & 27 / 26 & 26 (1976) & 28 (2012) \\
CompOR & 7 / 28 & 26 / 28 & 36 / 34 & 28 (2010) & 28 (2012) \\
EJOR & 10 / 25 & 17 / 28 & 10 / 25 & 25 (1978) & 28 (2012) \\
IIE & 27 / 31 & 40 / 20 & 27 / 11 & 31 (1991) & 20 (2012) \\
IJComp & 15 / 16 & 16 / 28 & 39 / 28 & 28 (2008) & 25 (2012) \\
Interf & 24 / 3 & 17 / 3 & 36 / 35 & 17 (1992) & 3 (2012) \\
IJPEcon & 14 / 3 & 20 / 40 & 14 / 20 & 3 (1997) & 40 (2012) \\
IJPRes & 24 / 29 & 29 / 12 & 27 / 29 & 29 (1965) & 12 (2012) \\
JCombOpt & 40 / 33 & 3 / 33 & 18 / 33 & 33 (2007) & 33 (2012) \\
JGlobOpt & 39 / 2 & 16 / 27 & 7 / 2 & 16 (1992) & 27 (2012) \\
JManSys & 27 / 29 & 31 / 29 & 27 / 26 & 29 (1998) & 29 (2006) \\
JOpMan & 27 / 40 & 11 / 40 & 27 / 40 & 40 (2010) & 40 (2012) \\
JOptApp & 39 / 27 & 29 / 27 & 39 / 2 & 2 (1971) & 2 (2012) \\
JSched & 15 / 28 & 37 / 28 & 15 / 37 & 37 (2005) & 37 (2012) \\
ManSci & 24 / 4 & 11 / 4 & 33 / 40 & 4 (2012) & 4 (2012) \\
ManServOM & 33 / 11 & 23 / 35 & 33 / 18 & 18 (2011) & 7 (2012) \\
MathMethOR & 10 / 6 & 37 / 27 & 10 / 33 & 2 (2003) & 27 (2012) \\
MathP & 35 / 13 & 34 / 33 & 35 / 13 & 13 (2004) & 16 (2012) \\
MathPComp & 7 / 16 & 28 / 16 & 40 / 16 & 16 (2012) & 16 (2012) \\
MathOR & 40 / 2 & 2 / 6 & 40 / 27 & 2 (2005) & 27 (2012) \\
MilOR & 11 / 15 & 21 / 19 & 33 / 15 & 15 (1995) & 15 (2012) \\
NavalResLog & 5 / 39 & 13 / 39 & 40 / 21 & 39 (2002) & 39 (2012) \\
Networks & 4 / 33 & 14 / 25 & 35 / 33 & 33 (1987) & 33 (2012) \\
Omega & 36 / 20 & 17 / 28 & 36 / 40 & 40 (2001) & 34 (2012) \\
OpRes & 39 / 22 & 17 / 18 & 10 / 22 & 22 (1964) & 22 (2012) \\
OpResLet & 12 / 33 & 33 / 22 & 12 / 33 & 33 (1981) & 25 (2012) \\
OptLet & 12 / 16 & 36 / 37 & 8 / 16 & 16 (2008) & 27 (2012) \\
ORSpectrum & 4 / 21 & 22 / 28 & 4 / 21 & 21 (1993) & 28 (2012) \\
ProdOpMan & 33 / 40 & 40 / 7 & 33 / 40 & 40 (2001) & 7 (2012) \\
ProdPlanCon & 2 / 40 & 26 / 40 & 9 / 31 & 12 (2004) & 40 (2012) \\
QueueSys & 3 / 22 & 21 / 32 & 39 / 22 & 22 (2002) & 22 (2012) \\
SIAMJOpt & 7 / 16 & 16 / 2 & 7 / 2 & 16 (1991) & 16 (2012) \\
SurvORMS & 8 / 21 & 16 / 22 & 8 / 21 & 21 (2006) & 22 (2012) \\
TransResB & 15 / 9 & 38 / 21 & 29 / 9 & 9 (1988) & 9 (2012) \\
TransResE & 33 / 40 & 40 / 1 & 37 / 40 & 40 (1998) & 9 (2012) \\
TransSci & 4 / 9 & 9 / 28 & 4 / 9 & 9 (1967) & 9 (2012) \\  \hline
All journals & 39 / 2 & 17 / 28 & 7 / 2 & 2 (1973) & 28 (2012) \\ \hline
\end{tabular}
}
\begin{minipage}[b]{5in}
\caption[Journal topic statistics]{For each journal, this table presents the topics that have significant topic proportion dynamics over the course of the publication history; metrics for all journals combined are provided in the last row. See the related text for a description of the topic dynamics.}
\label{tab:jtopicstats}
\end{minipage}
\end{table}

\begin{table}[!h]
\centering
{\footnotesize
\begin{tabular}{c p{4cm} c p{6cm}}
\hline
Metric & Description & Topic numbers & Terms \\ 
& & (journal count) \\ \hline
$\phi^{(j)}_{\min}$ & Min topic range & 27 (4) & \{game(s), equilibrium, dual(ity), paper, solution\} \\
$\phi^{(j)}_{\max}$ & Max topic range & 16 (4) & \{method(s), algorithm, convergence, optimization, global\} \\
$\psi^{(j)}_{\textup{dec}}$ & Decreasing topic change & 17 (5) & \{decision, management, planning, project, development\} \\ 
$\psi^{(j)}_{\textup{inc}}$ & Increasing topic change & 28 (9) & \{algorithm, heuristic, search, solution(s), proposed\} \\
$\tau^{(j)}_{\min}$ & Least dynamic topic & 27 (5) & \{game(s), equilibrium, dual(ity), paper, solution\} \\
$\tau^{(j)}_{\max}$ & Most dynamic topic & 40 (5) & \{performance, study, quality, management, firms\} \\
$p_{\max}$ & Max topic proportion & 16 (4) & \{method(s), algorithm, convergence, optimization, global\} \\
$p^{\textup{mry}}_{\max}$ & Max topic proportion at most recent year & 28 (5) & \{algorithm, heuristic, search, solution(s), proposed\} \\ \hline
\end{tabular}
}
\begin{minipage}[b]{5in}
\caption[Journal topic statistics 2]{Summarization of topic metric counts from Table \ref{tab:jtopicstats}.}
\label{tab:jtopicstats2}
\end{minipage}
\end{table}

\begin{figure}[H]
\centering
\includegraphics[width=0.95\linewidth]{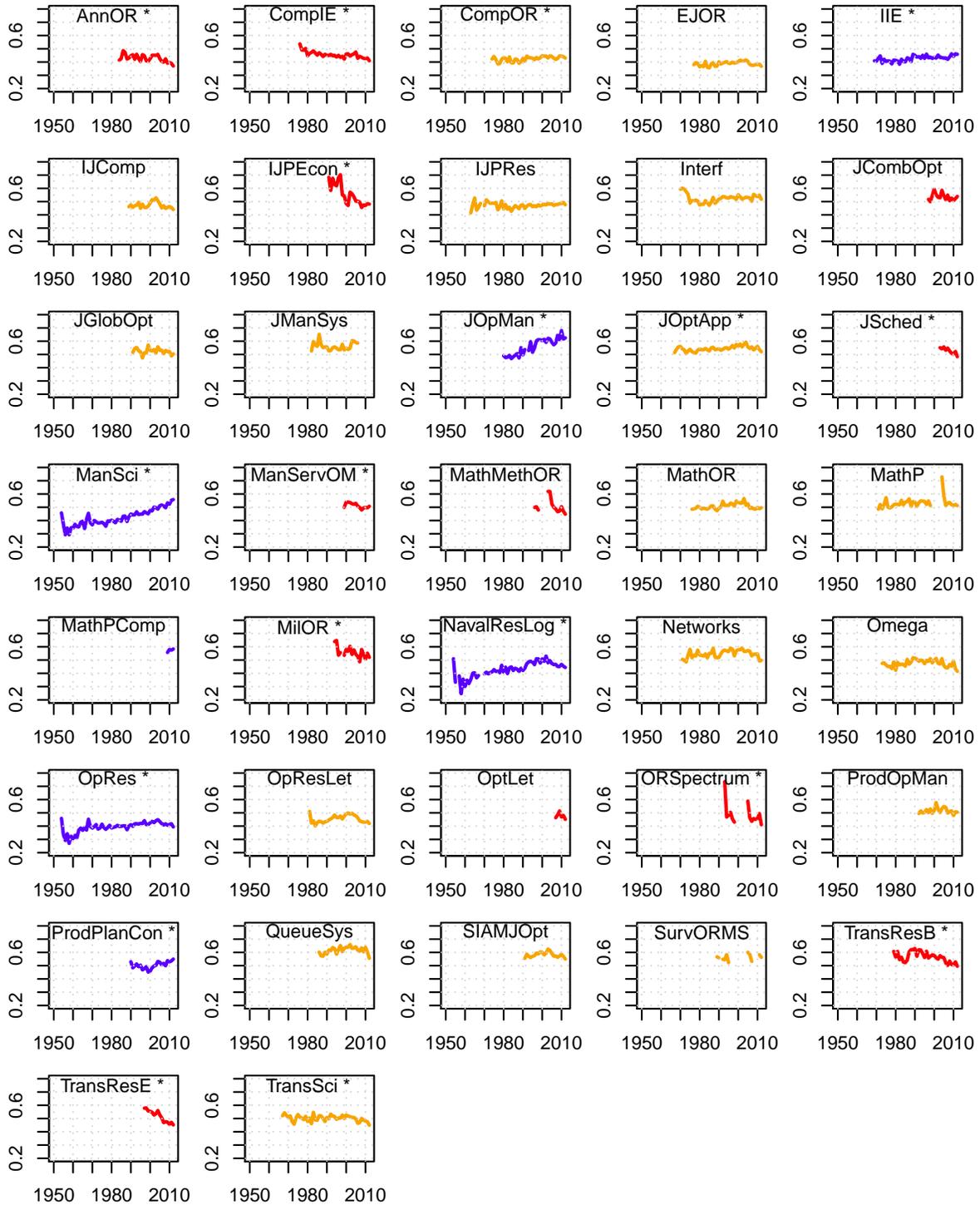}
\begin{minipage}[b]{5in}
\caption{Uniqueness over time for each journal. The temporal uniqueness of each journal is colored by its trend: red -- decreasing trend (12 journals); orange -- neutral trend (18 journals); blue -- increasing trend (7 journals). Journal names that have an asterisk (*) have a statistically significant ($p<0.05$) trend.}
\label{fig:uniquenessovertime}
\end{minipage}
\end{figure}

A detailed analysis of each journal's uniqueness over time is shown in Figure \ref{fig:uniquenessovertime}. At each year, all journals' uniqueness was computed relative to all other journals, and the uniqueness values for each journal are plotted individually. We used a basic linear regression to determine the general trend of each journal's temporal uniqueness, and the slope coefficients for these journals are provided in Table \ref{tab:pvals}. Journals that have a large decreasing trend over time have become less unique and more similar to other journals; journals that have a large increasing trend over time have become more unique and distinct from other journals. Additionally, the journals (and their trends) are colored based on four categories to help distinguish positive versus negative trends and their magnitudes (see figure caption).

\section{Discussion}
\label{sec:discussion}

This work analyzed the OR/MS literature to understand the current content, as well as the historical content, of the published work in these fields. This was a purely observational study based on a descriptive quantitative analysis that attempted to capture and characterize the dynamics of individual journals as well as the field at large. Our analysis is unique in that it is based on textual information from the journal abstracts of a representative set of journals from a single field that extends back to the early years of the field, and thus this data is an extremely rich data set. We hope that the analysis presented herein is insightful to both researchers and journal editors of the OR/MS fields. However, we also feel that the work presented herein merely scratches the surface of this data set, and that future work and exploration could yield additional valuable insights.

We found that the field has been consistently growing since it's inception, and it continues to grow steadily both in terms of the number of published articles and the number of journals. This raises the question of how has this growth affected the impact or quality of research. While the current work is unable to answer this, it would be interesting to determine if the increase in the number of journals and publications are positively and substantially contributing to subsequent

\begin{table}[!h]
\centering
{\footnotesize
\begin{tabular}{ l c c }
\hline
Journal & $m$ & $p$-value \\ \hline
AnnOR & $-1.751 \times 10^{-3}$ & $\mathbf{<0.01}$ \\
CompIE & $-1.632 \times 10^{-3}$ & $\mathbf{<0.001}$ \\
CompOR & $8.102 \times 10^{-4}$ & $\mathbf{<0.001}$ \\
EJOR & $3.628 \times 10^{-4}$ & $0.1604$ \\
IIE & $1.022 \times 10^{-3}$ & $\mathbf{<0.001}$ \\
Interf & $-1.792 \times 10^{-5}$ & $0.9627$ \\
IJPEcon & $-1.014 \times 10^{-2}$ & $\mathbf{<0.001}$ \\
IJPRes & $2.912 \times 10^{-4}$ & $0.1869$ \\
JCombOpt & $-1.054 \times 10^{-3}$ & $0.4841$ \\
JGlobOpt & $-7.025 \times 10^{-4}$ & $0.3543$ \\
JManSys & $1.523 \times 10^{-4}$ & $0.8599$ \\
JOpMan & $5.563 \times 10^{-3}$ & $\mathbf{<0.001}$ \\
JOptApp & $6.416 \times 10^{-4}$ & $\mathbf{<0.01}$ \\
JSched & $-6.712 \times 10^{-3}$ & $\mathbf{<0.01}$ \\
JComp & $-3.381 \times 10^{-4}$ & $0.6364$ \\
ManSci & $3.153 \times 10^{-3}$ & $\mathbf{<0.001}$ \\
ManServOM & $-2.945 \times 10^{-3}$ & $\mathbf{<0.05}$ \\
MathMethOR & $-2.964 \times 10^{-3}$ & $0.3319$ \\
MathP & $8.89 \times 10^{-4}$ & $0.0951$ \\
MathPComp & $7.261 \times 10^{-3}$ & $0.1936$ \\
MathOR & $5.622 \times 10^{-4}$ & $0.0861$ \\
MilOR & $-4.521 \times 10^{-3}$ & $\mathbf{<0.01}$ \\
NavalResLog & $2.487 \times 10^{-3}$ & $\mathbf{<0.001}$ \\
Networks & $6.359 \times 10^{-4}$ & $0.0822$ \\
Omega & $-3.533 \times 10^{-5}$ & $0.9237$ \\
OpRes & $1.462 \times 10^{-3}$ & $\mathbf{<0.001}$ \\
OpResLet & $4.105 \times 10^{-4}$ & $0.4522$ \\
OptLet & $-3.44 \times 10^{-3}$ & $0.5716$ \\
ORSpectrum & $-6.406 \times 10^{-3}$ & $\mathbf{<0.01}$ \\
ProdOpMan & $3.887 \times 10^{-5}$ & $0.9625$ \\
ProdPlanCon & $1.962 \times 10^{-3}$ & $\mathbf{<0.05}$ \\
QueueSys & $7.557 \times 10^{-4}$ & $0.242$ \\
SIAMJOpt & $-9.298 \times 10^{-5}$ & $0.8889$ \\
SurvORMS & $9.713 \times 10^{-4}$ & $0.2987$ \\
TransResB & $-2.039 \times 10^{-3}$ & $\mathbf{<0.01}$ \\
TransResE & $-8.893 \times 10^{-3}$ & $\mathbf{<0.001}$ \\
TransSci & $-4.932 \times 10^{-4}$ & $\mathbf{<0.05}$ \\  \hline
\end{tabular}
}
\begin{minipage}[b]{5in}
\caption[Uniqueness p-values]{Linear regression slope ($m$) coefficients and significance of uniqueness trends. A bold $p$-value indicates $p < 0.05$.}
\label{tab:pvals}
\end{minipage}
\end{table}

\noindent works, or if the majority of these works only make a minuscule or negligible epistemic contribution to the field.

We found that the OR/MS fields has grown to a very uniformly distributed representation of topics today. The large proportion of topic 30 (which contains very general research/literature terms) in the earliest years and the subsequent decline in proportion is likely due to the change in actual research content that is presented in the abstracts. At the latest time point (2012), there is not a single topic that clearly dominates the field. It is interesting that while, as participants and observers of the OR/MS community, there are trends in both research and funding initiatives, we also see that core models and methods often remain. The fact that the temporal topic distributions do not have large fluctuations in the latest years may suggest that the field progresses via the collective effort of the community in many diverse areas, rather than via ground-breaking novelties. However, we also acknowledge that the field has undergone a rebranding effort with terms such as `Analytics' and `Systems Engineering'. One could argue that, based on the stability of the topic proportions in recent years, the rebranding is more of a change in verbiage rather than a significant shift in the focus of the field.

While many of the journals have relatively flat uniqueness curves, some have rather large trends or sharp peaks or changes, and all of these features could potentially tell a story. We could speculate that journals that have consistent uniqueness over time may have maintained their niche of the field (e.g., {\it IIE Transactions}, {\it SIAM Journal on Optimization}, {\it Surveys in Operations Research} and {\it Management Science}). Journals that have some non-zero trend in either direction may have slowly been introducing/removing different topics from their scope. Similarly, the individual topic distributions over time for each journal may also be telling of a story of the journal's state and scope (e.g., {\it Journal on Optimization Theory and Applications}, {\it International Journal of Production Research}, {\it Transportation Research Part E}). Changes in a journal's uniqueness or in the topic distributions may be due to a change in editors, or the appearance/disappearance of another journal. However, although it would be extremely interesting, a thorough analysis of the driving factors behind these changes is beyond the scope of this work.

Contrary to the stability of the temporal topic proportions as a whole, the temporal topic proportions of the individual journals can be quite dynamic. We see that a single topic, or a few topics, can dominate a journal's content, and that the historical prevalence of these topics can be quite dynamic with significant changes in their topic proportions over time. These topic dynamics are quite interesting and stimulate questions regarding the reasons behind the characteristics that are seen: Are sudden and dramatic shifts in topic proportions related to changes of the journal's editorial board? Are increasing or decreasing topic proportions motivated by changes in funding initiatives? Our analysis does not have the ability to answer these questions, however, the current work is useful for identifying changes in a journal's composition that stimulate historical interest and inquiry.

The fields of operations research and management science are application-driven fields. Some of the topics found from the model are well-aligned with these applications. For example, topic 1 for routing, topic 9 for transportation, and topic 15 for finance and business. However, many topics do not have clear and concise application areas. Instead, many of the topics represent methodologies developed to solve problems in different applications. For instance, topic 13 for (linear) programming, topic 22 for queuing methods, topic 27 for game theory, and topic 33 for graph problems. It could be said that, as a community, we have developed a core set of competencies and methodologies that are used to solve problems in many different application areas. A more detailed analysis of our model results could yield additional insight though, and could further support the claims of historical surveys of the field of OR, such as the use of 'hard' versus 'soft' OR methods and the differences and similarities between the British and American perspectives \cite{Ackoff1987, Reisman1994, Kirby2000, Kirby2007}.

As the OR/MS fields continue to grow, a common discussion point among members of this community is whether or not the field is adequately and sufficiently covered by the current population of journals. Our analysis indicates that there are very few journals that are distinct to a single topic or a couple of topics. This could suggest that the introduction of new, focused journals are justified, if there is a sufficient active subcommunity of researchers in these topics and subfields. The current analysis as presented is not detailed enough to answer these questions, however, an extended analysis based on the same topic model has the ability to provide valuable insight to the discussions of creating new journals, specifically with regards to the scope and focus of the journal. On the contrary, our analysis is also able to highlight pairs or groups of journals that have substantially similar and possibly redundant content, which may support the removal of a journal.

From and author's perspective, the analysis (specifically Figure \ref{fig:journaltopicdist}) presented herein could provide valuable guidance when considering where to submit new and original research. As an author, the lack of very specialized journals, or finding journals with similar content, could be advantageous in the sense that there could be multiple publication outlets for the same work. Furthermore, the current work could be extended to assist authors by helping them select the most appropriate publication outlet for a new and original work. Our results may be especially useful when new research is specific in terms of it's application or methodology, or perhaps when new research spans multiple areas, yet it does not seem to fit appropriately based solely on the description of the journal's scope. An {\it a posterior} topic distribution could be computed for the abstract of a pre-submission manuscript, and the resulting topic distribution could be compared to those of the topic distributions of the journals from our analysis. A ranked list of journals could then be generated for which the new work may potentially be appropriate.

This work can also be used in a wide variety of capacities for journal editors and publishers. It can be used to confirm that a journal's content is consistent with it's intended scope, and furthermore, to determine if a journal's actual scope is in line with the editor's intended scope. Conducting specific studies assessing the publicized scope versus the published literature could highlight efficiencies or deficiencies in the editorial and review process. Our analysis can also be used to help identify topic areas that are growing or that are under-represented, and to find niche topic areas that other journals do not cover. This may help journals encourage research in a particular area for which they want to be a leading publication. Furthermore, this work of using topic analysis to understand the content of journals and a field as a whole may be of interest to those in other fields as well, as the methodology we present can be applied to any field, given that the data set of manuscript abstracts is available.

The field of topic modeling has suggested guidelines and common procedures, though there are few established methodologies. Consequently, the models must be subjectively analyzed to ensure their credibility. We found that some journals have very broad scopes and publish articles from many different subject areas, whereas others are very specific in their scope, and this is consistent with our intuition regarding these journals. Based on hierarchical clustering, we also found pairs and groups of journals that had similar content, which were also consistent with our intuition. Additionally, the topics seem to be acceptably unique and the prominent words in each topic are cohesive in content with regards to the field. We acknowledge that, because topic modeling is somewhat of an art, though there are some quantitatively-supported guidelines for some methodologies \cite{Taddy2012, Wallach2009} and some work comparing topic similarity methods \cite{Aletras2014}, preprocessing and modeling of the data will not be perfect, these intuitively consistent observations add to the credence and believability of the model, and thus in our subsequent analysis.

\subsection{Limitations}

This work has been able to provide a wealth of data and knowledge about the content of OR/MS journals, yet there are a number of limitations. The size of the data set used in this work required custom and automated preprocessing to sort content articles from non-content articles and to pare down the abstract text to that which was input to the topic model. Such a procedure cannot guarantee perfect preprocessing, but we are confident of the final data that was used in the model, based on selected manual inspection of the data. We are also confident in the resulting topic model based on our analysis and the consistency of the model with our intuition. Additionally, the size of the data limited the size of the model (namely, the number of topics $K$), though we still feel that the current model sufficiently models the original texts. A larger model, or one that includes additional abstracts, would certainly require some form of parallelization \cite{Liu2011}.

An LDA model, by nature, finds latent topics in the documents from which it is developed. Yet, these topics may not clearly align with topics that are conceived by humans such that each individual topic has tight, semantically-related content. Furthermore, the topics that are found by the model may not be clear-cut and precisely coincide with those that are traditionally thought of when considering OR/MS literature. A possible aid to including such human-conceived topics is to use seeded topics as done by Hall and colleagues \cite{Hall2008}. However, manual inspection shows that the topics that were found in this work still are quite coherent in their content.

As well, the LDA model has a number of parameter settings that are user-defined and that affect the resulting topic model, and this is an issue for all topic models. In particular, and for the topic model package used in this work, the main user-defined parameters include the number of topics $K$ and the Dirichlet prior $\alpha$ that defines the topic distribution. The model presented herein was selected, based on manual inspection and review, from a set of models that used different values for $K$ and $\alpha$. While a thorough study of the effects of changing these parameters (and others) would be useful, it is beyond the scope of this work.

\subsection{Future directions}

The goal of observational studies is often to gain an understanding of a process or phenomenon. However, observational analyses also often lead to additional questions and research, and there are many future directions and potential applications for this work. The data set used and the resulting topic model is a rich source of information that is likely of interest to many in the operations research and management science communities. Consequently, there were many ways to analyze and display the results, all of which could not be shown here. A more dynamic presentation of this data, such as an interactive tool similar to that of \cite{Gunnemann2013}, would certainly be a great tool in the dissemination of this work, and this may be the focus of future work.

A possible future direction of this work would be to relate the growth of journals and their content to measures of scientific impact. We acknowledge that many measures have been criticized for one reason or another \cite{Campbell2008}, and that there is no single and universally accepted measure of scientific impact \cite{Bollen2009}. However, such an analysis, possibly using a series of metrics, may be able to shed light on which topics have both positively and negatively contributed to a journal's scientific impact as well as on the growth of the field. A preliminary analysis relating journal impact factor (from {\tt www.scimagojr.com}) showed that, as a whole, there is no significant relationship between journal uniqueness and impact factor for the set of journals included in this work, although there may be relationships between these two measures for individual journals. Another use of the current model could be to use it to determine the most appropriate journal for a new work. This could be done by determining the posterior probabilities of the abstract text of a new journal article and matching these topic proportions to those of the journals from our model. The purpose of this use would be to identify the journal with the most similar content as the new work, with the idea that the potential for publications is greatest in the best matched journal.

We refrain from using the term \emph{evolution} with regards to the content and dynamics of the literature because this term often conjures biologically-related notions of content breeding, exchange, and extinction, etc. An evolutionary analysis of the epistemology of the OR/MS fields, however, would be extremely interesting. In this case, we would be interested in understanding how knowledge and content has grown from itself and new ideas \cite{Radnitzky1993}. Such an analysis is likely possible with the current data set, though with other modeling and analysis methodologies.

\section{Conclusion}

We have created a topic model of the scientific literature from the fields of operations research and management science. Our analysis has highlighted the growth of these fields, in terms of journals and articles, and illustrated how the content of these fields has changed over time. We have characterized journals by the dynamics of their content over time and we have clearly shown how some journals have very wide scopes whereas others are very focused in their content. These observations are consistent with intuition, which gives us confidence in the model in order to look for new and interesting artifacts. There are many possible directions for further investigation of the data set used herein and the model created, both to identify and understand the history of the OR/MS fields but also to aid researchers in understanding these fields. \\

\noindent \textbf{Acknowledgment:} The authors would like to thank all journal publishers for their assistance in obtaining the metadata used in this work, including: Elsevier ({\tt www.elsevier.com}); INFORMS (Institute for Operations Research and the Management Sciences, {\tt www.informs.org}); MORS (the Military Operations Research Society, {\tt www.mors.org}); Palgrave Macmillan ({\tt www.palgrave.com}); SIAM (Society for Industrial and Applied Mathematics, {\tt www.siam.org}); Springer \newline ({\tt www.springer.com}); Taylor and Francis ({\tt www.tandfonline.com}); and Wiley ({\tt www.wiley.com}). Additionally, we greatly appreciate the thoughtful discussions and input from Dr. Thomas Willemain, Dr. William A. Wallace, and Dr. David Mendon\c{c}a.\\

\clearpage



\newpage
\clearpage

\newgeometry{bottom=1in, top=1.25cm}

\section*{Appendix A: Wordclouds}

\begin{figure}[!hb]
\center
\subfigure[Topic \#1.]{\includegraphics[trim=1.2cm 1.2cm 1.2cm 1.2cm, clip=true, width=0.4\textwidth]{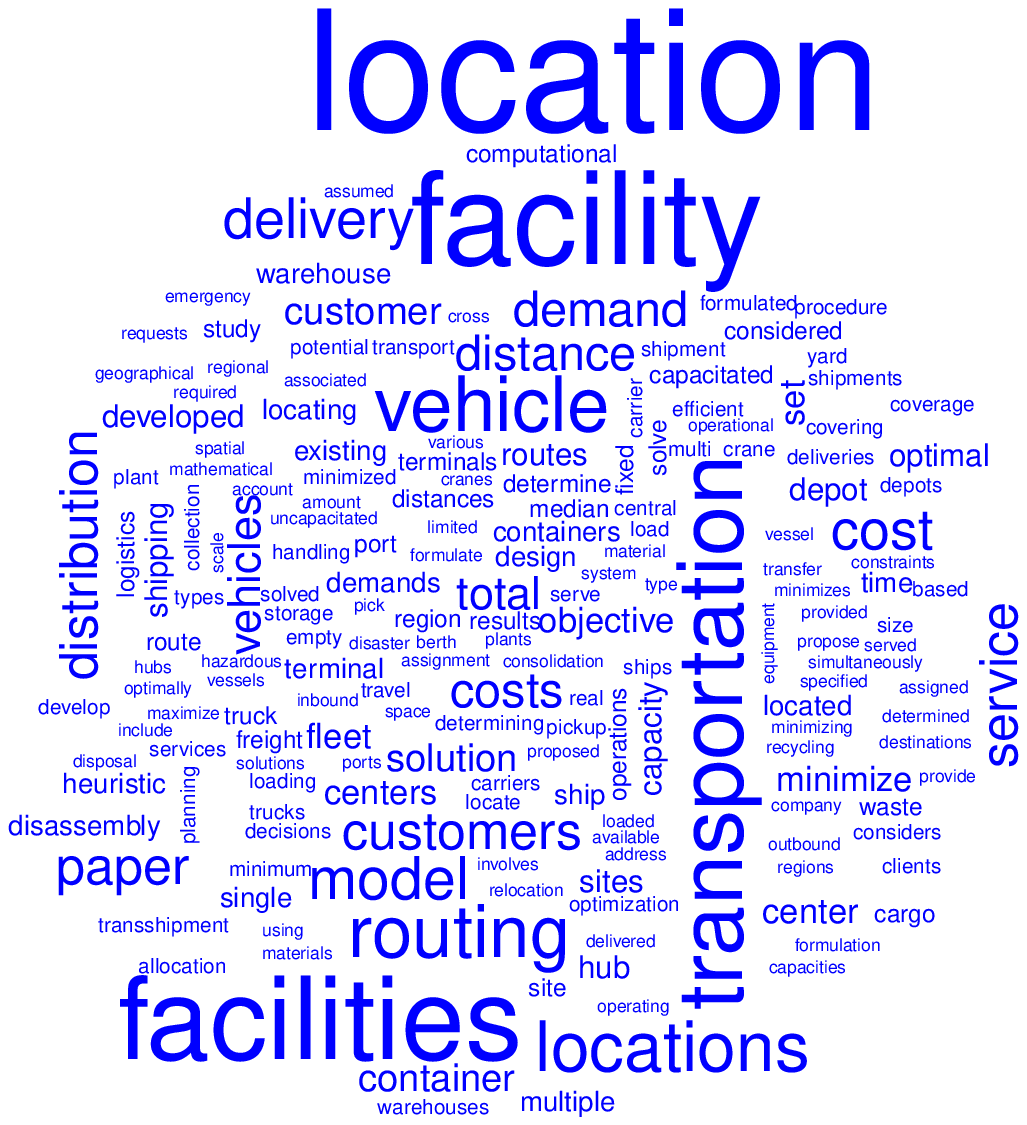} }
\subfigure[Topic \#2.]{\includegraphics[trim=1.2cm 1.2cm 1.2cm 1.2cm, clip=true, width=0.4\textwidth]{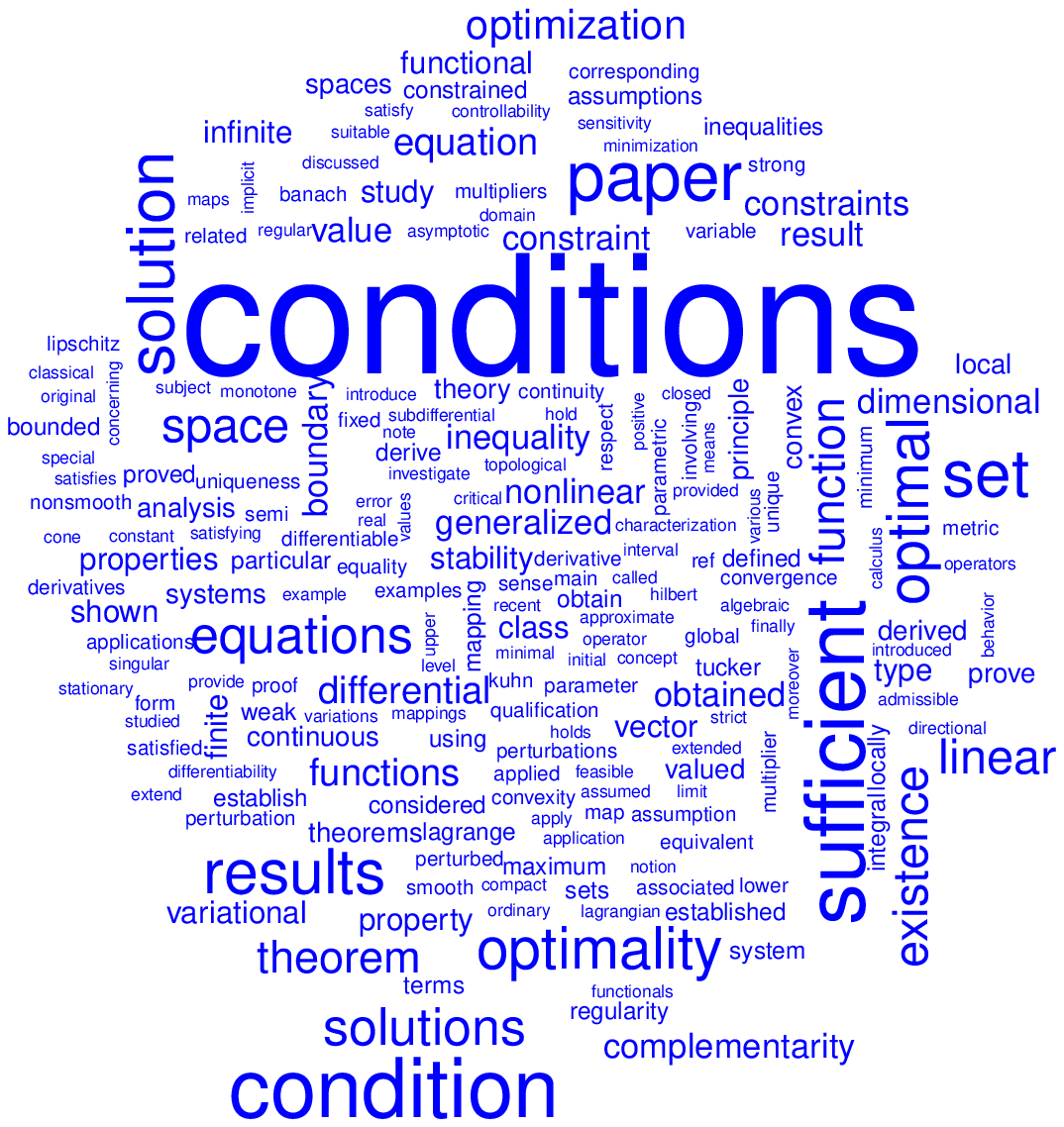} } \\
\subfigure[Topic \#3.]{\includegraphics[trim=1.2cm 1.2cm 1.2cm 1.2cm, clip=true, width=0.4\textwidth]{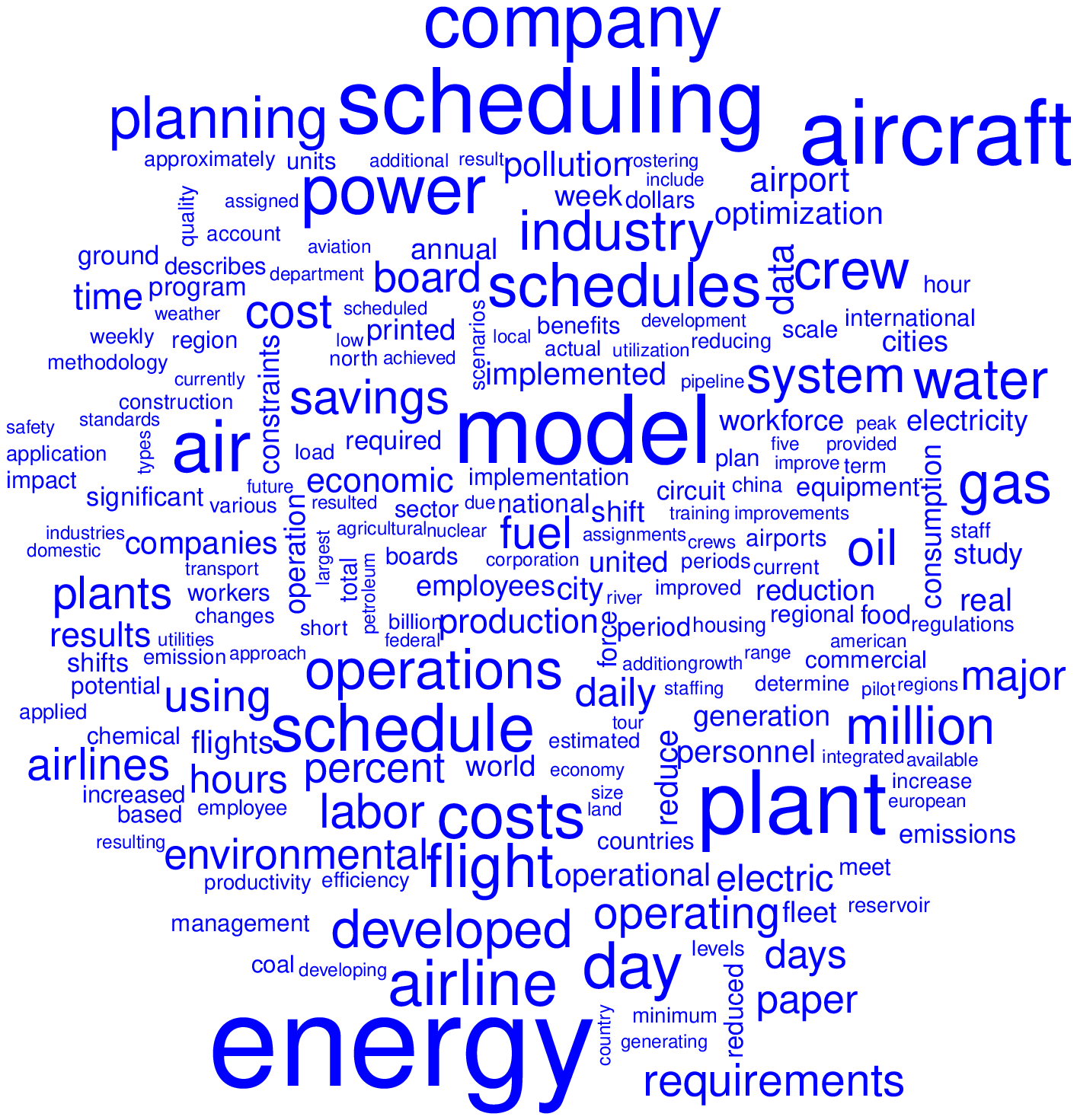} }
\subfigure[Topic \#4.]{\includegraphics[trim=1.2cm 1.2cm 1.2cm 1.2cm, clip=true, width=0.4\textwidth]{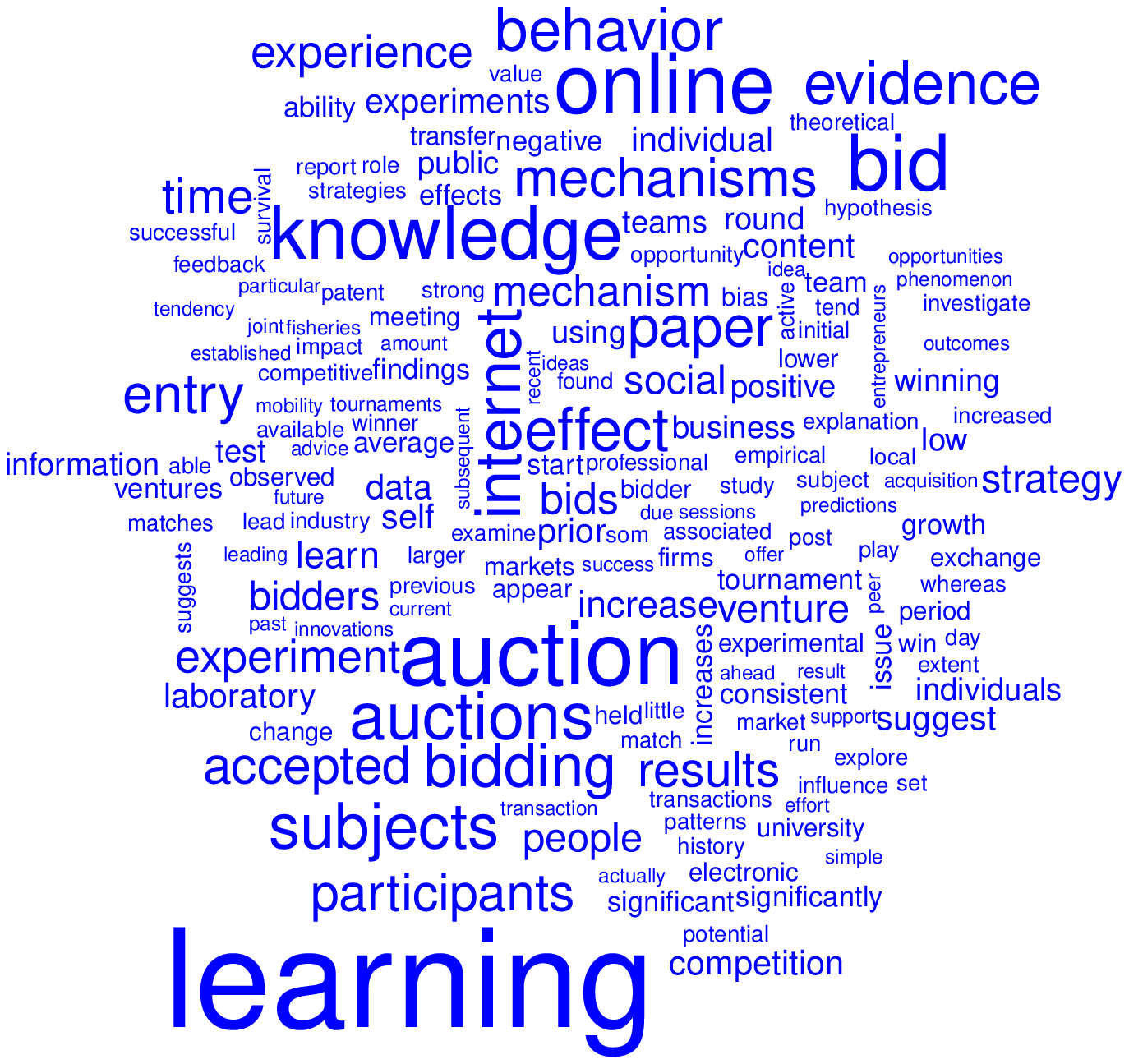} } \\
\subfigure[Topic \#5.]{\includegraphics[trim=1.2cm 1.2cm 1.2cm 1.2cm, clip=true, width=0.4\textwidth]{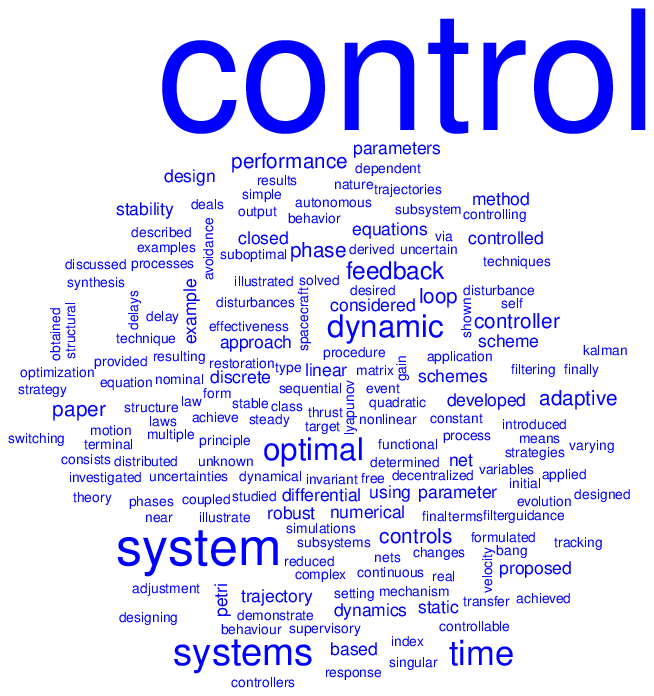} }
\subfigure[Topic \#6.]{\includegraphics[trim=1.2cm 1.2cm 1.2cm 1.2cm, clip=true, width=0.4\textwidth]{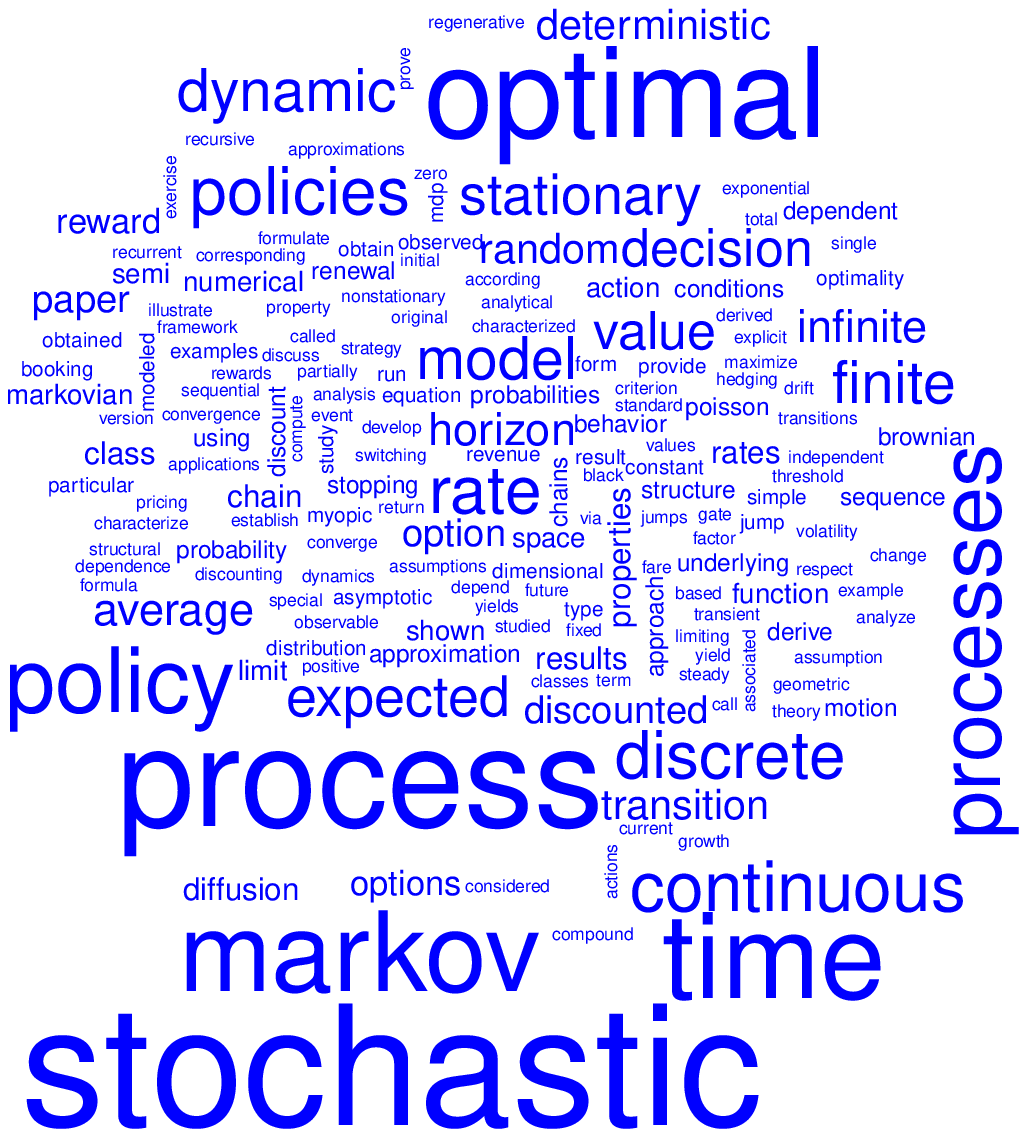} }
\caption{Wordclouds for topics 1--6.}
\label{fig:wordclouds1}
\end{figure}

\restoregeometry
\clearpage
\newpage

\begin{figure}[!h]
\centering
\subfigure[Topic \#7.]{\includegraphics[trim=1.2cm 1.2cm 1.2cm 1.2cm, clip=true, width=0.4\textwidth]{wordcloud_topic7.eps} }
\subfigure[Topic \#8.]{\includegraphics[trim=1.2cm 1.2cm 1.2cm 1.2cm, clip=true, width=0.4\textwidth]{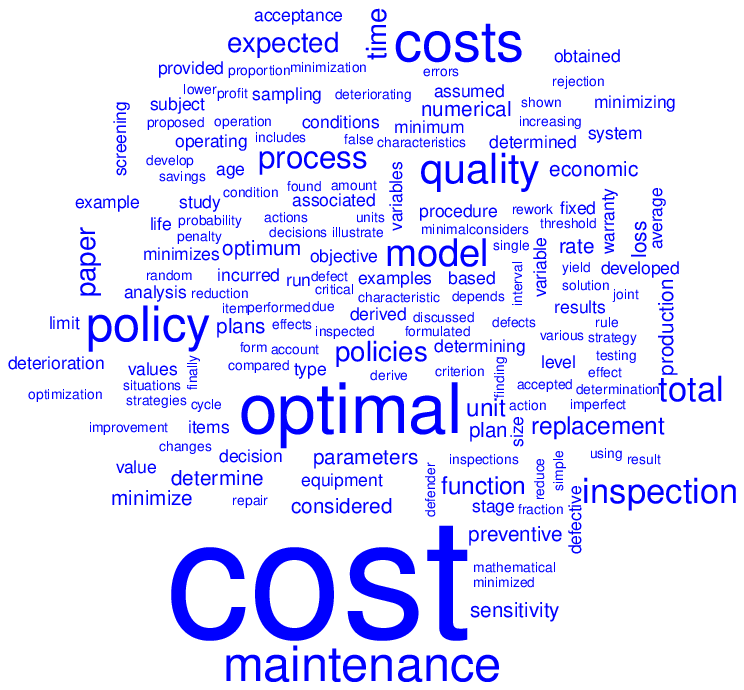} } \\
\subfigure[Topic \#9.]{\includegraphics[trim=1.2cm 1.2cm 1.2cm 1.2cm, clip=true, width=0.4\textwidth]{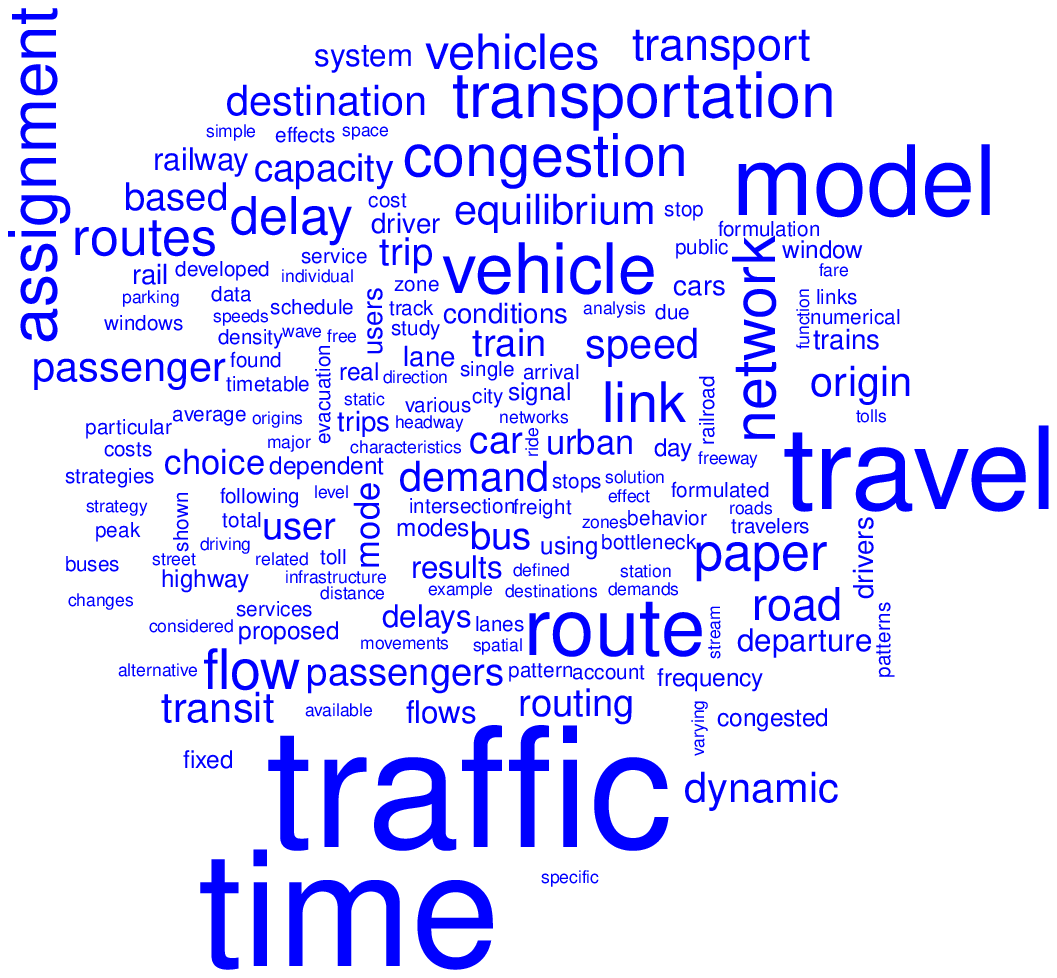} }
\subfigure[Topic \#10.]{\includegraphics[trim=1.2cm 1.2cm 1.2cm 1.2cm, clip=true, width=0.4\textwidth]{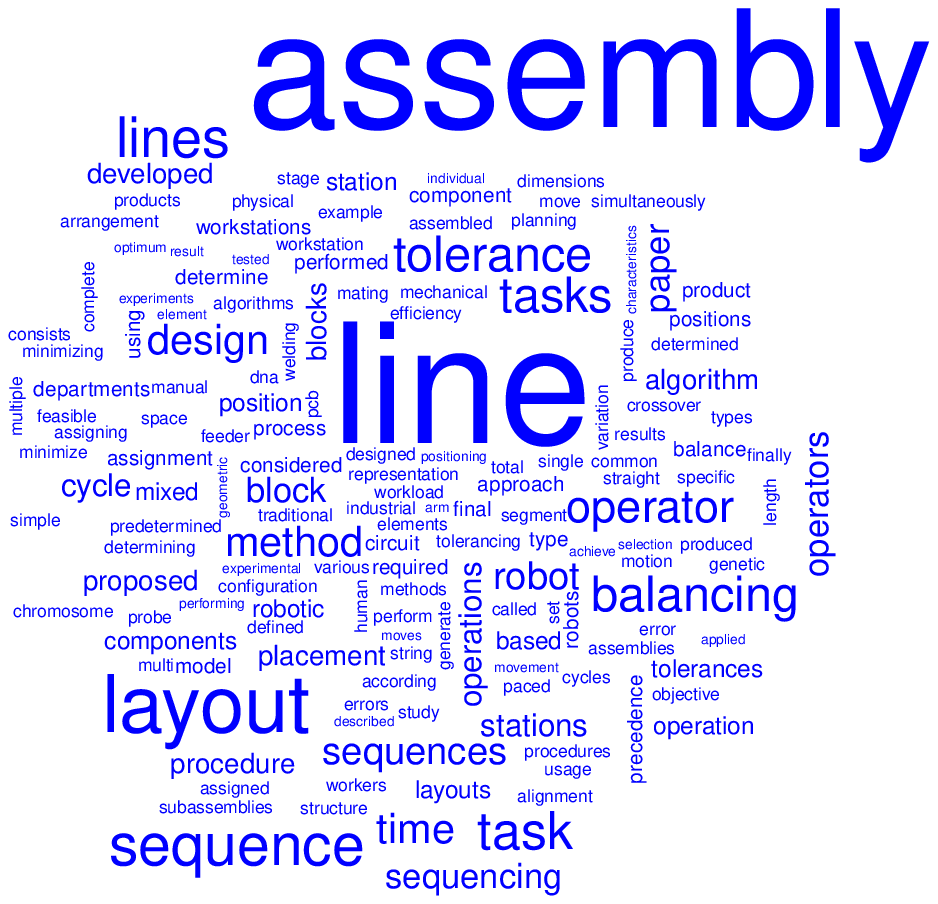} } \\
\subfigure[Topic \#11.]{\includegraphics[trim=1.2cm 1.2cm 1.2cm 1.2cm, clip=true, width=0.4\textwidth]{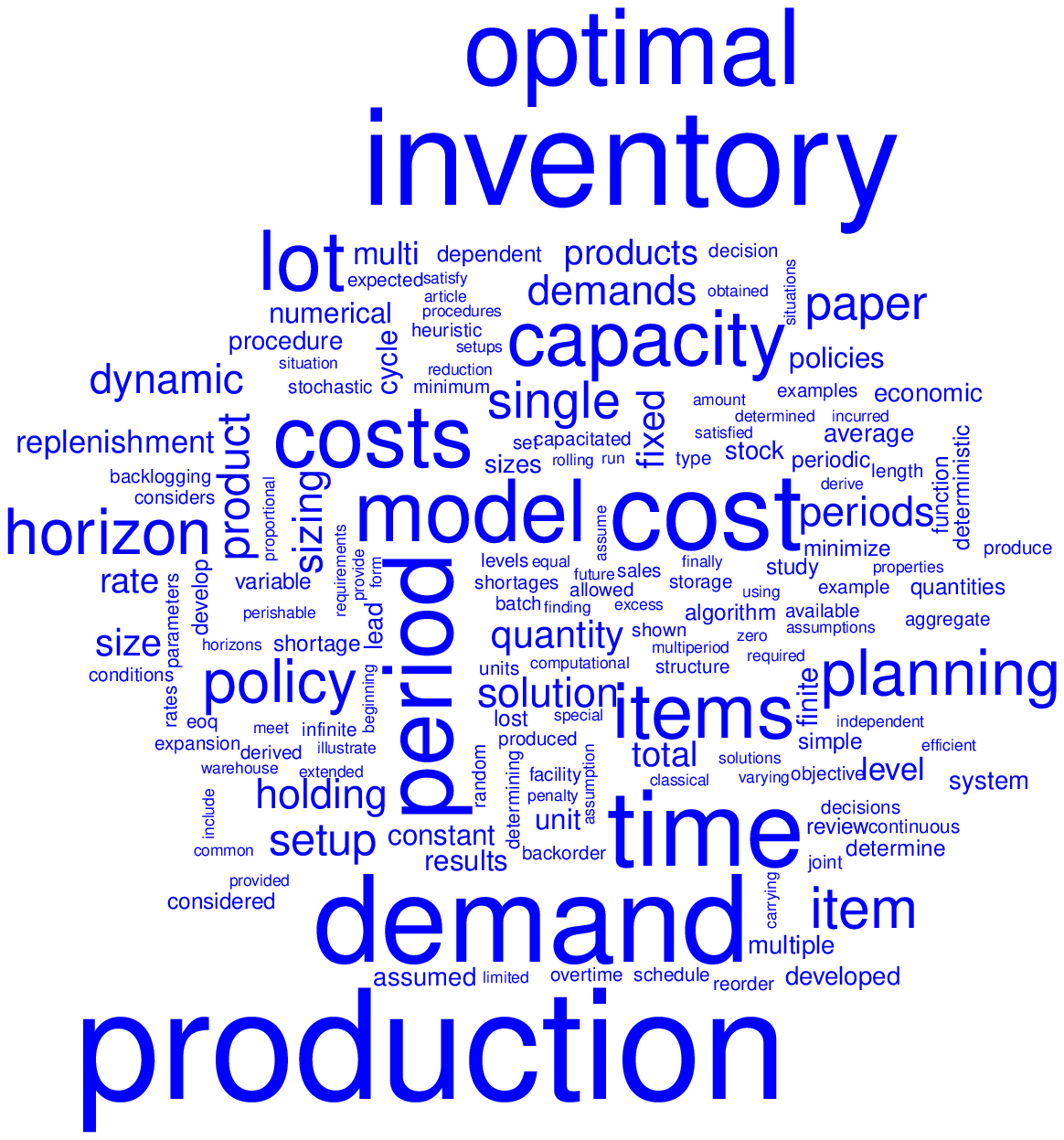} }
\subfigure[Topic \#12.]{\includegraphics[trim=1.2cm 1.2cm 1.2cm 1.2cm, clip=true, width=0.4\textwidth]{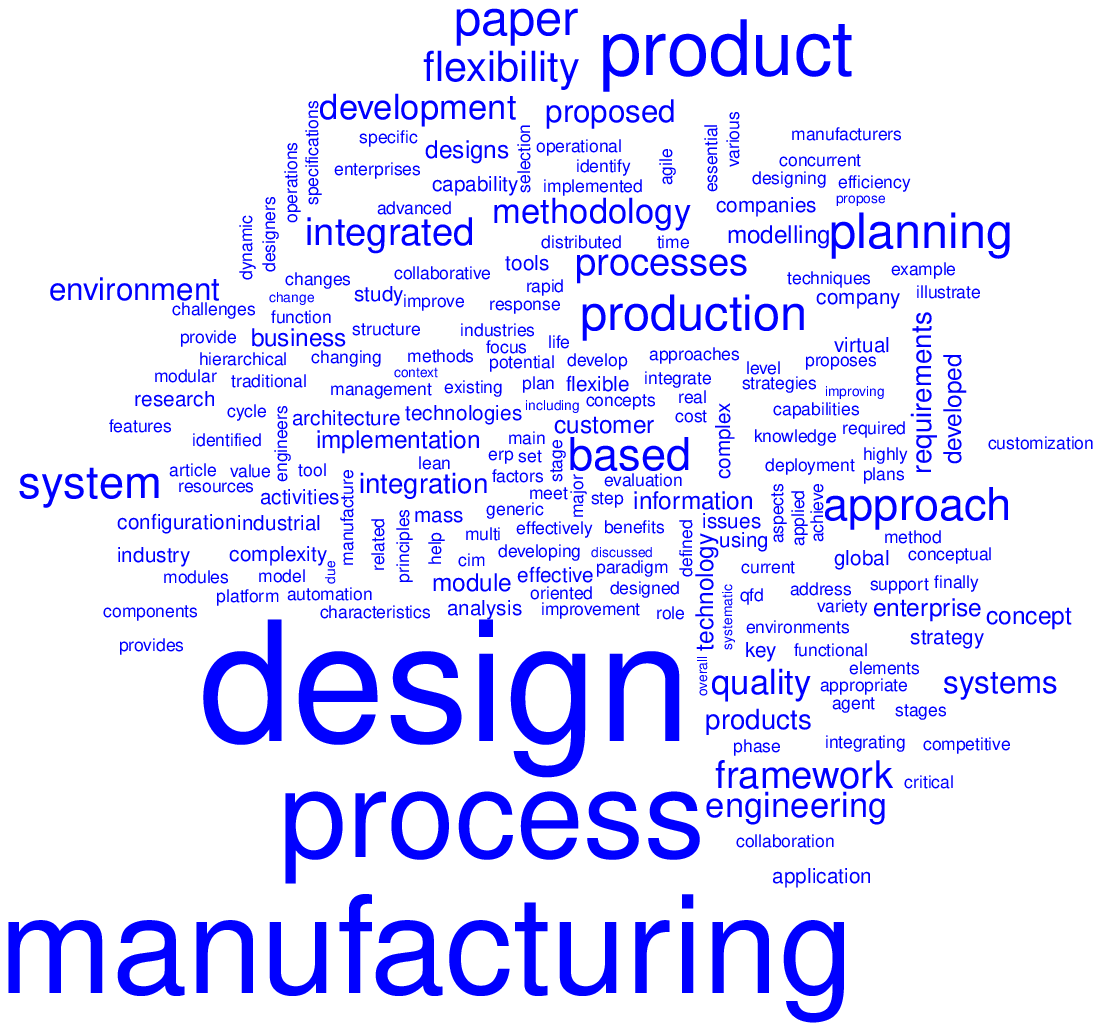} }
\caption{Wordclouds for topics 7--12.}
\label{fig:wordclouds2}
\end{figure}

\newpage

\begin{figure}[!h]
\centering
\subfigure[Topic \#13.]{\includegraphics[trim=1.2cm 1.2cm 1.2cm 1.2cm, clip=true, width=0.4\textwidth]{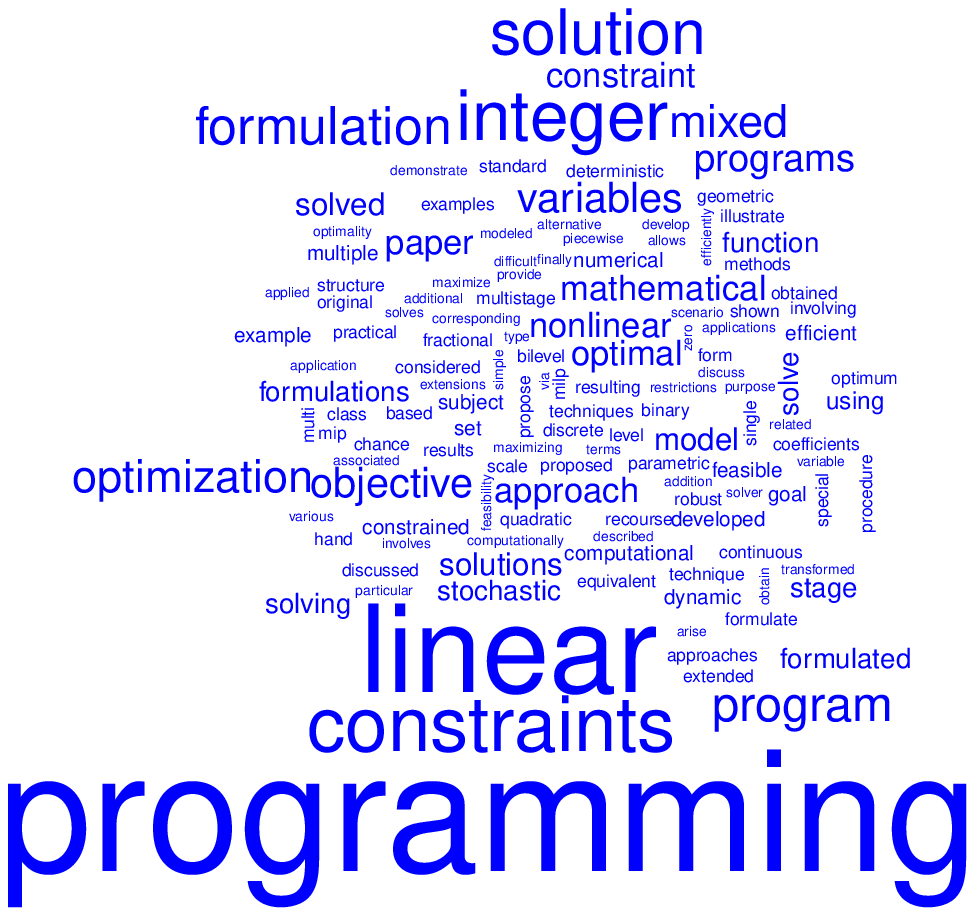} }
\subfigure[Topic \#14.]{\includegraphics[trim=1.2cm 1.2cm 1.2cm 1.2cm, clip=true, width=0.4\textwidth]{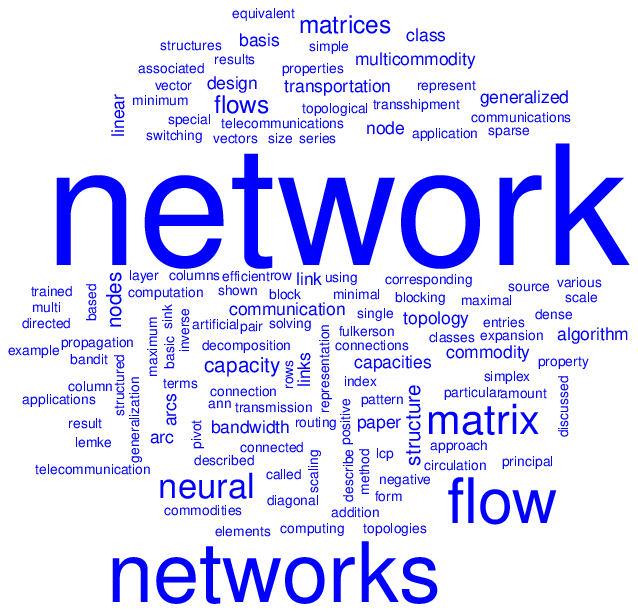} } \\
\subfigure[Topic \#15.]{\includegraphics[trim=1.2cm 1.2cm 1.2cm 1.2cm, clip=true, width=0.4\textwidth]{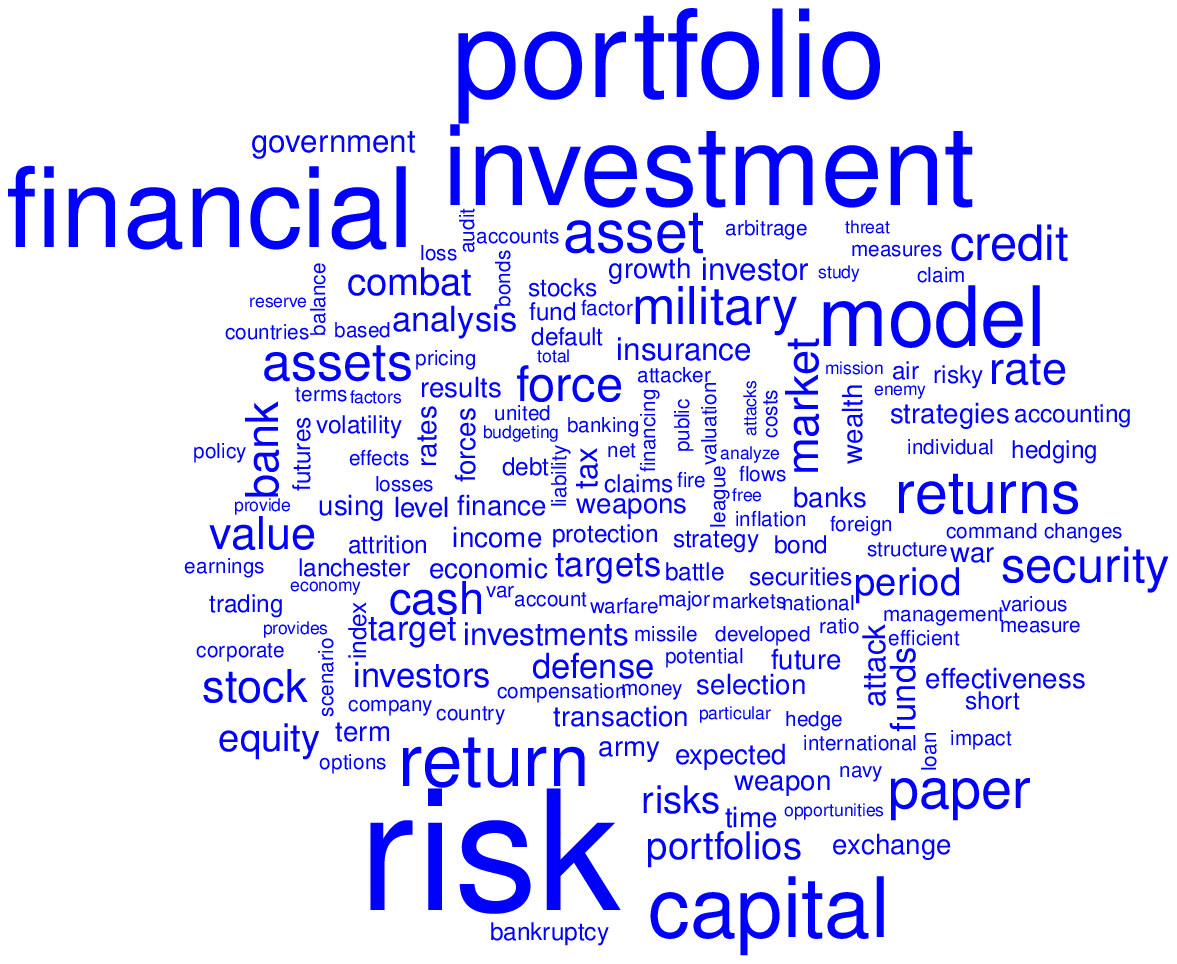} }
\subfigure[Topic \#16.]{\includegraphics[trim=1.2cm 1.2cm 1.2cm 1.2cm, clip=true, width=0.4\textwidth]{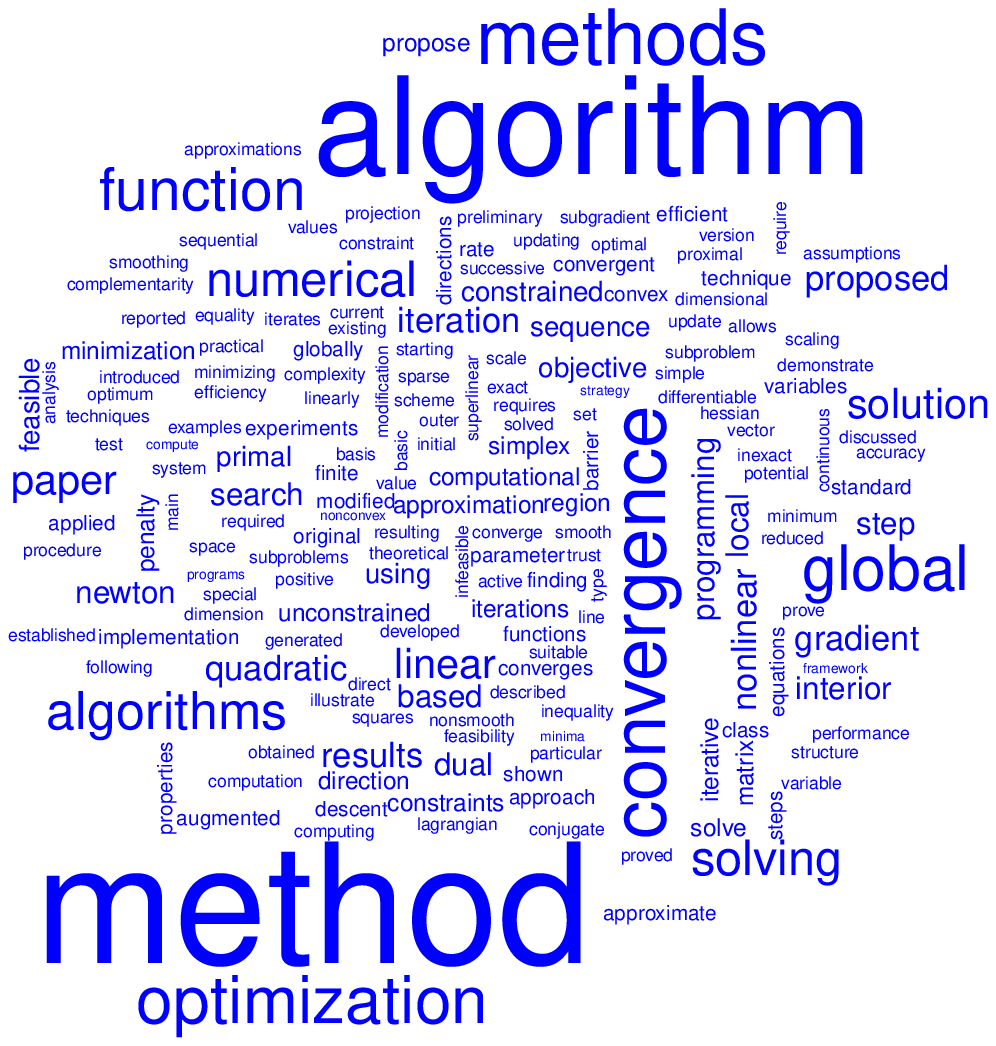} } \\
\subfigure[Topic \#17.]{\includegraphics[trim=1.2cm 1.2cm 1.2cm 1.2cm, clip=true, width=0.4\textwidth]{wordcloud_topic17.eps} }
\subfigure[Topic \#18.]{\includegraphics[trim=1.2cm 1.2cm 1.2cm 1.2cm, clip=true, width=0.4\textwidth]{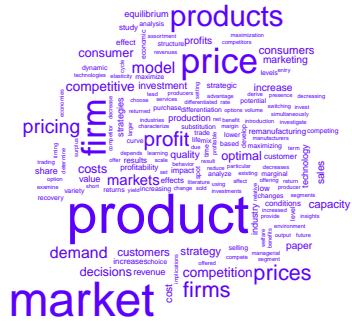} }
\caption{Wordclouds for topics 13--18.}
\label{fig:wordclouds3}
\end{figure}

\newpage

\begin{figure}[!h]
\centering
\subfigure[Topic \#19.]{\includegraphics[trim=1.2cm 1.2cm 1.2cm 1.2cm, clip=true, width=0.4\textwidth]{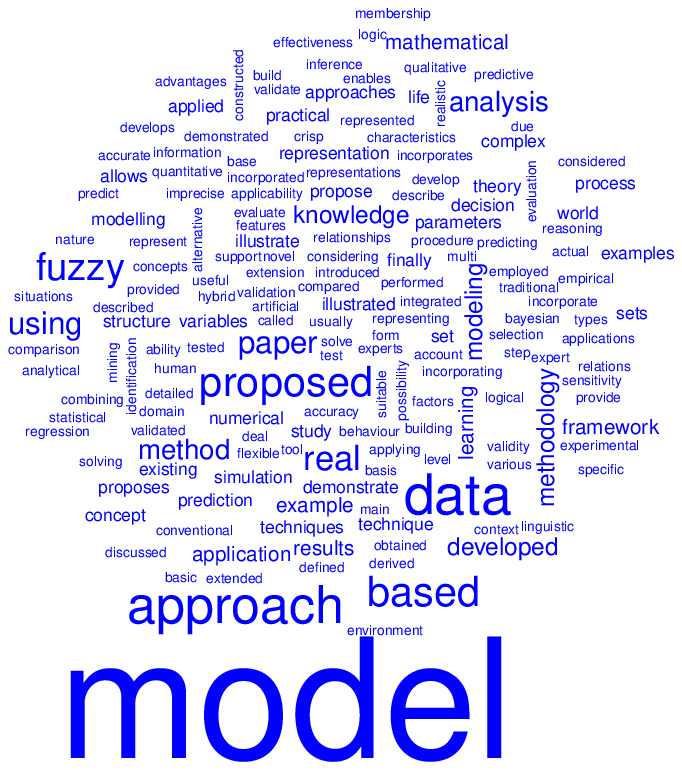} }
\subfigure[Topic \#20.]{\includegraphics[trim=1.2cm 1.2cm 1.2cm 1.2cm, clip=true, width=0.4\textwidth]{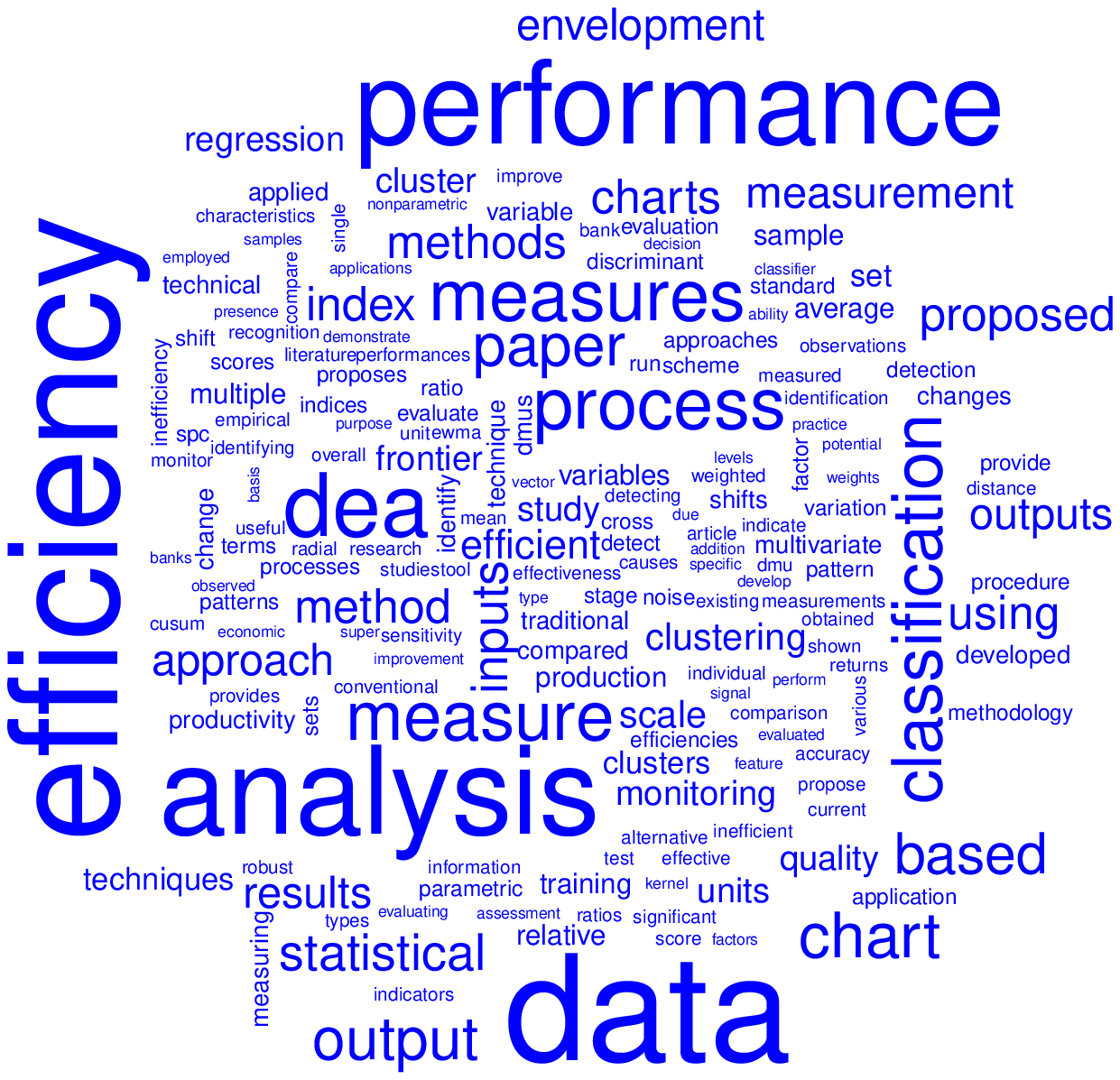} } \\
\subfigure[Topic \#21.]{\includegraphics[trim=1.2cm 1.2cm 1.2cm 1.2cm, clip=true, width=0.4\textwidth]{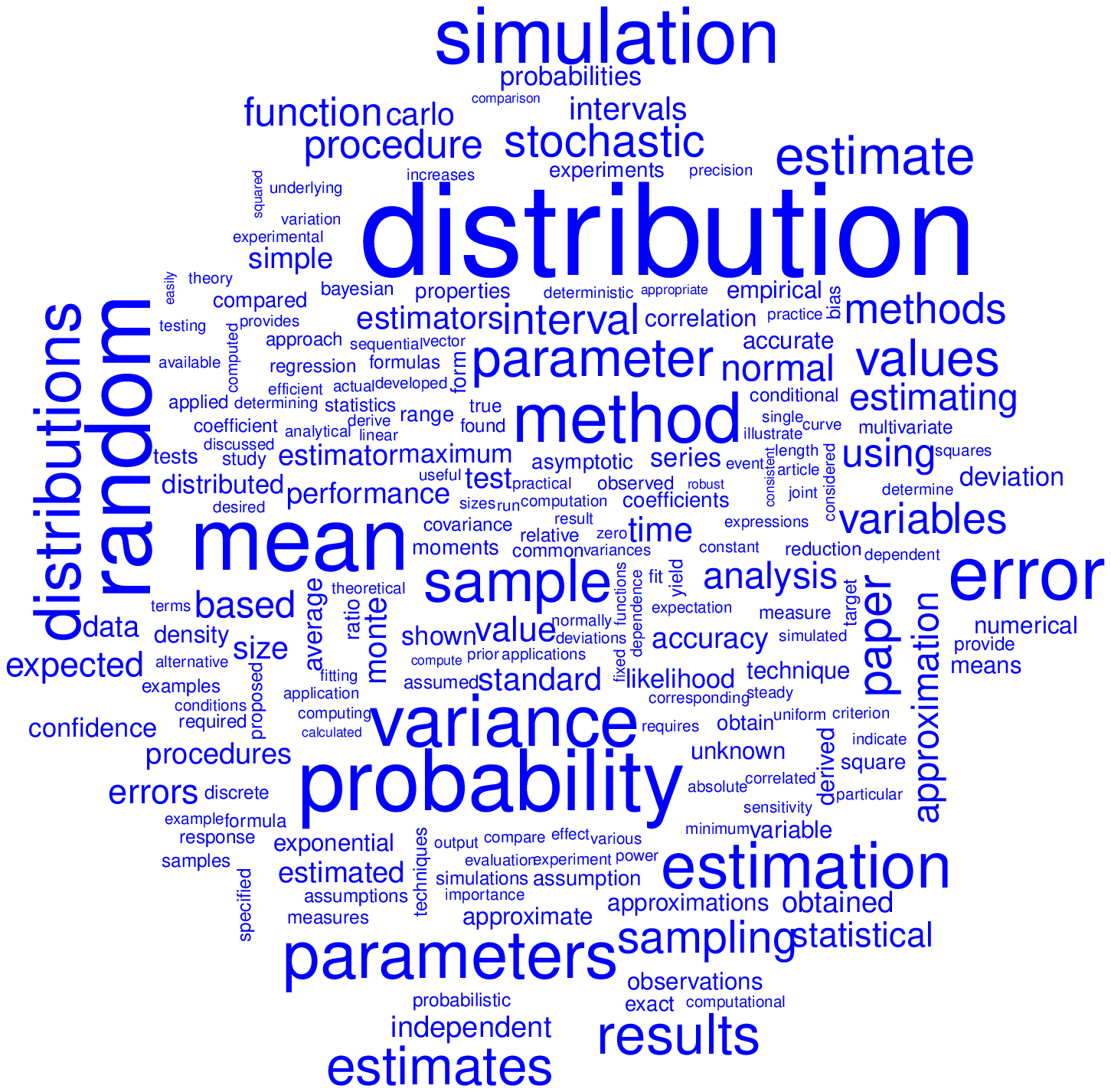} }
\subfigure[Topic \#22.]{\includegraphics[trim=1.2cm 1.2cm 1.2cm 1.2cm, clip=true, width=0.4\textwidth]{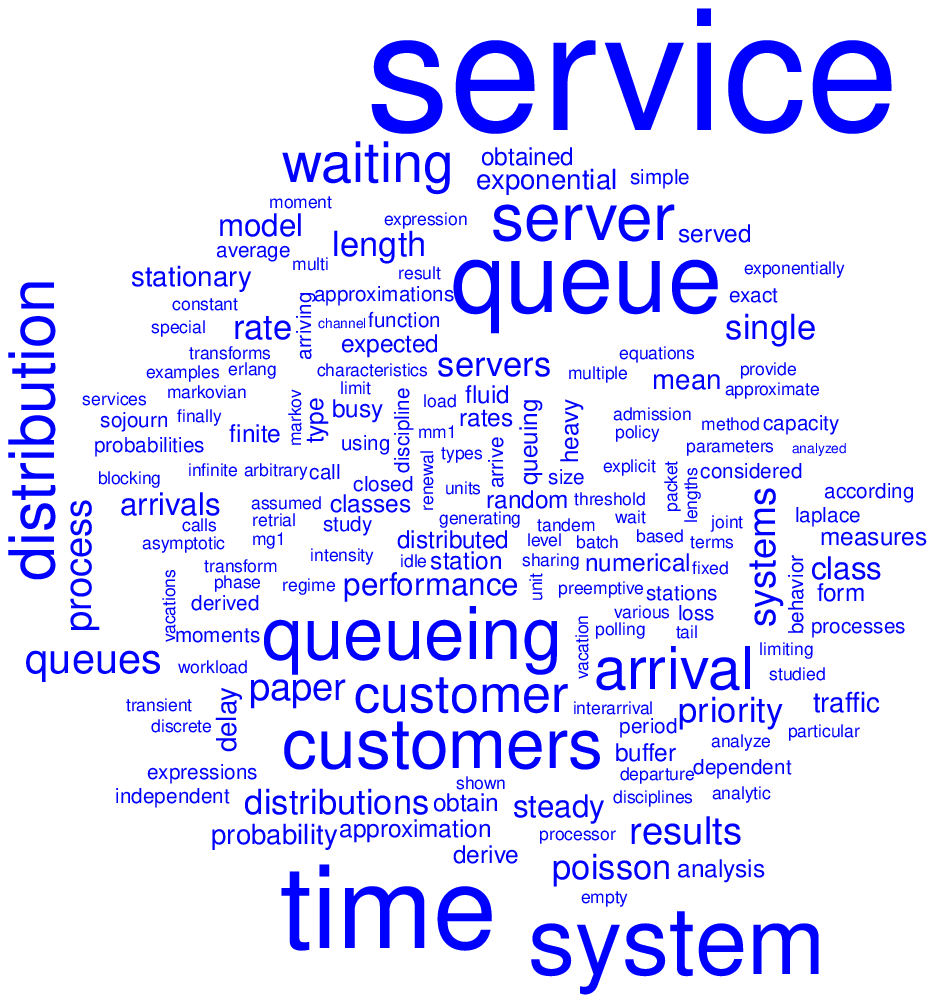} } \\
\subfigure[Topic \#23.]{\includegraphics[trim=1.2cm 1.2cm 1.2cm 1.2cm, clip=true, width=0.4\textwidth]{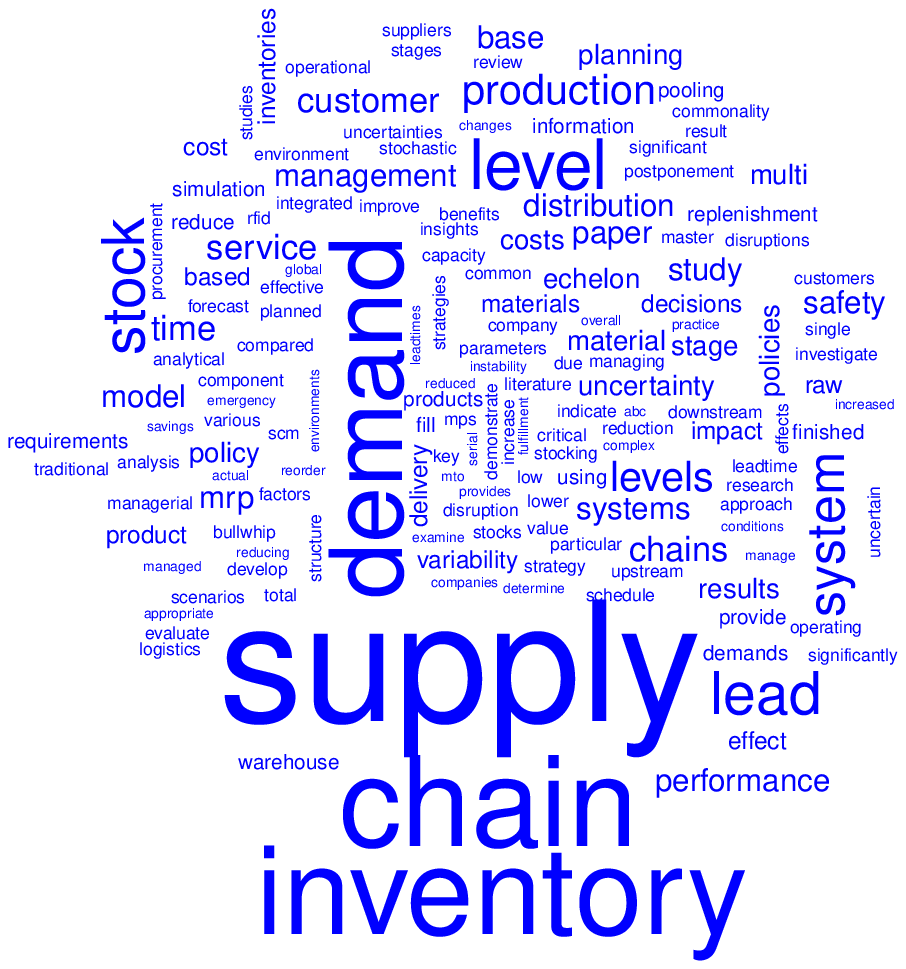} }
\subfigure[Topic \#24.]{\includegraphics[trim=1.2cm 1.2cm 1.2cm 1.2cm, clip=true, width=0.4\textwidth]{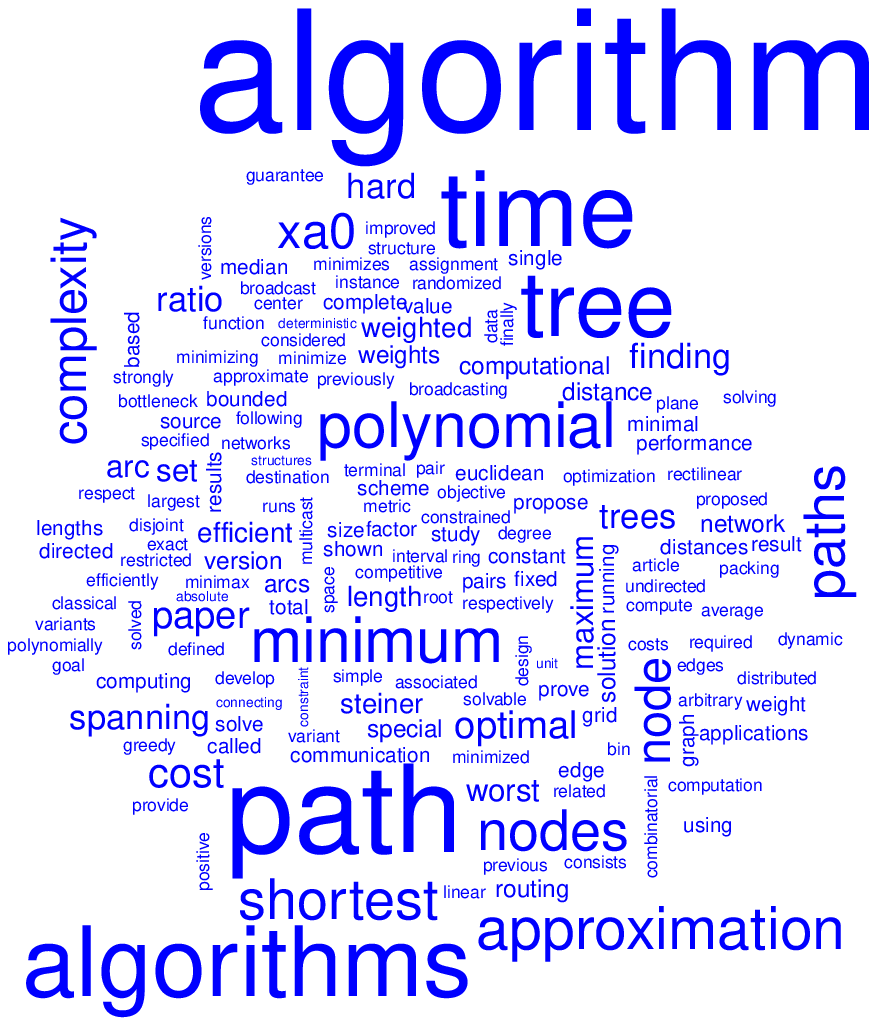} }
\caption{Wordclouds for topics 19--24.}
\label{fig:wordclouds4}
\end{figure}

\newpage

\begin{figure}[!h]
\centering
\subfigure[Topic \#25.]{\includegraphics[trim=1.2cm 1.2cm 1.2cm 1.2cm, clip=true, width=0.4\textwidth]{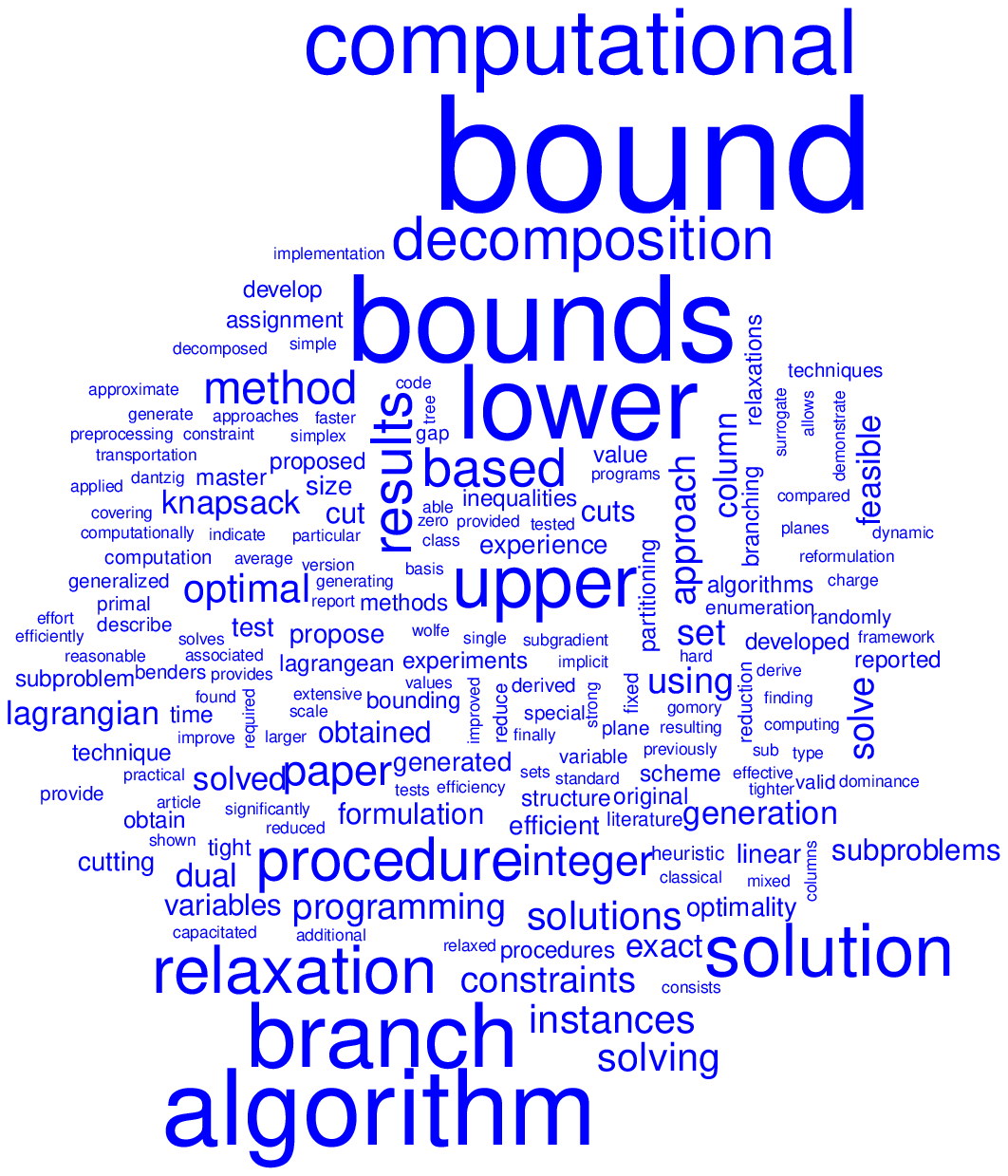} }
\subfigure[Topic \#26.]{\includegraphics[trim=1.2cm 1.2cm 1.2cm 1.2cm, clip=true, width=0.4\textwidth]{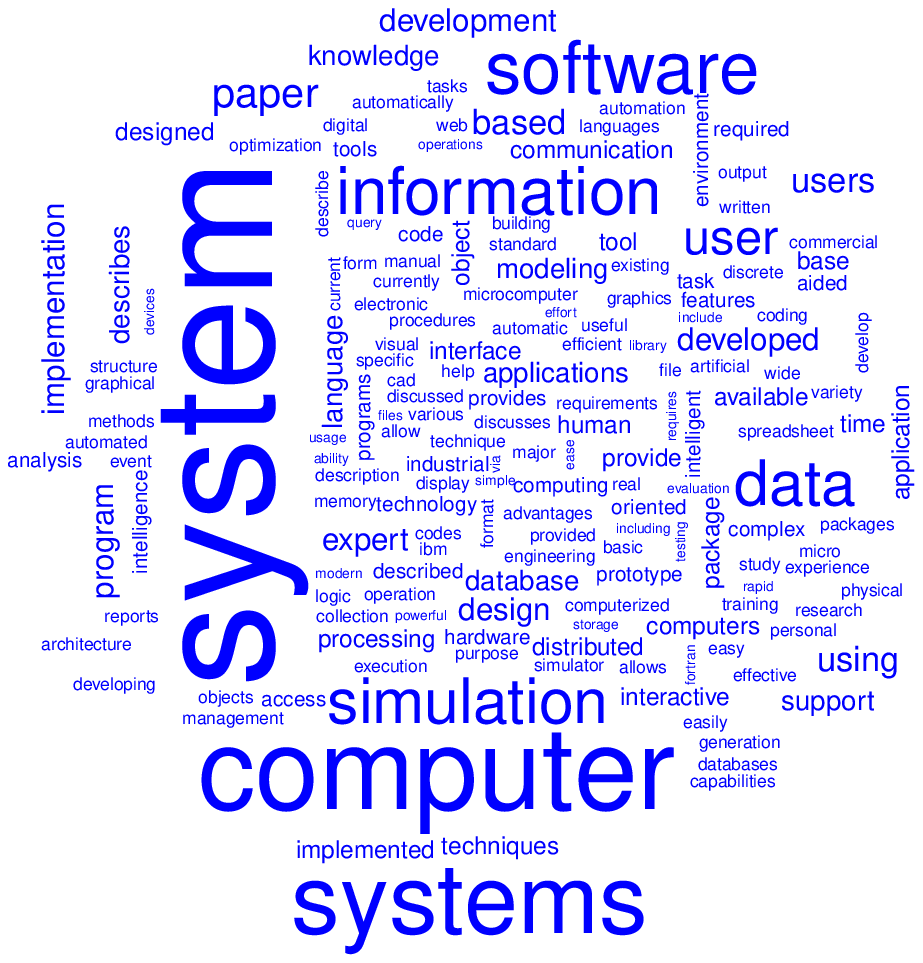} } \\
\subfigure[Topic \#27.]{\includegraphics[trim=1.2cm 1.2cm 1.2cm 1.2cm, clip=true, width=0.4\textwidth]{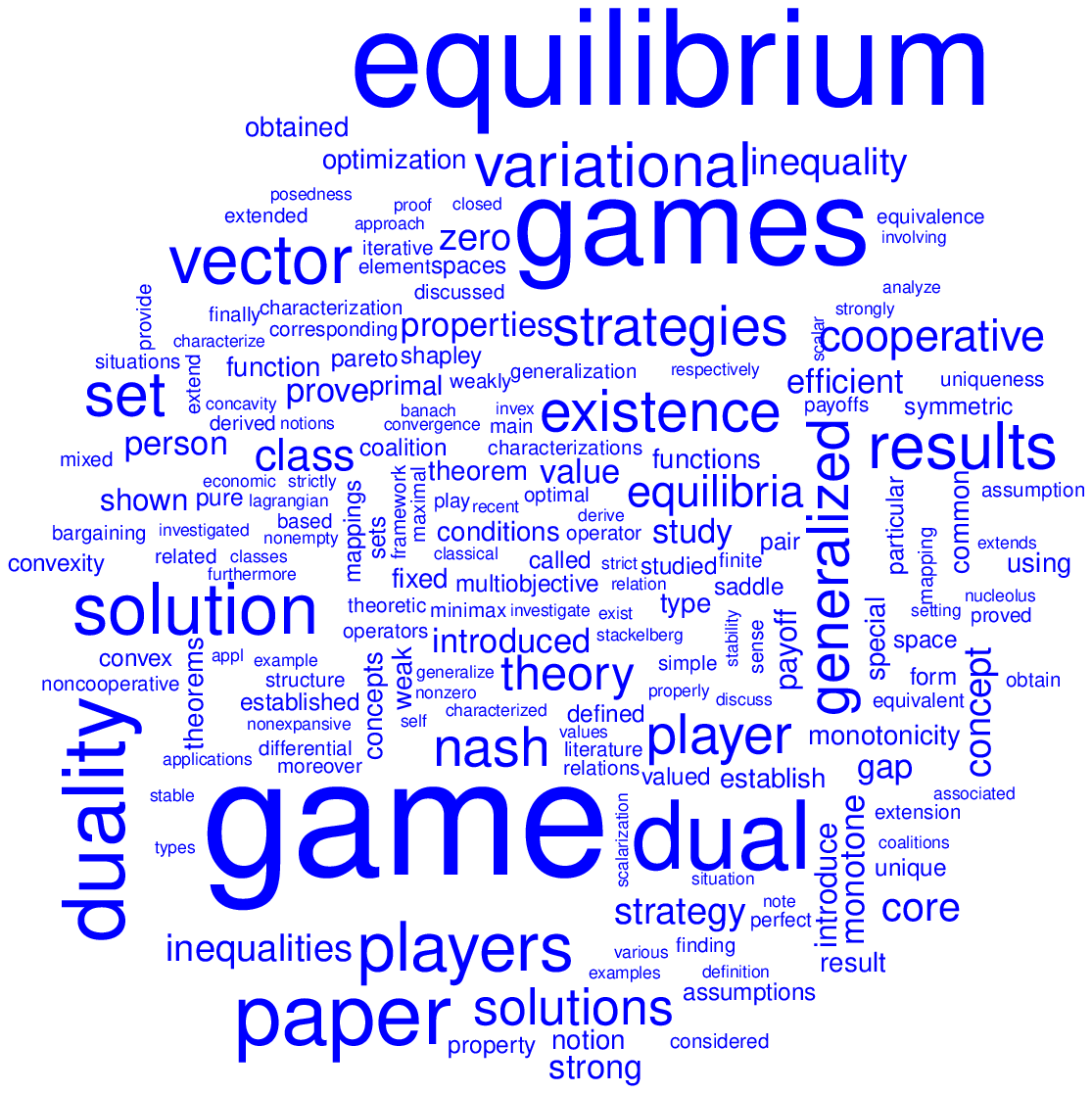} }
\subfigure[Topic \#28.]{\includegraphics[trim=1.2cm 1.2cm 1.2cm 1.2cm, clip=true, width=0.4\textwidth]{wordcloud_topic28.eps} } \\
\subfigure[Topic \#29.]{\includegraphics[trim=1.2cm 1.2cm 1.2cm 1.2cm, clip=true, width=0.4\textwidth]{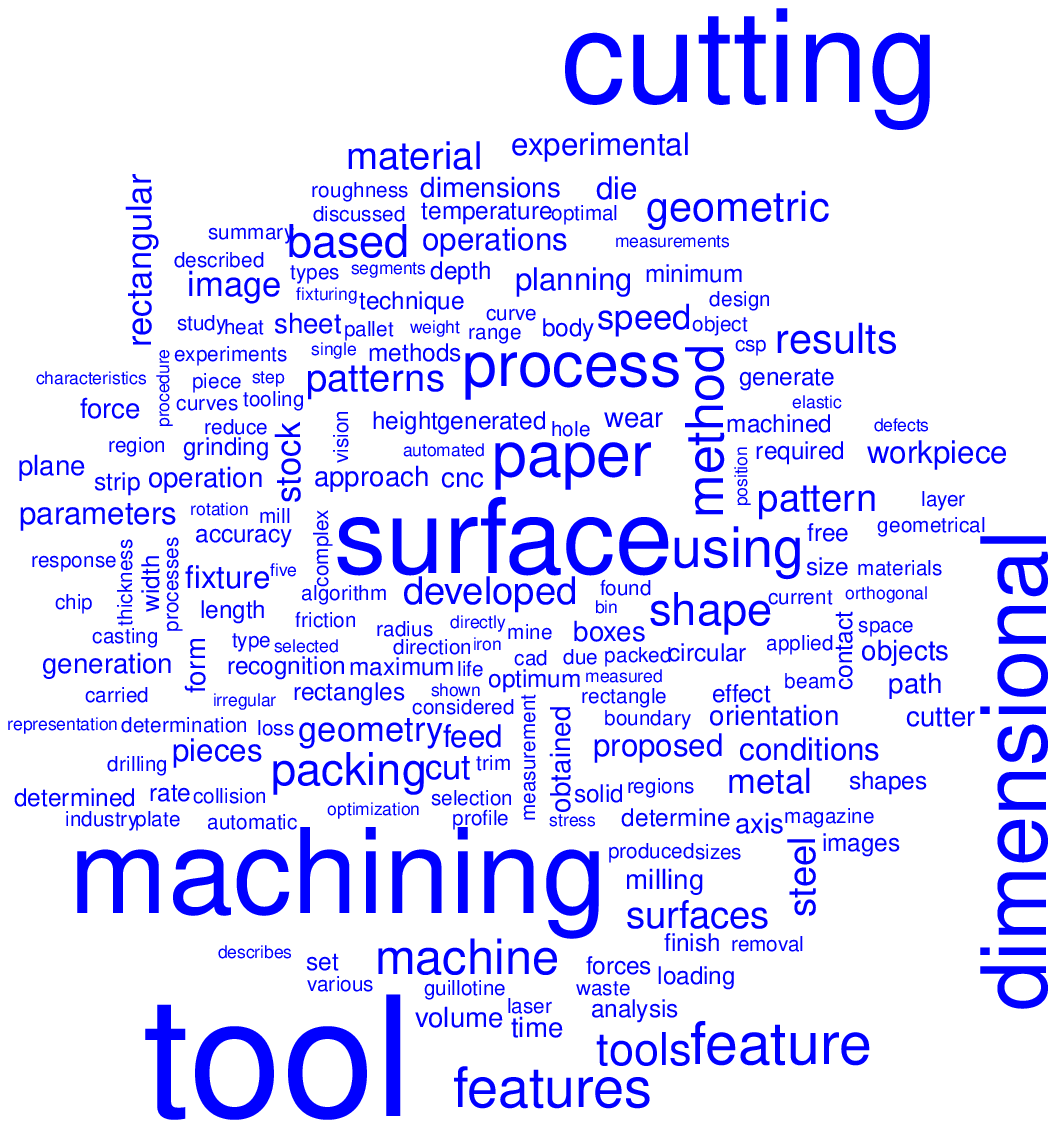} }
\subfigure[Topic \#30.]{\includegraphics[trim=1.2cm 1.2cm 1.2cm 1.2cm, clip=true, width=0.4\textwidth]{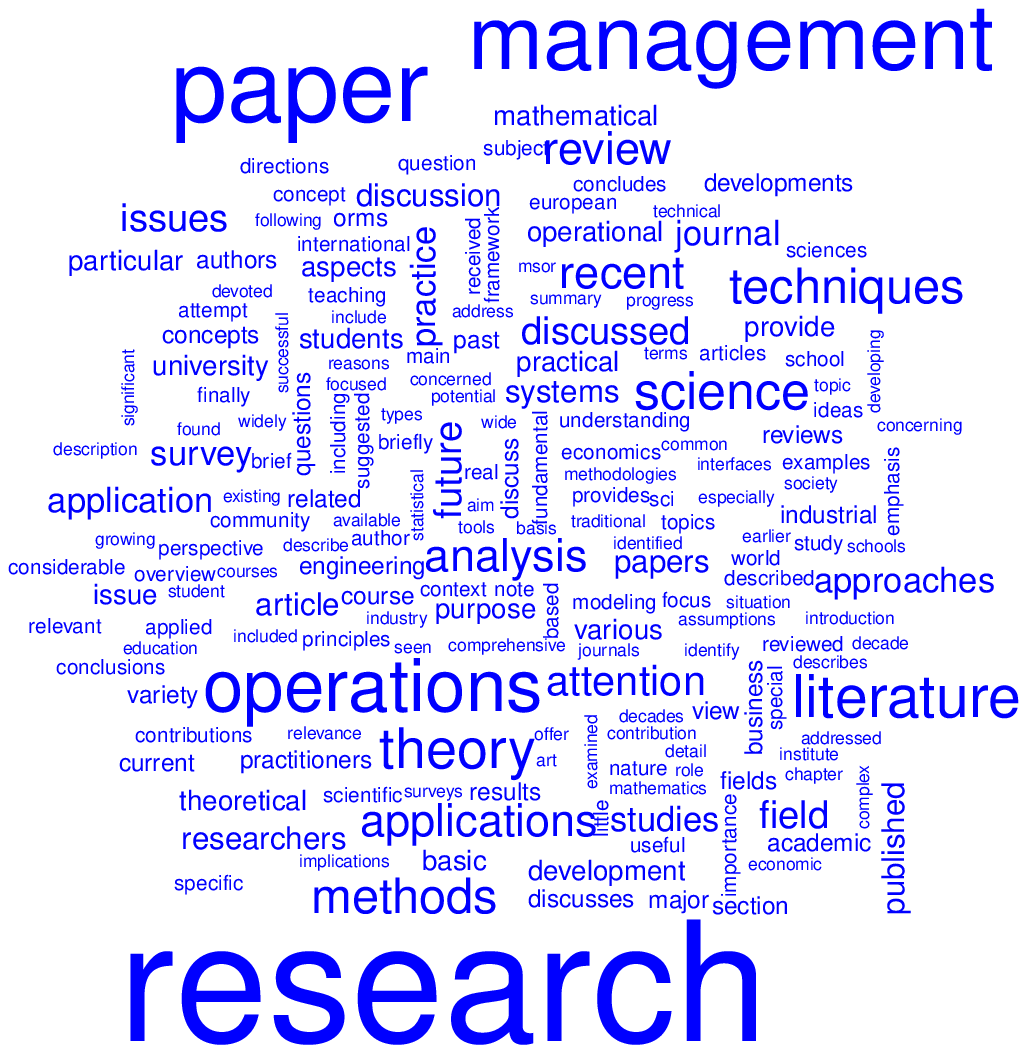} }
\caption{Wordclouds for topics 25--30.}
\label{fig:wordclouds5}
\end{figure}

\newpage

\begin{figure}[!h]
\centering
\subfigure[Topic \#31.]{\includegraphics[trim=1.2cm 1.2cm 1.2cm 1.2cm, clip=true, width=0.4\textwidth]{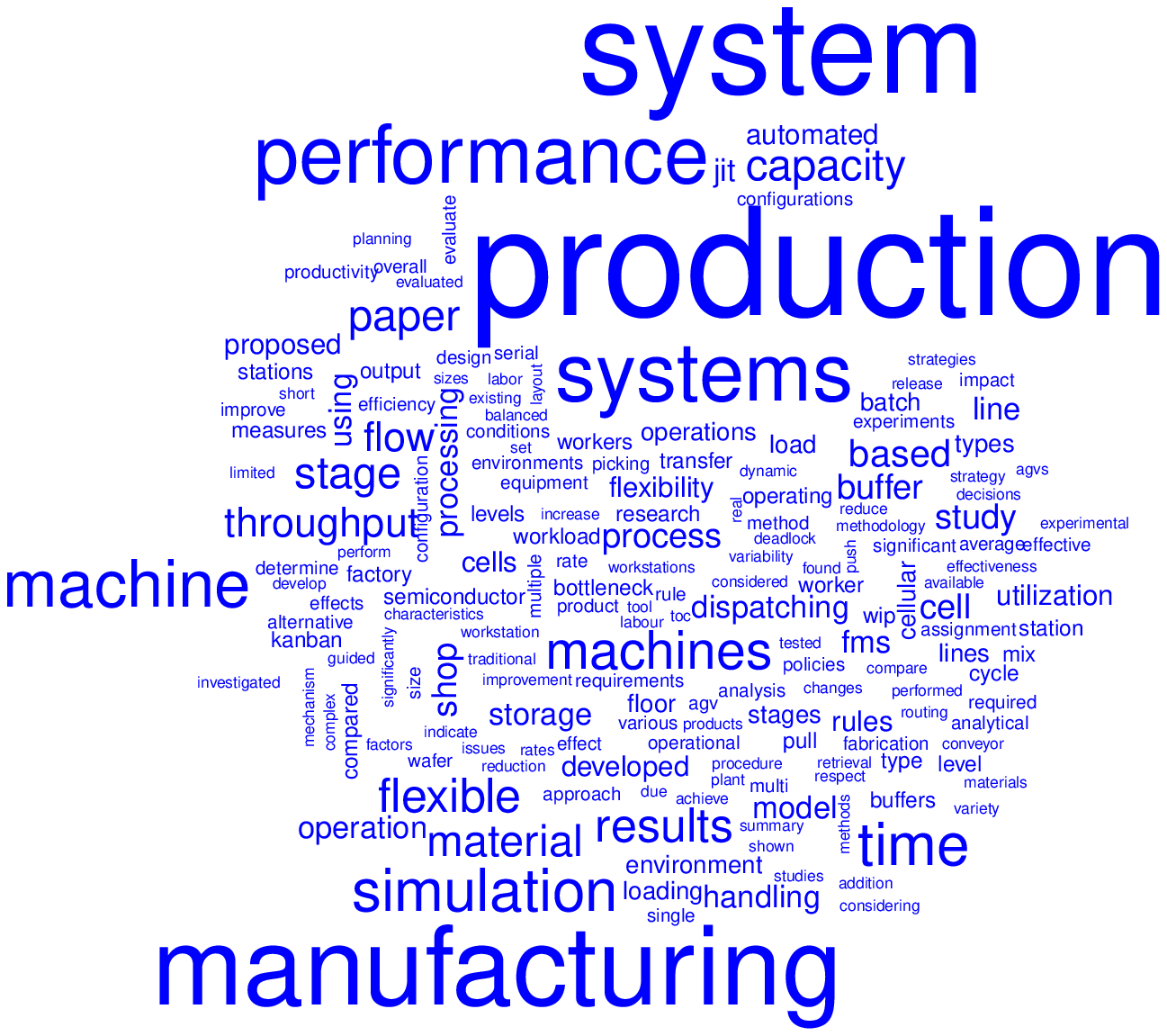} }
\subfigure[Topic \#32.]{\includegraphics[trim=1.2cm 1.2cm 1.2cm 1.2cm, clip=true, width=0.4\textwidth]{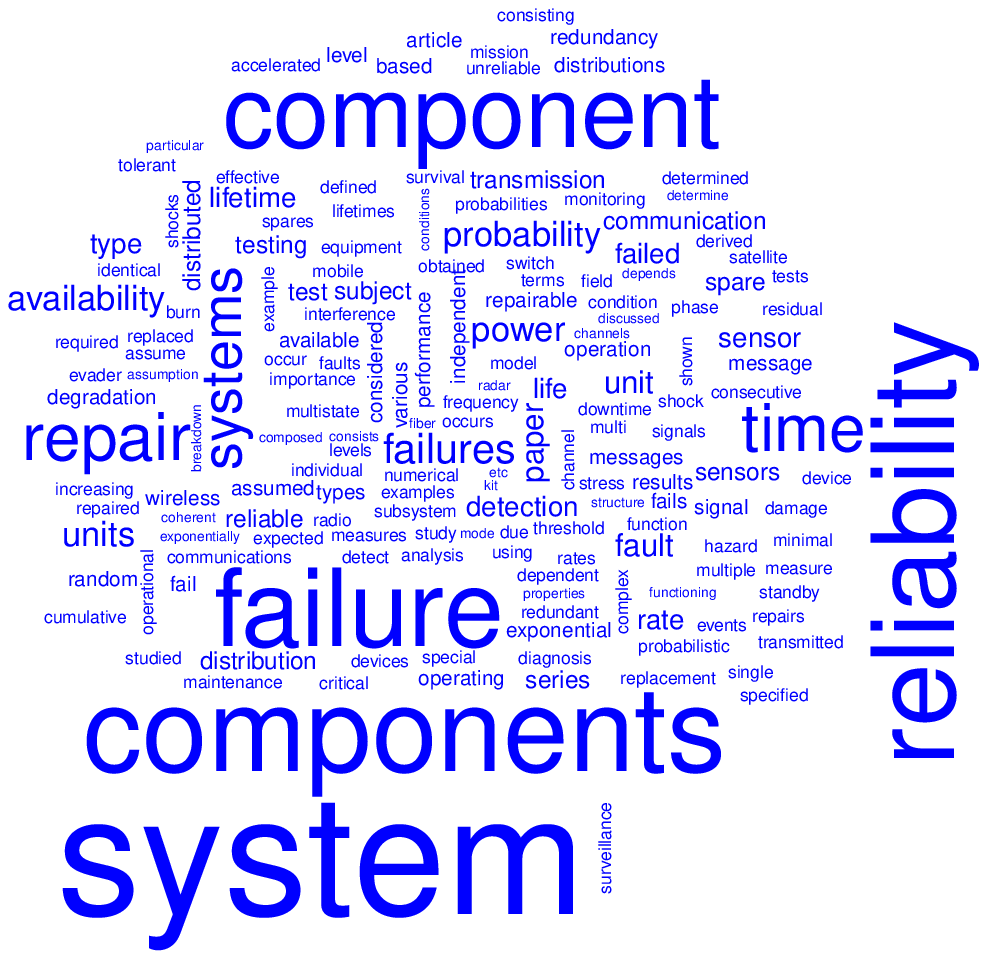} } \\
\subfigure[Topic \#33.]{\includegraphics[trim=1.2cm 1.2cm 1.2cm 1.2cm, clip=true, width=0.4\textwidth]{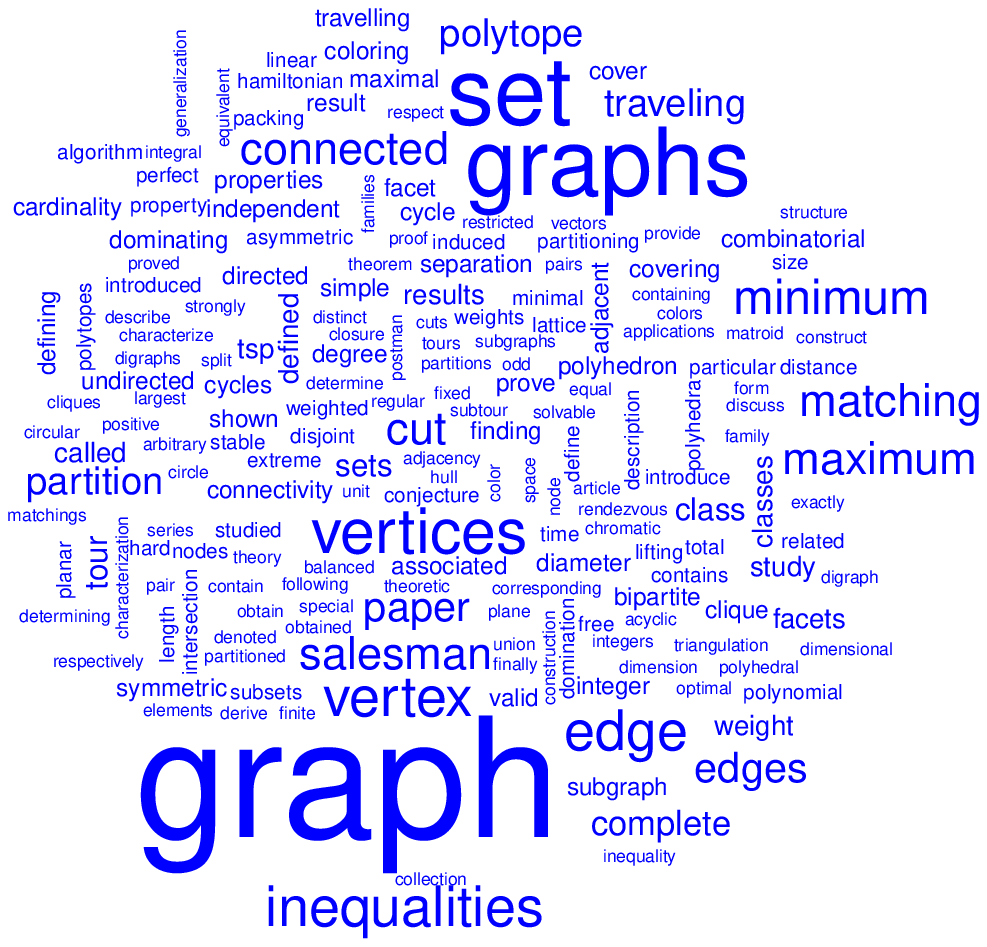} }
\subfigure[Topic \#34.]{\includegraphics[trim=1.2cm 1.2cm 1.2cm 1.2cm, clip=true, width=0.4\textwidth]{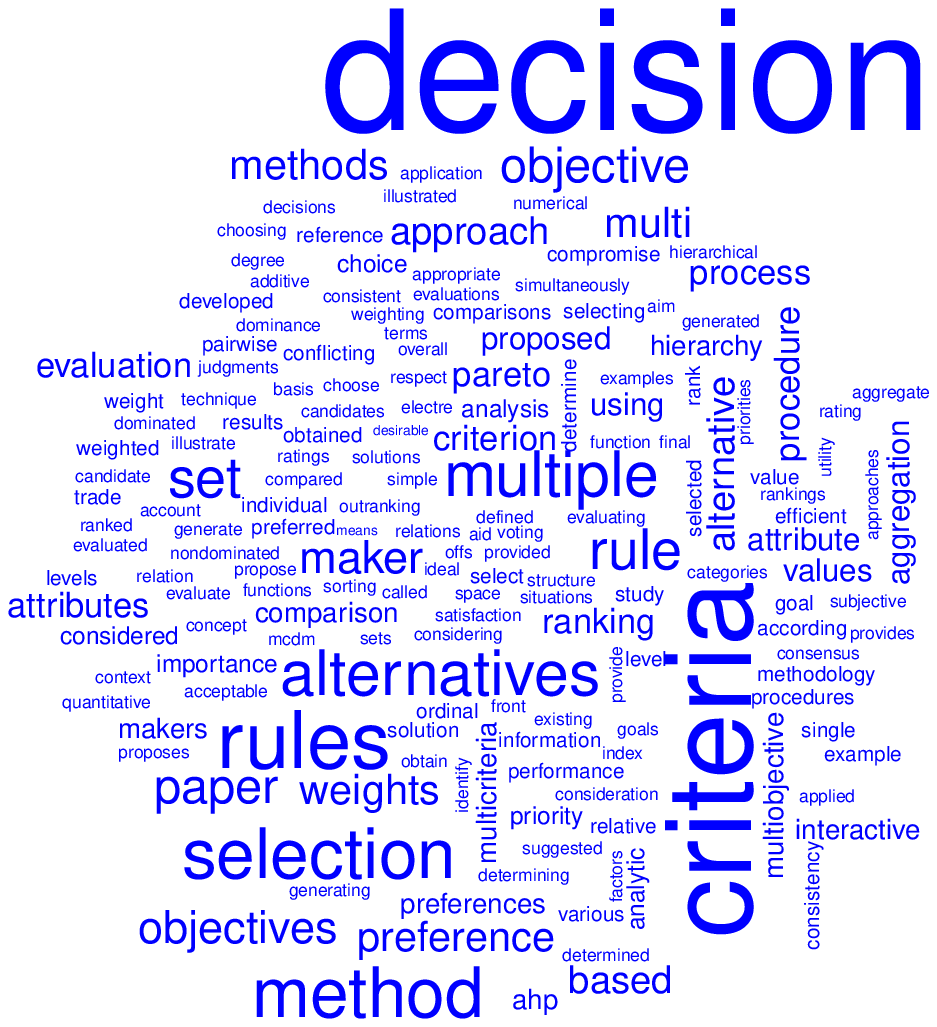} } \\
\subfigure[Topic \#35.]{\includegraphics[trim=1.2cm 1.2cm 1.2cm 1.2cm, clip=true, width=0.4\textwidth]{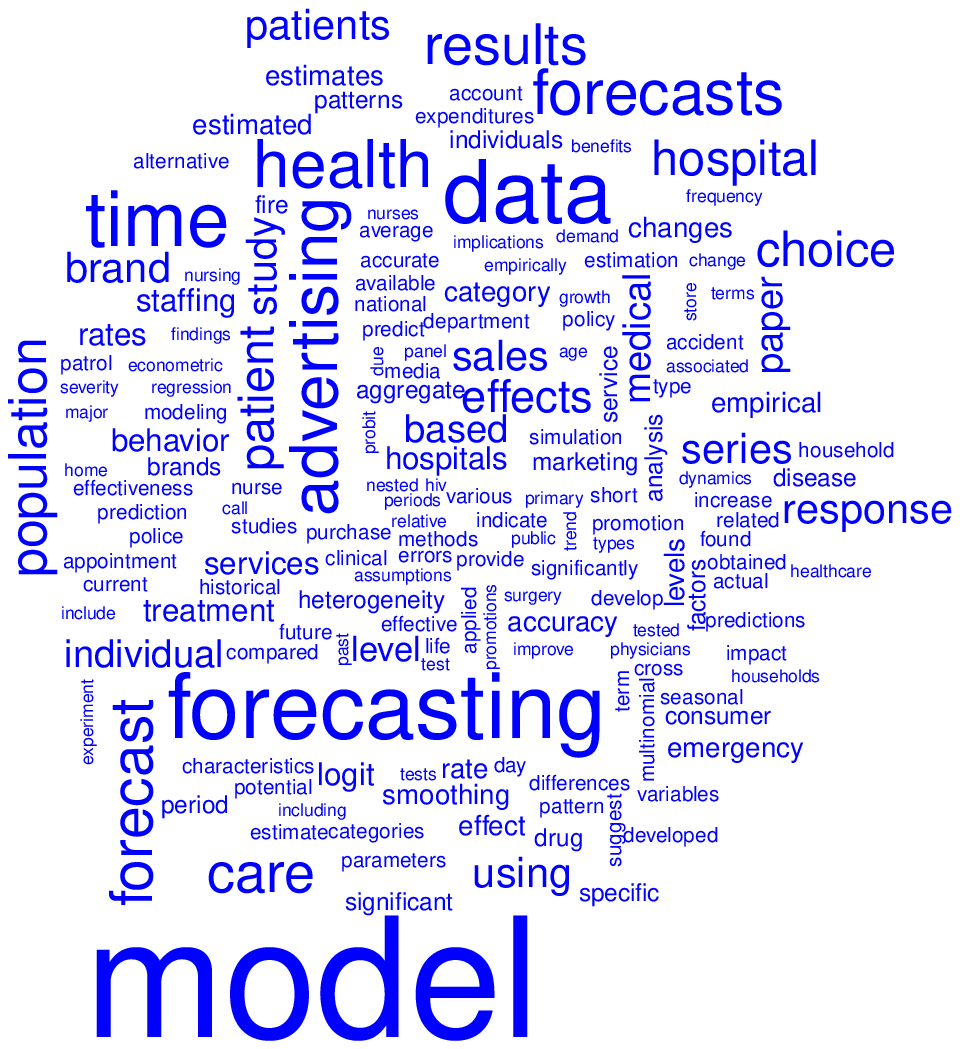} }
\subfigure[Topic \#36.]{\includegraphics[trim=1.2cm 1.2cm 1.2cm 1.2cm, clip=true, width=0.4\textwidth]{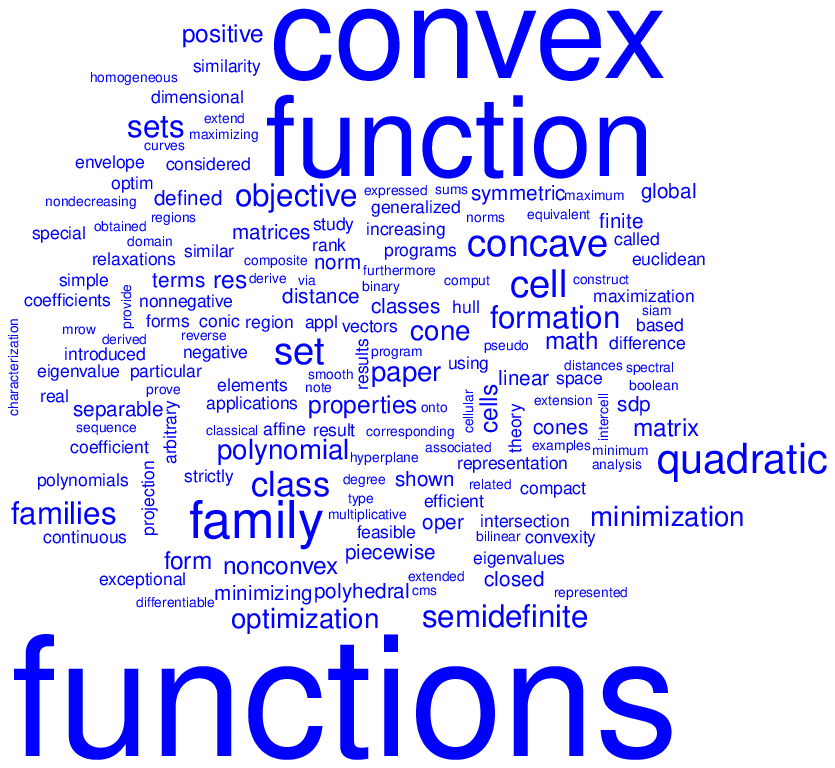} }
\caption{Wordclouds for topics 31--36.}
\label{fig:wordclouds6}
\end{figure}

\newpage

\begin{figure}[!h]
\centering
\subfigure[Topic \#37.]{\includegraphics[trim=1.2cm 1.2cm 1.2cm 1.2cm, clip=true, width=0.4\textwidth]{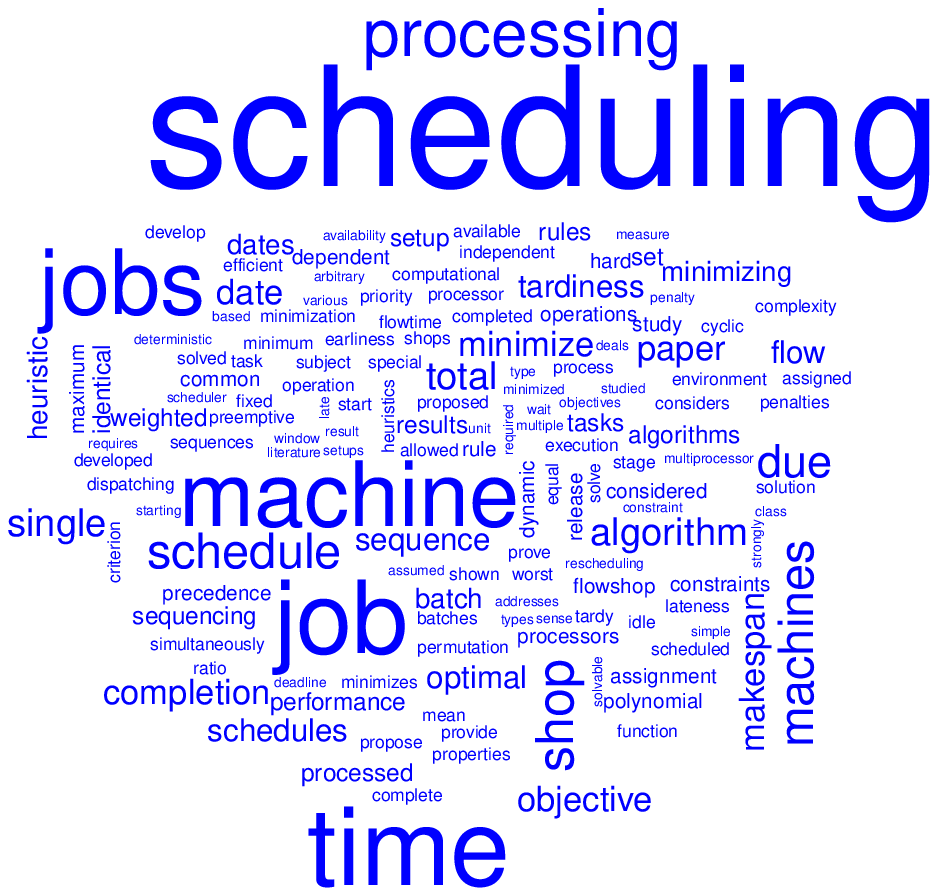} }
\subfigure[Topic \#38.]{\includegraphics[trim=1.2cm 1.2cm 1.2cm 1.2cm, clip=true, width=0.4\textwidth]{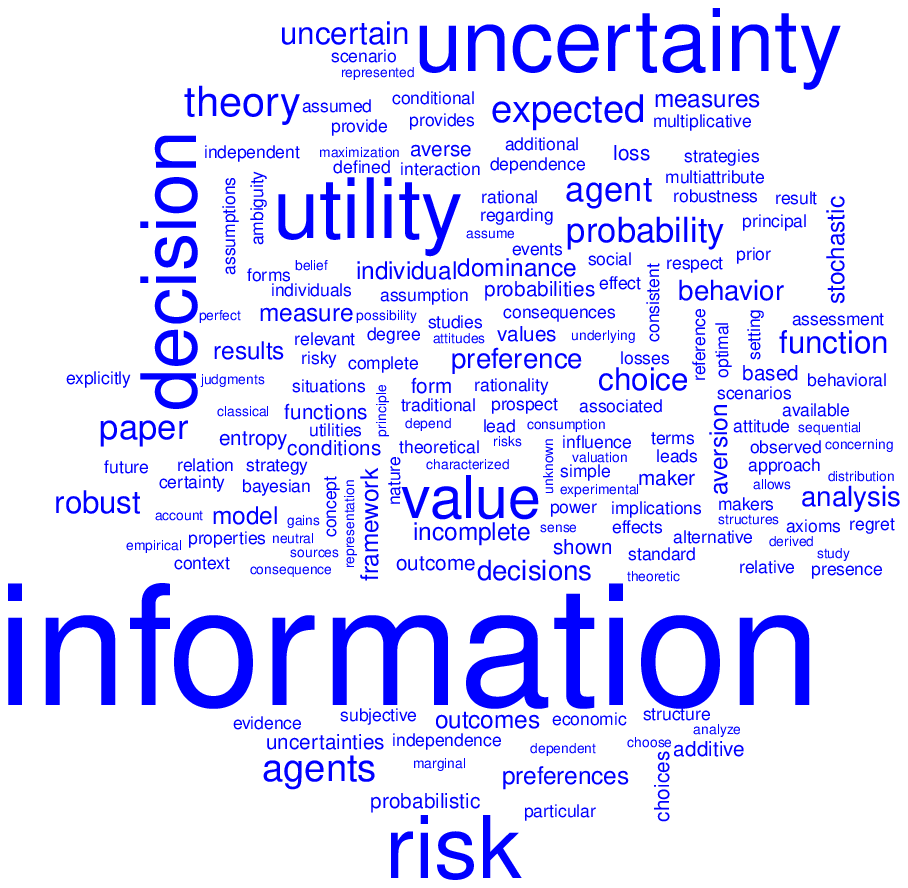} } \\
\subfigure[Topic \#39.]{\includegraphics[trim=1.2cm 1.2cm 1.2cm 1.2cm, clip=true, width=0.4\textwidth]{wordcloud_topic39.eps} }
\subfigure[Topic \#40.]{\includegraphics[trim=1.2cm 1.2cm 1.2cm 1.2cm, clip=true, width=0.4\textwidth]{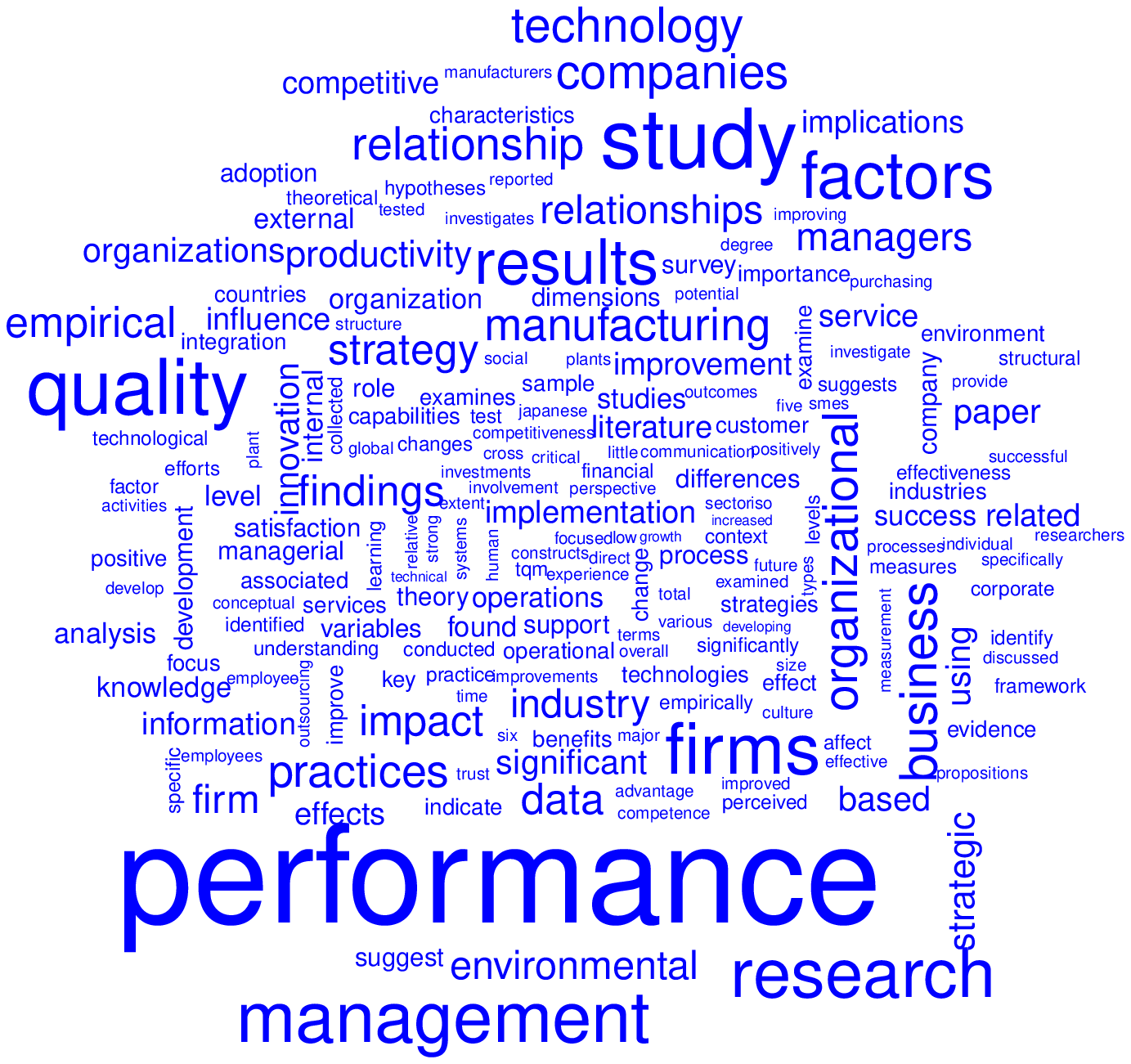} }
\caption{Wordclouds for topics 37--40.}
\label{fig:wordclouds7}
\end{figure}


\clearpage

\section*{Appendix B: Temporal topic distributions}

\begin{figure}[!ht]
\centering
\subfigure[Annals of Operations Research.]{\includegraphics[width=0.48\textwidth]{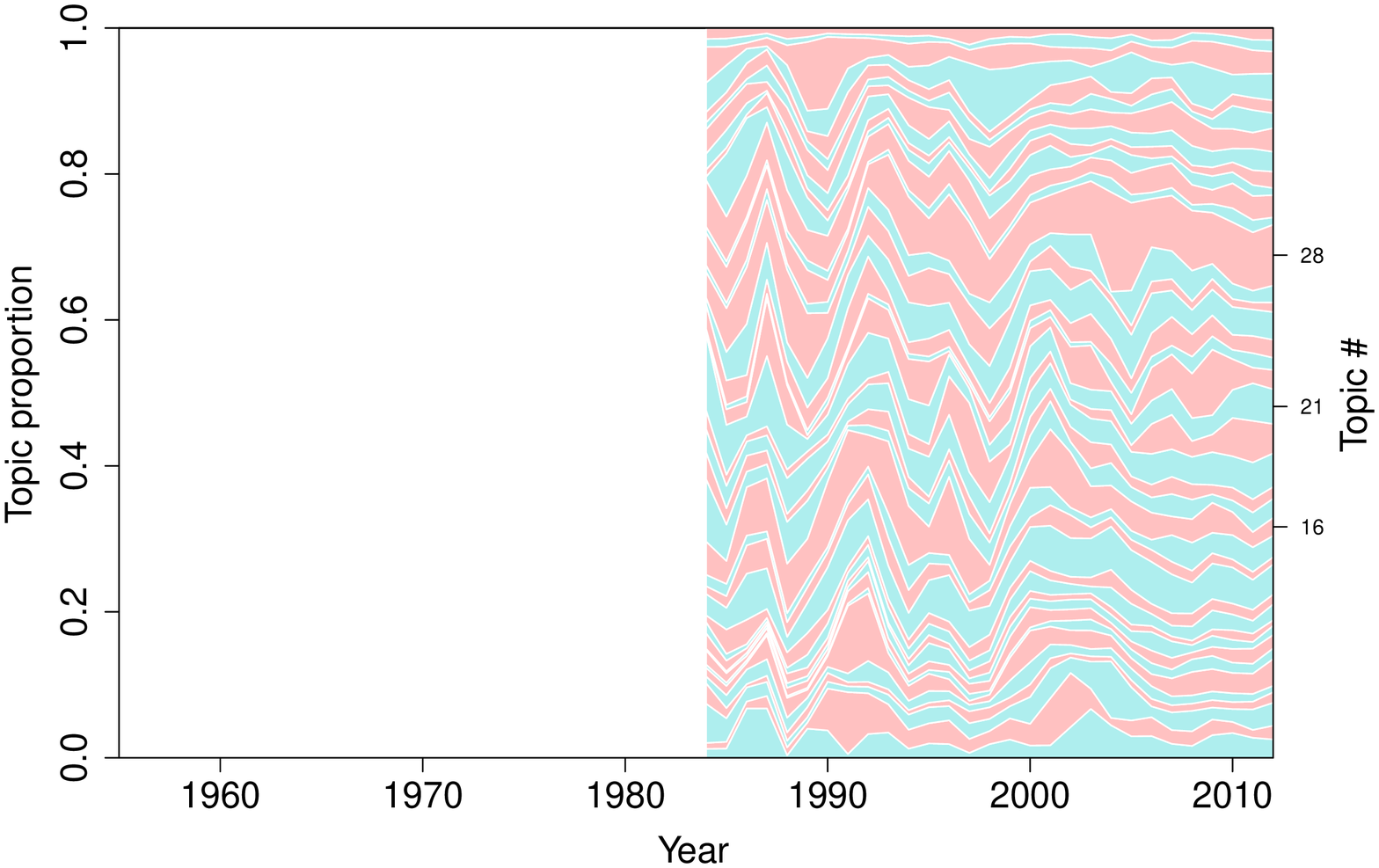} }
\subfigure[European Journal of Operational Research.]{\includegraphics[width=0.48\textwidth]{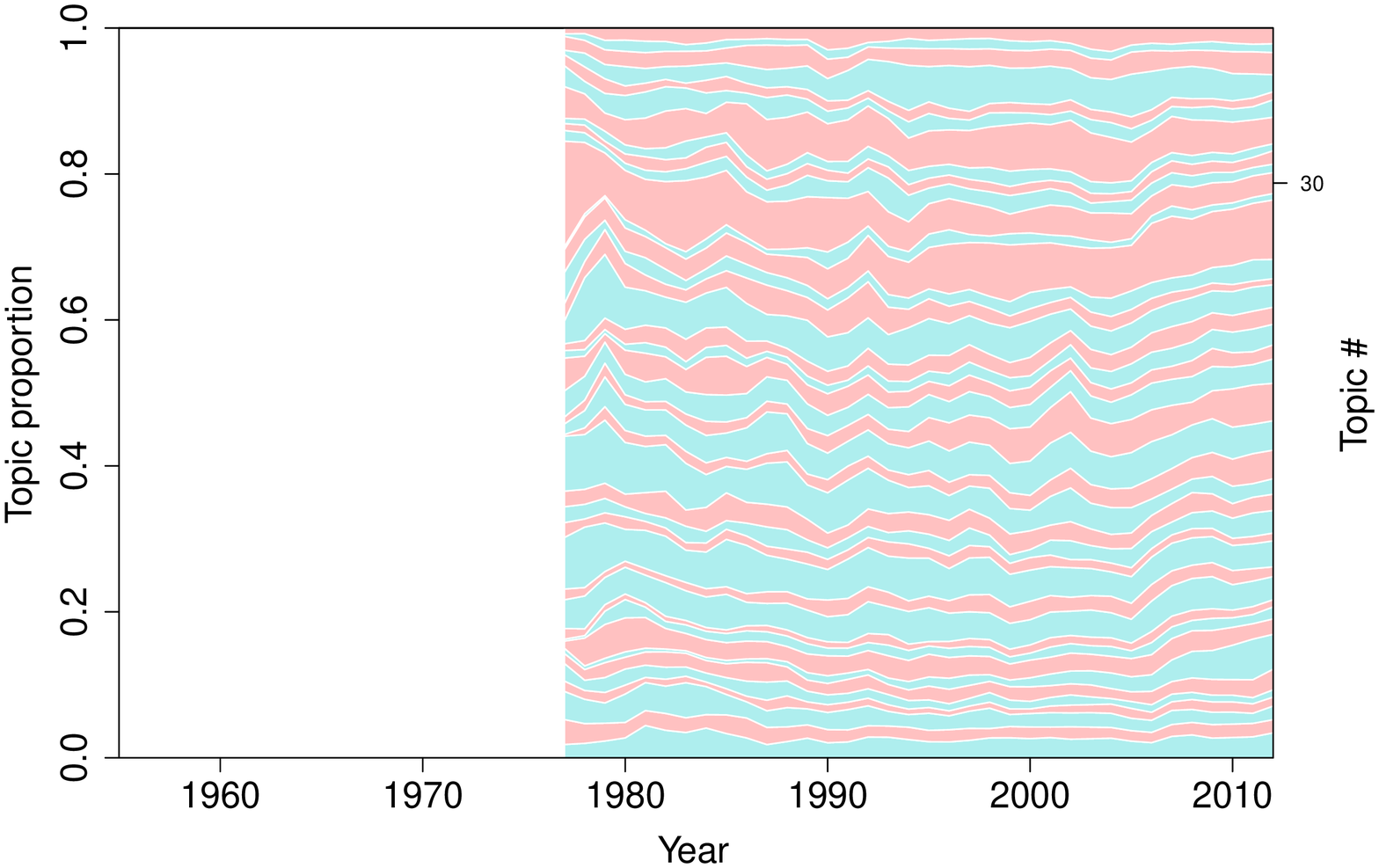} }
\subfigure[IIE Transactions.]{\includegraphics[width=0.48\textwidth]{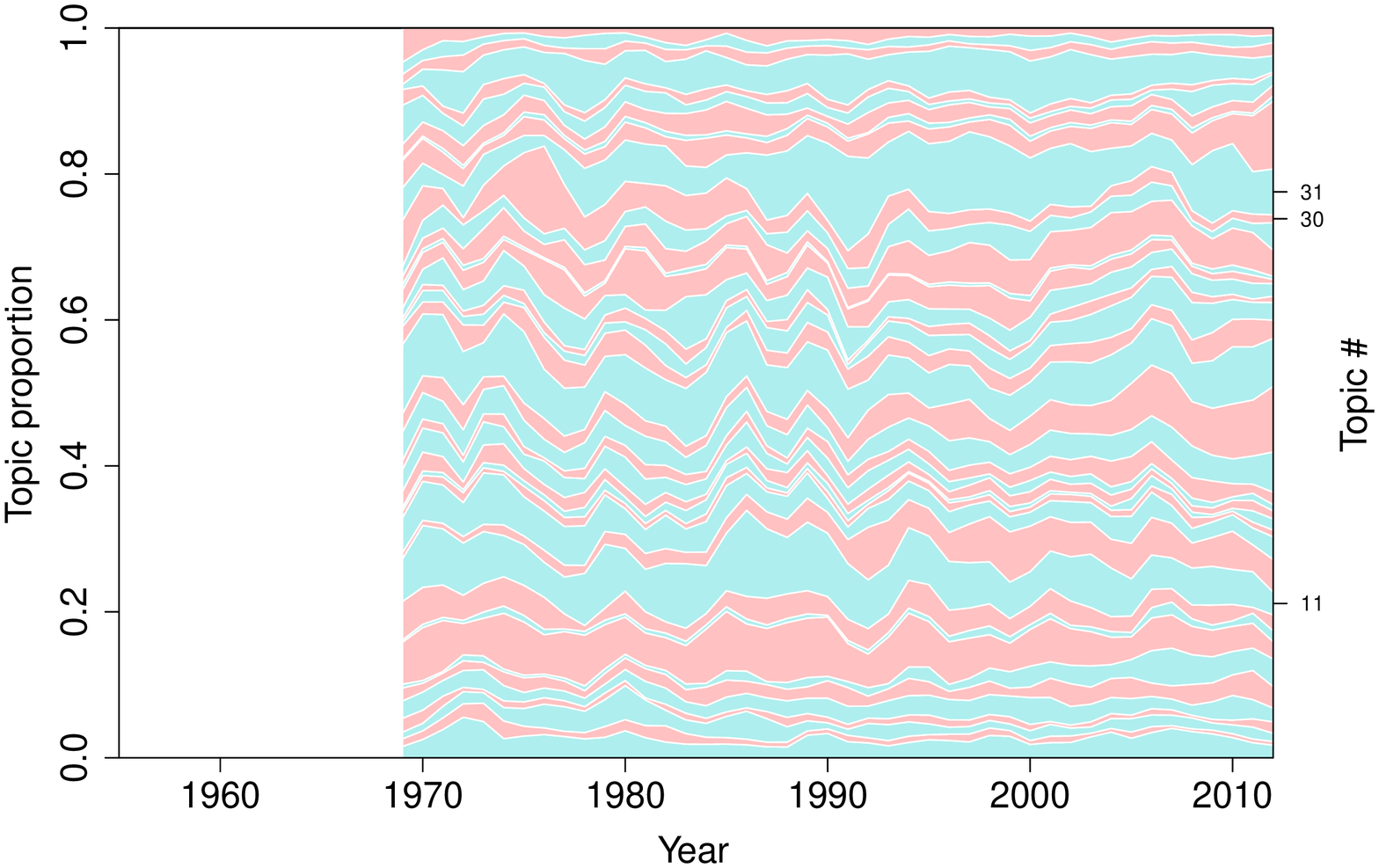} }
\subfigure[International Journal of Production Economics.]{\includegraphics[width=0.48\textwidth]{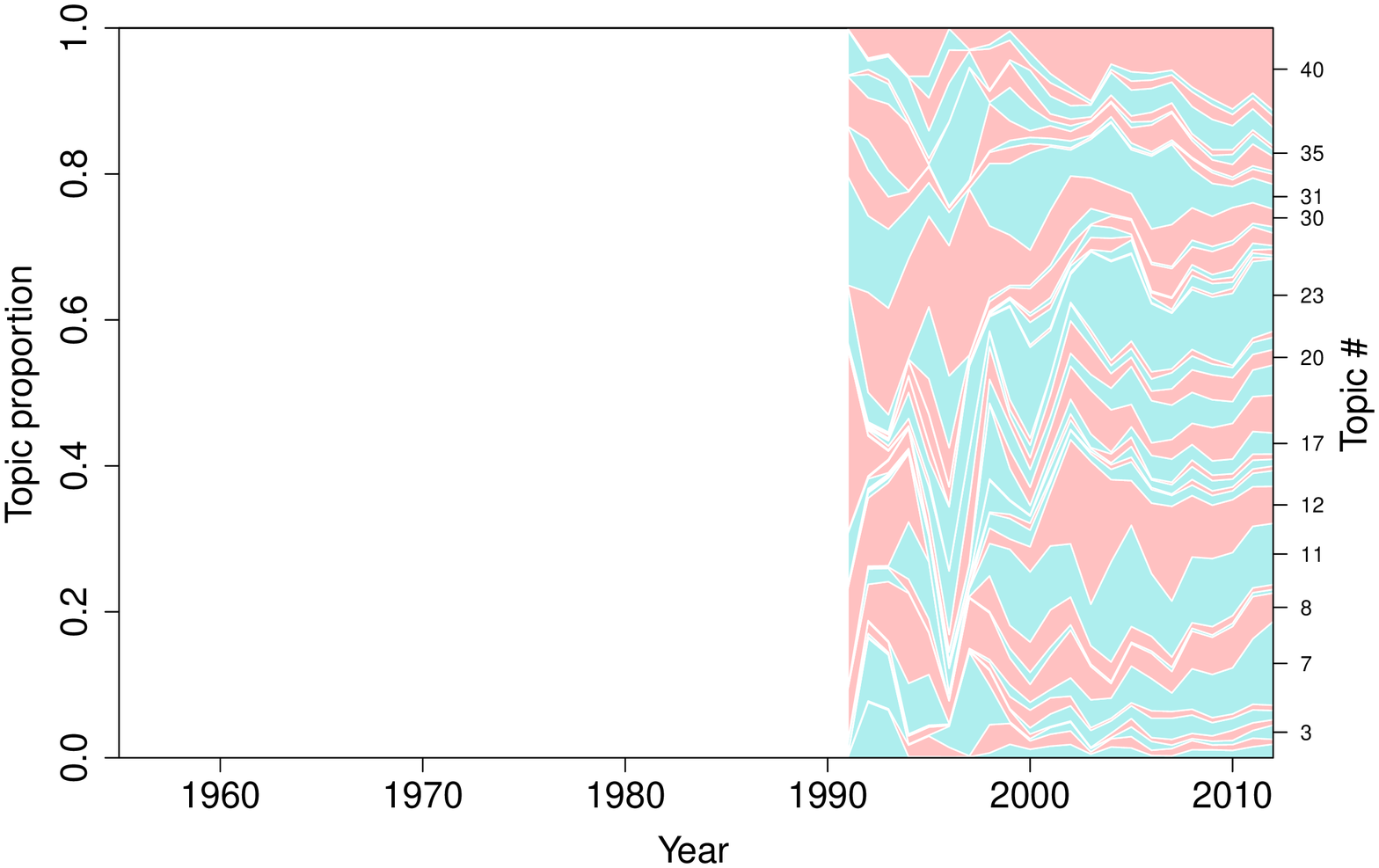} }
\subfigure[Journal of Combinatorial Optimization.]{\includegraphics[width=0.48\textwidth]{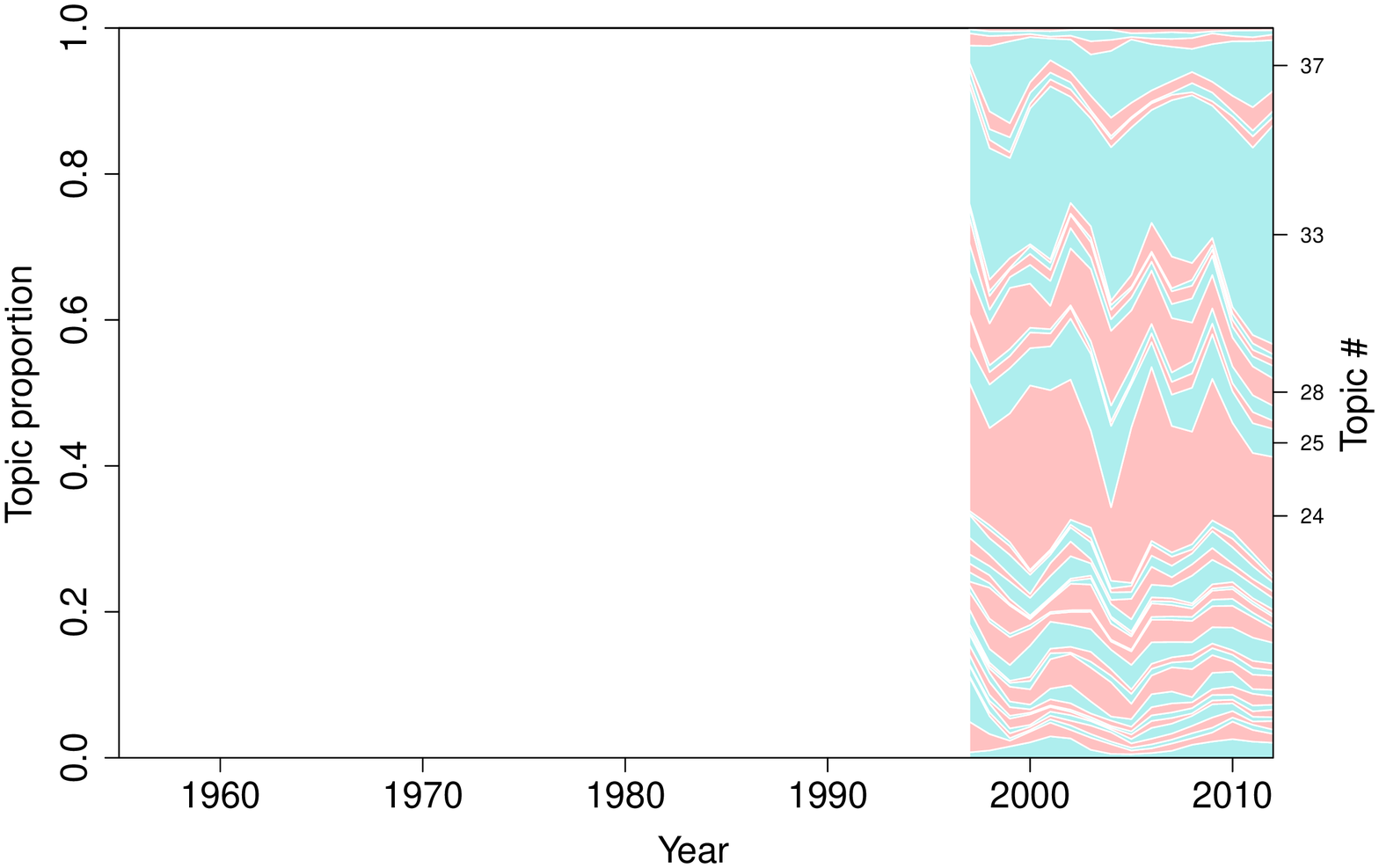} }
\subfigure[Journal of Optimization Theory and Applications.]{\includegraphics[width=0.48\textwidth]{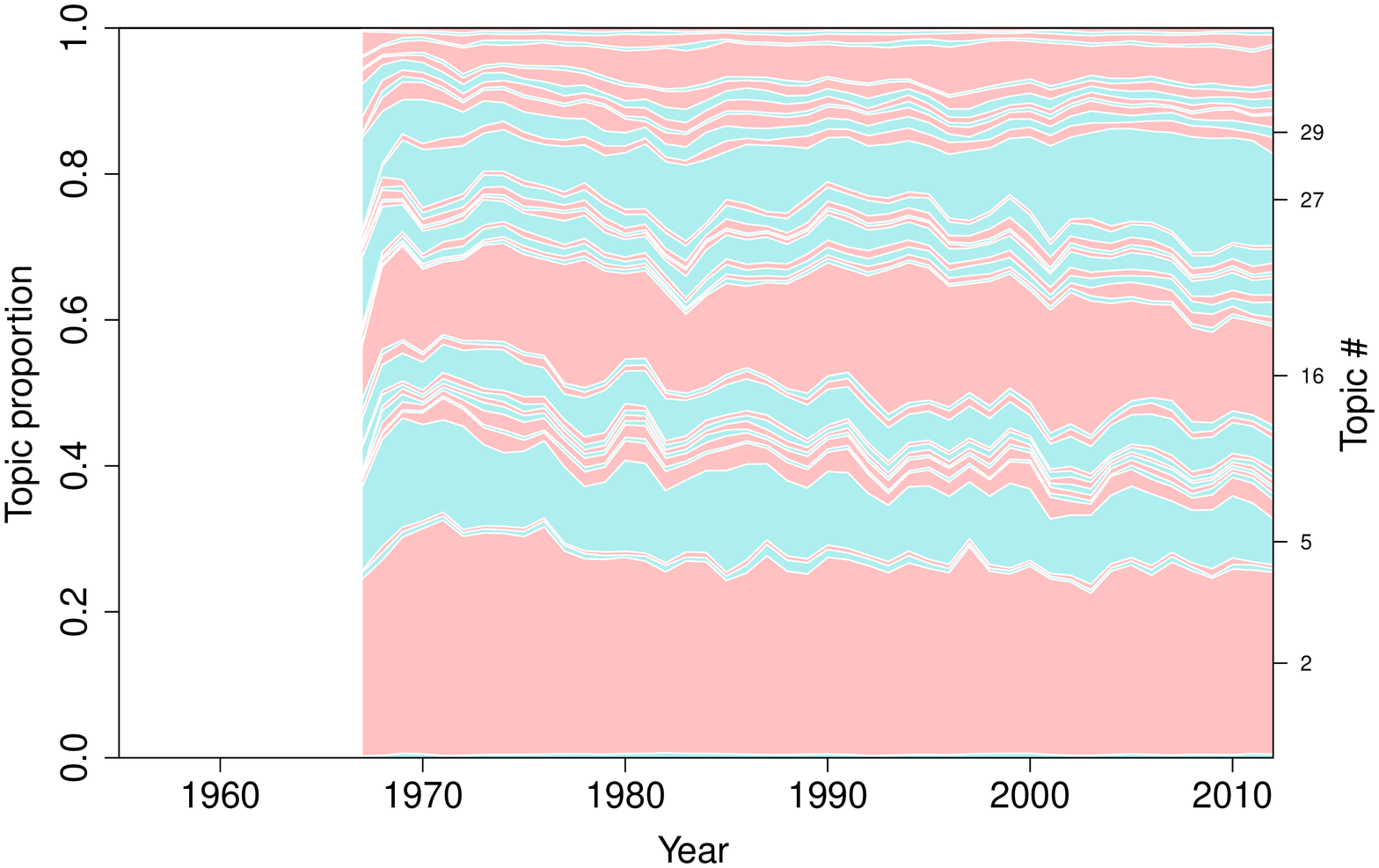} }
\begin{minipage}[b]{5in}
\caption{Journal topic distributions over time.}
\end{minipage}
\label{fig:jtopic_dist_overtime3}
\end{figure}

\begin{figure}[!ht]
\centering
\subfigure[Journal of Scheduling.]{\includegraphics[width=0.48\textwidth]{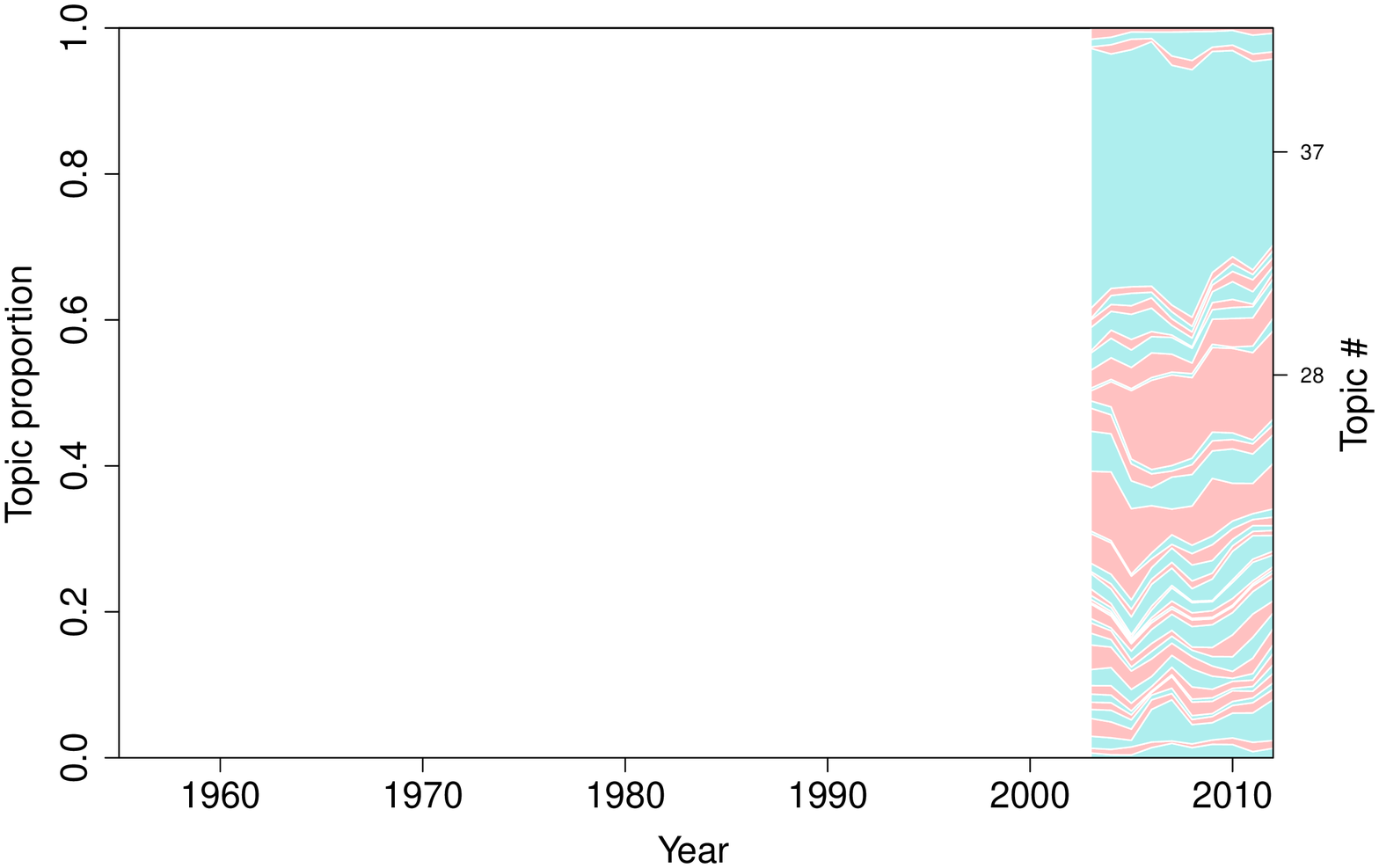} }
\subfigure[(INFORMS) Journal on Computing.]{\includegraphics[width=0.48\textwidth]{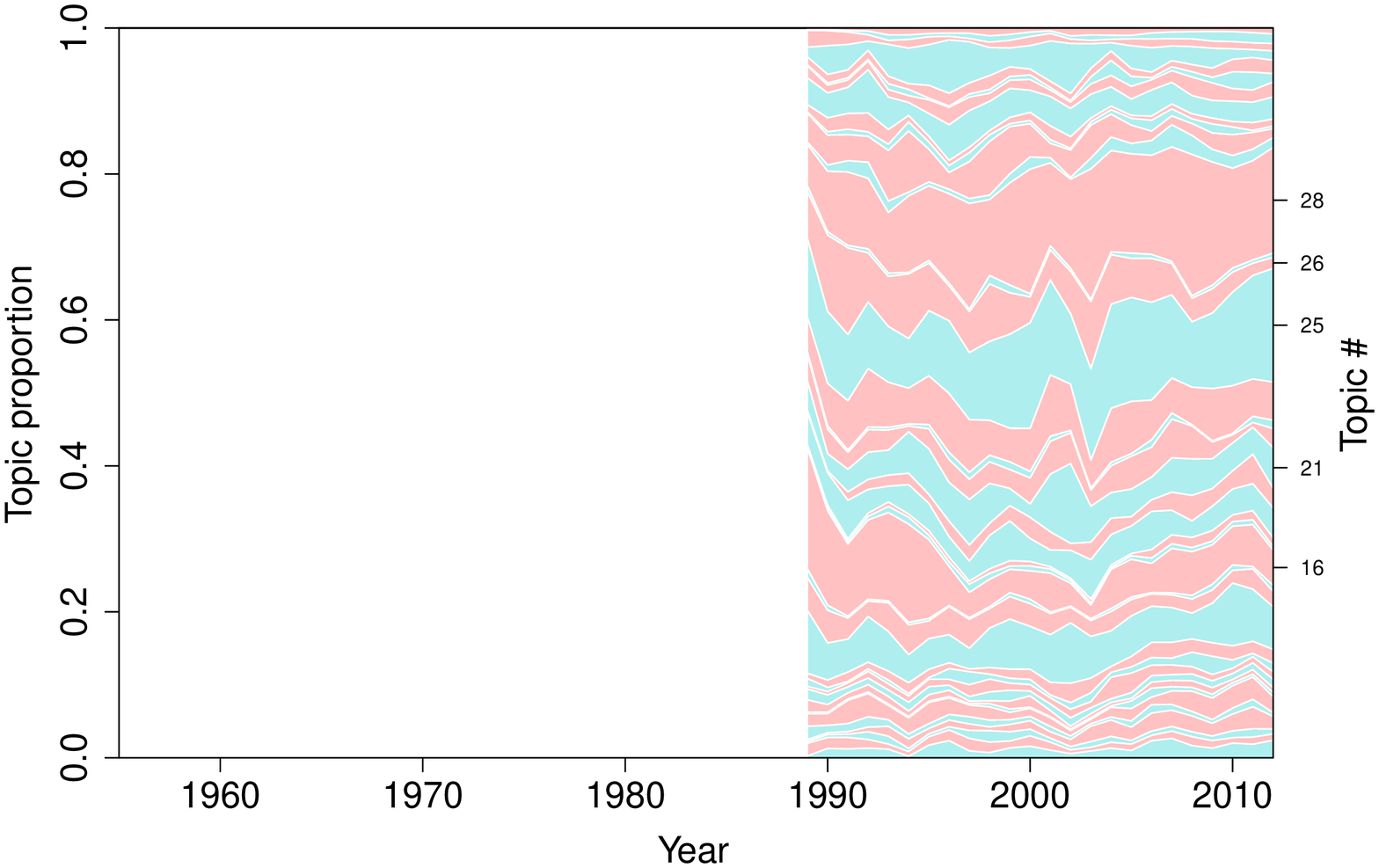} }
\subfigure[Manufacturing \& Service Operations Management.]{\includegraphics[width=0.48\textwidth]{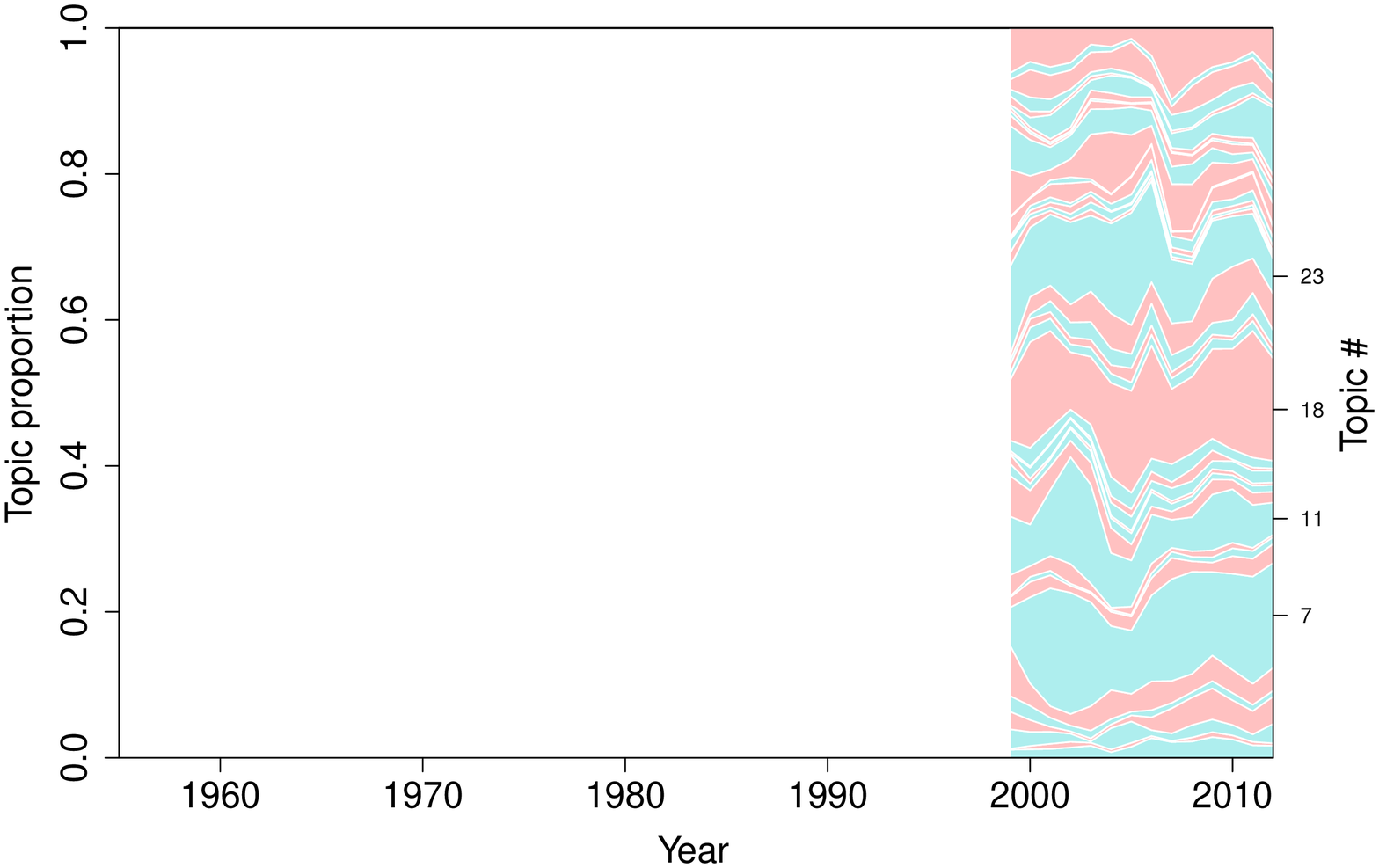} }
\subfigure[Mathematical Methods of Operations Research.]{\includegraphics[width=0.48\textwidth]{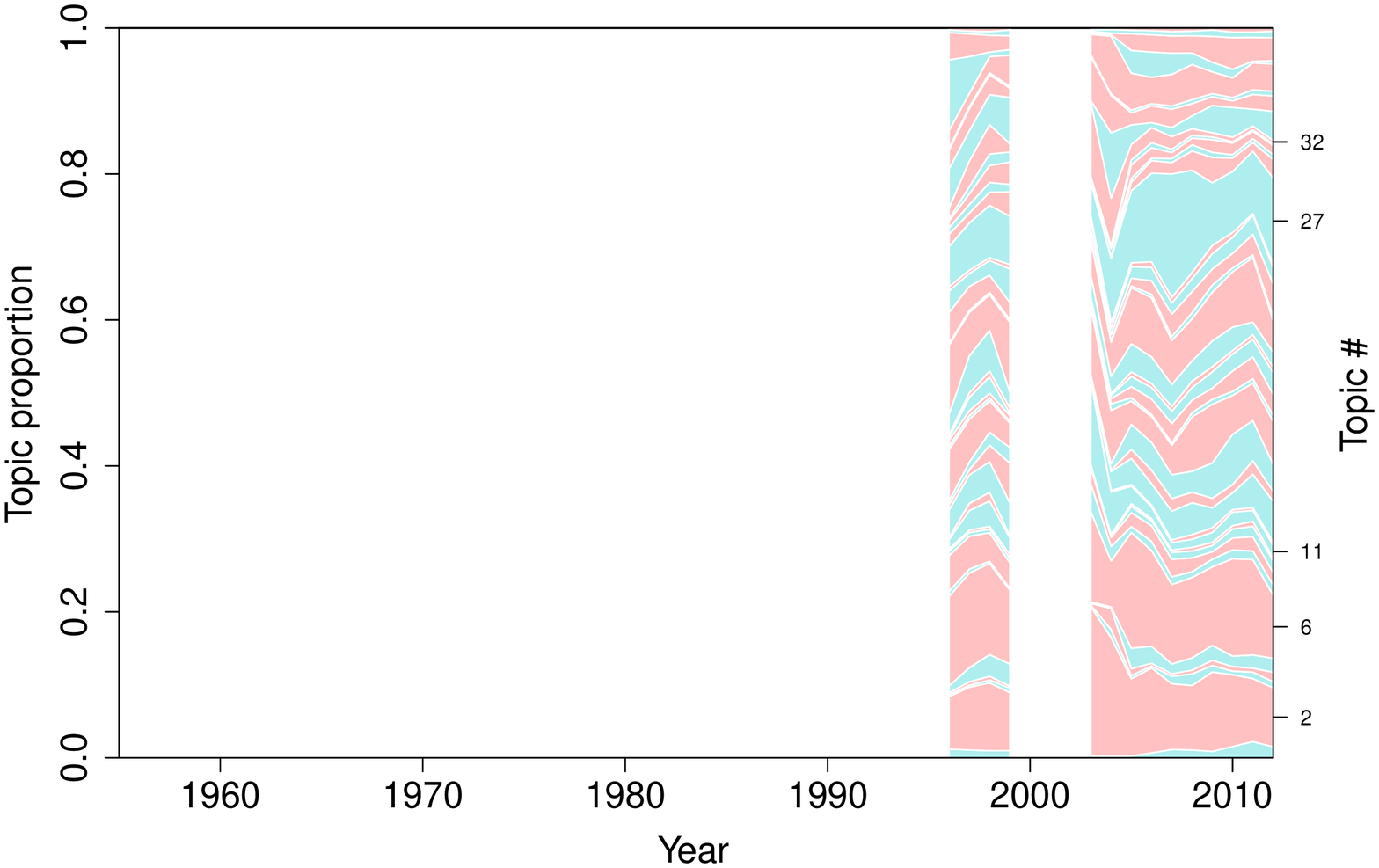} }
\subfigure[Mathematical Programming.]{\includegraphics[width=0.48\textwidth]{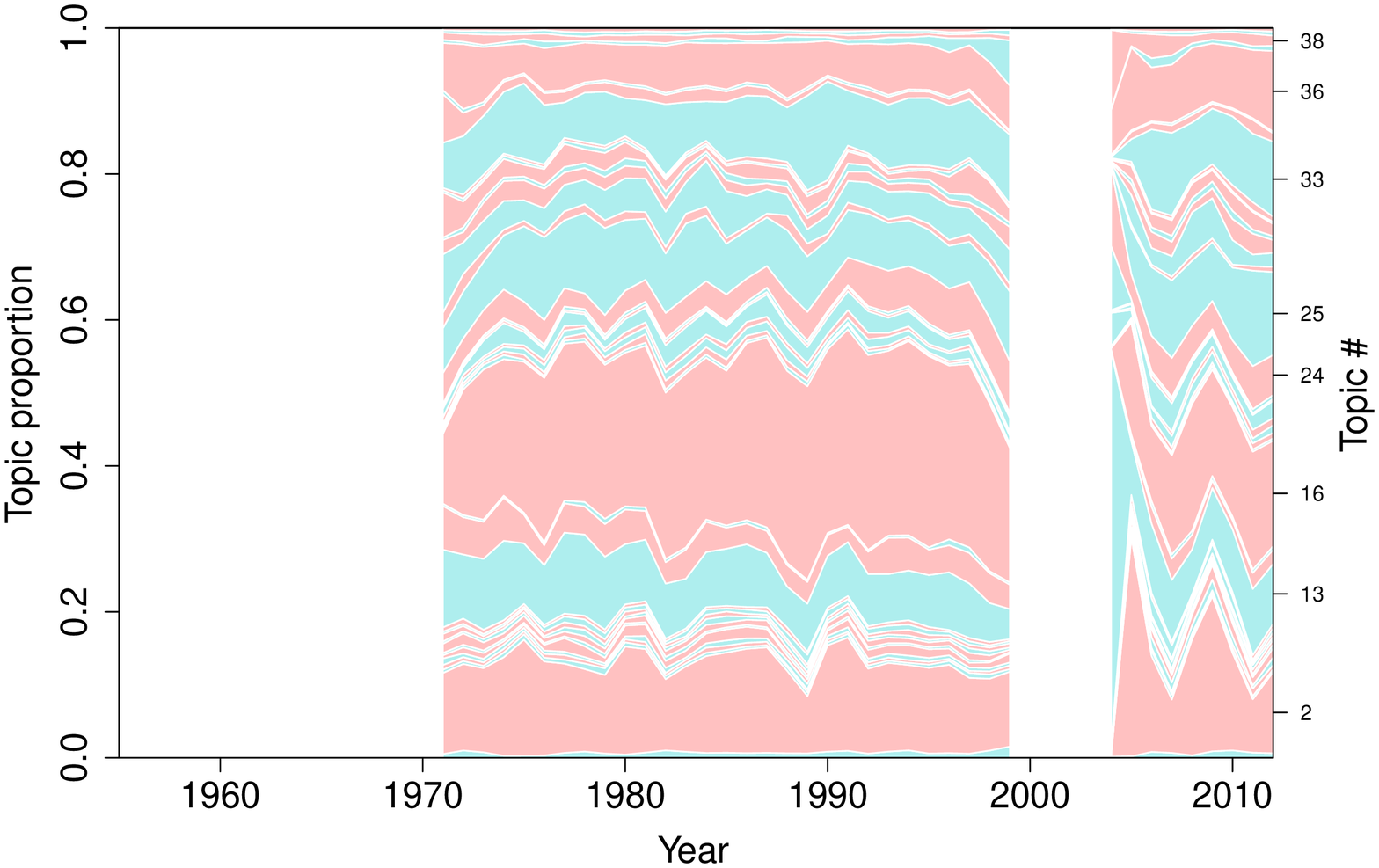} }
\subfigure[Mathematical Programming Computation.]{\includegraphics[width=0.48\textwidth]{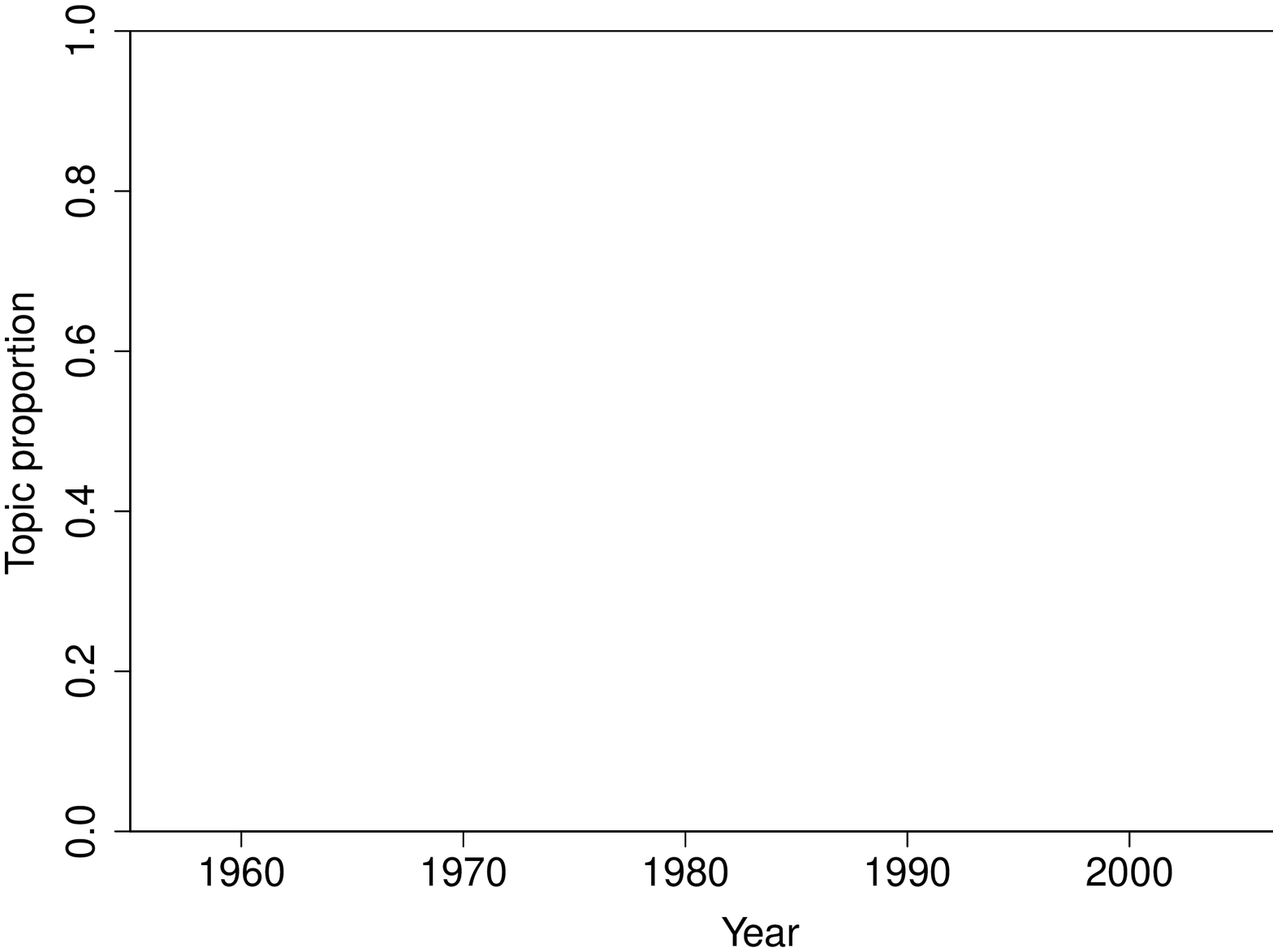} }
\begin{minipage}[b]{5in}
\caption{Journal topic distributions over time.}
\end{minipage}
\label{fig:jtopic_dist_overtime4}
\end{figure}

\begin{figure}[!ht]
\centering
\subfigure[Mathematics of Operations Research.]{\includegraphics[width=0.48\textwidth]{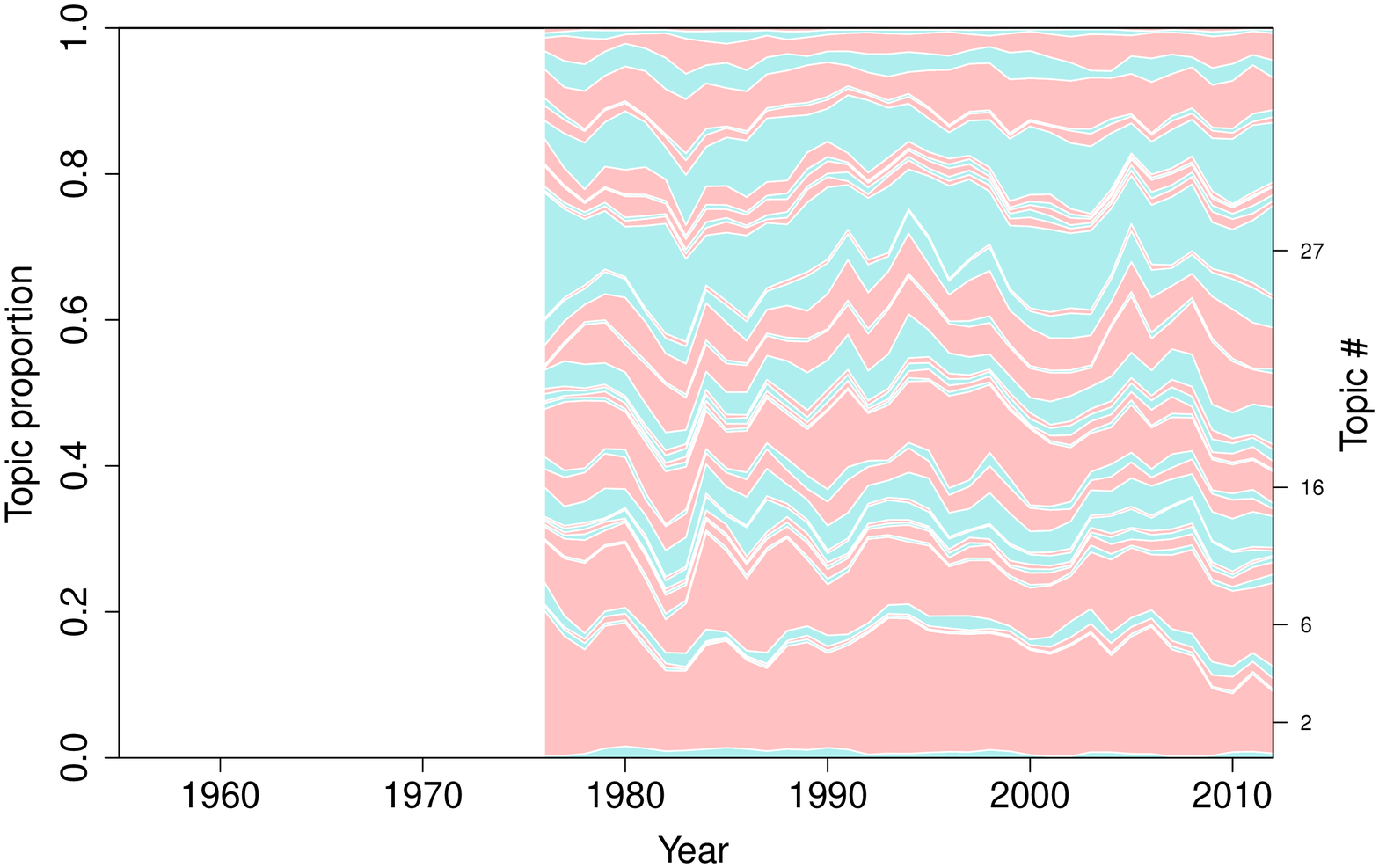} }
\subfigure[Military Operations Research.]{\includegraphics[width=0.48\textwidth]{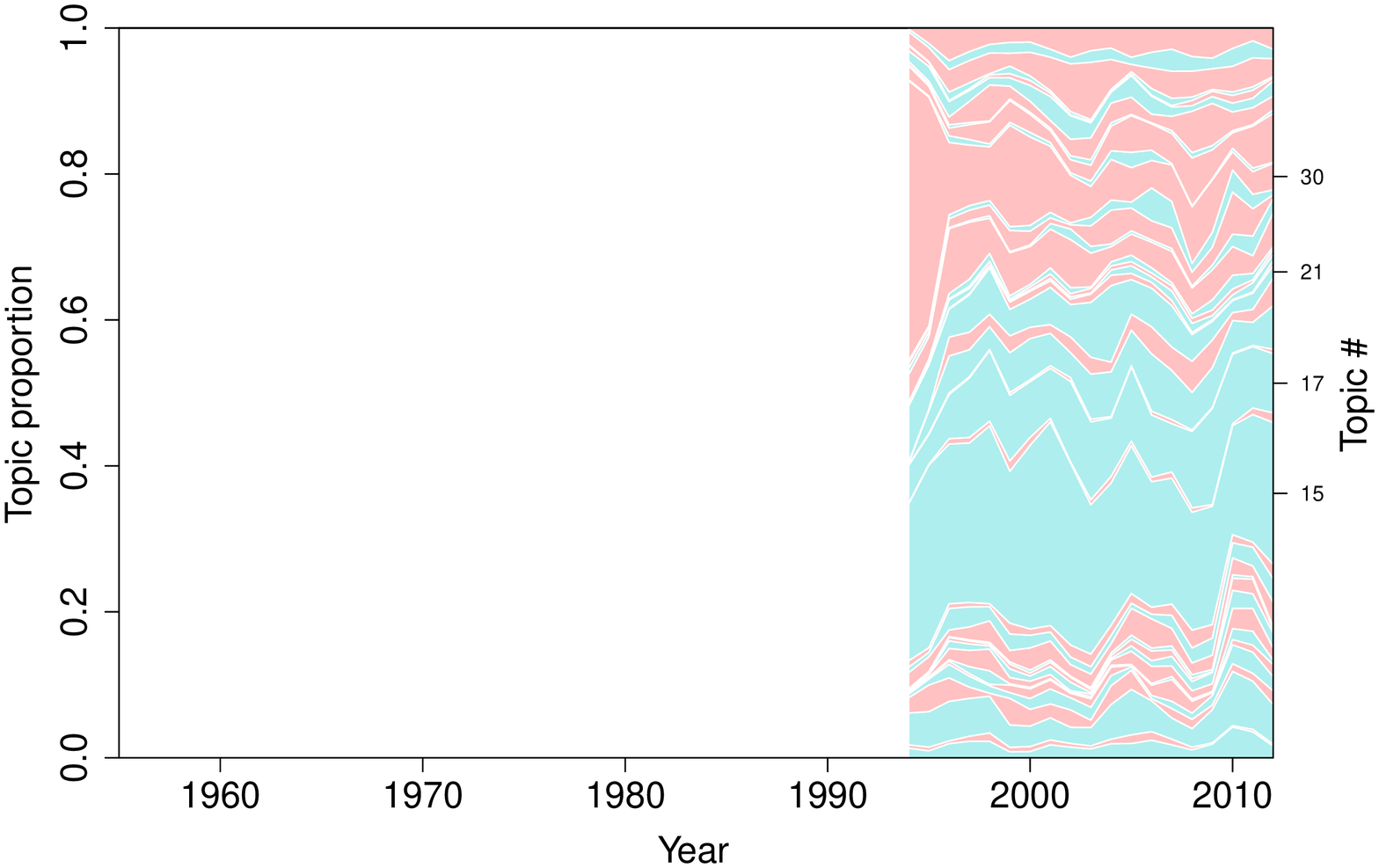} }
\subfigure[Networks.]{\includegraphics[width=0.48\textwidth]{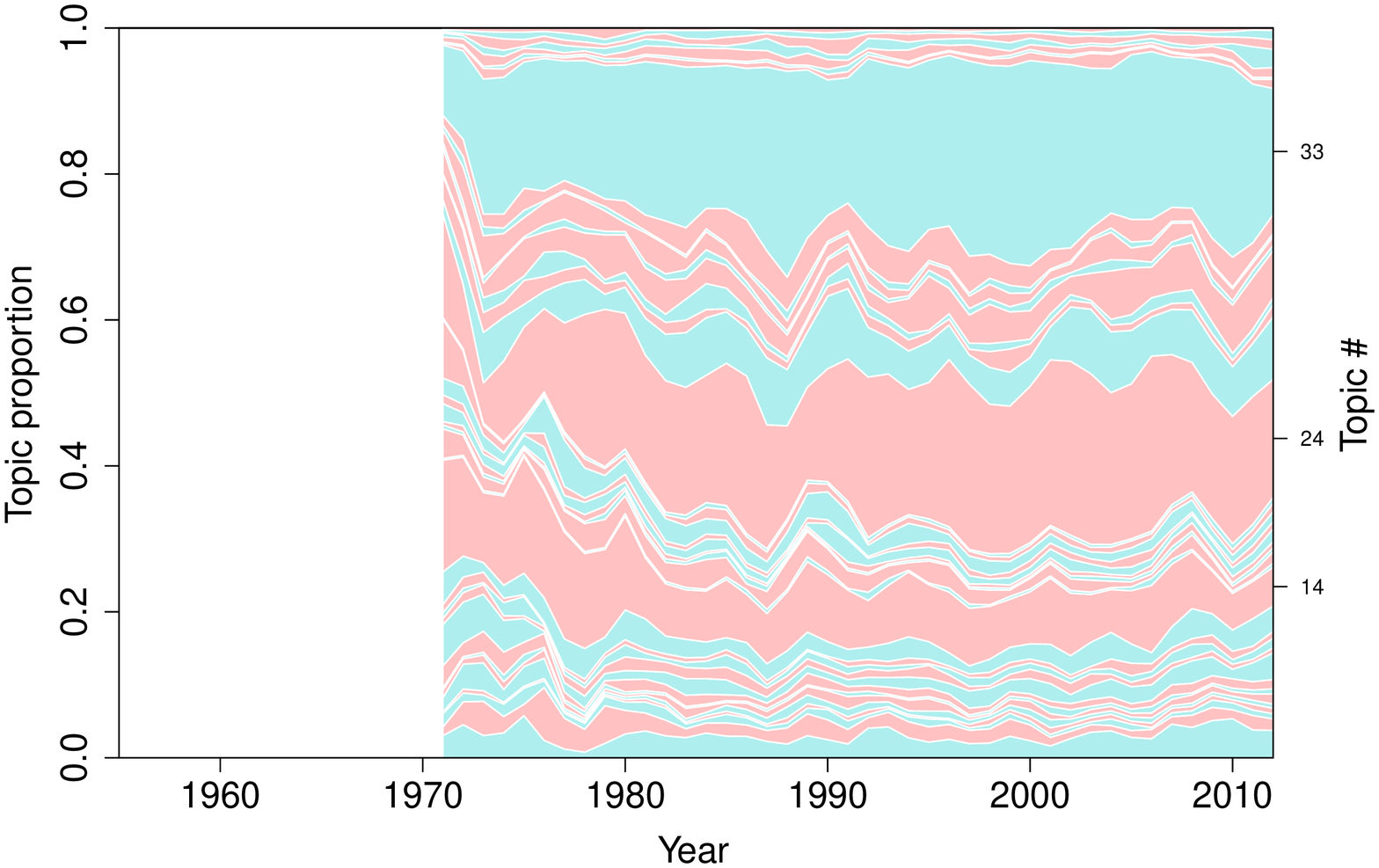} }
\subfigure[Omega.]{\includegraphics[width=0.48\textwidth]{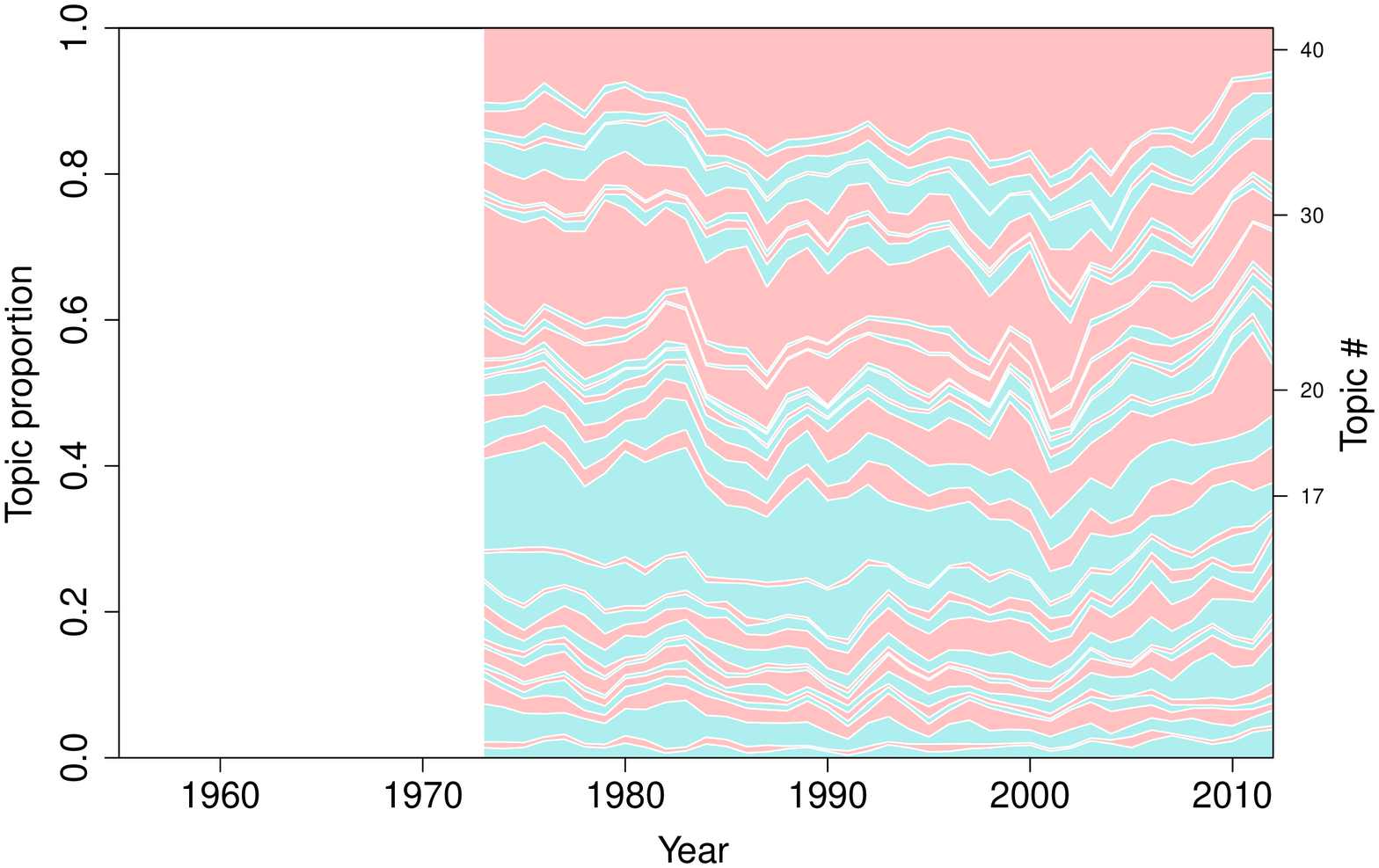} }
\subfigure[Operations Research.]{\includegraphics[width=0.48\textwidth]{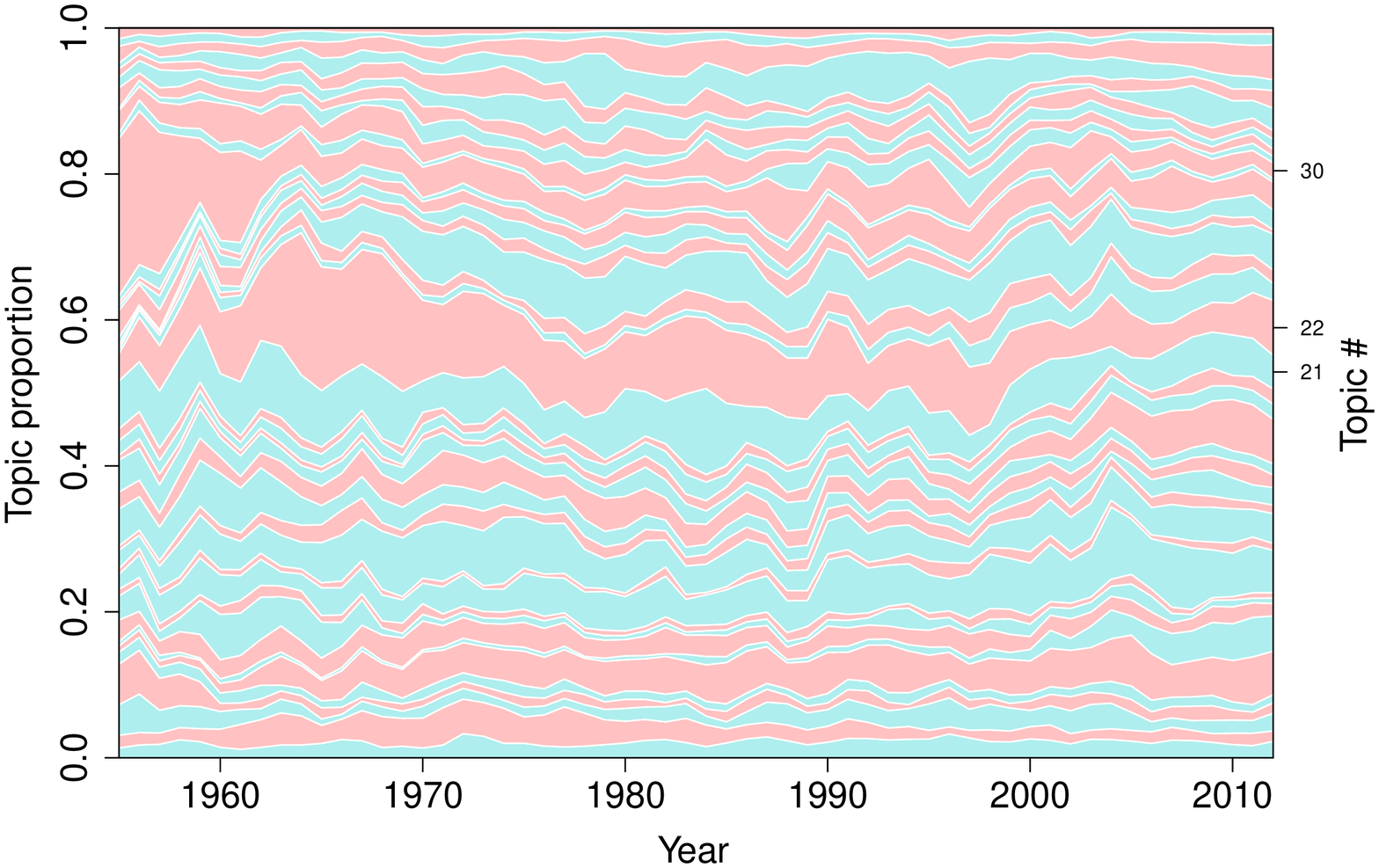} }
\subfigure[Operations Research Letters.]{\includegraphics[width=0.48\textwidth]{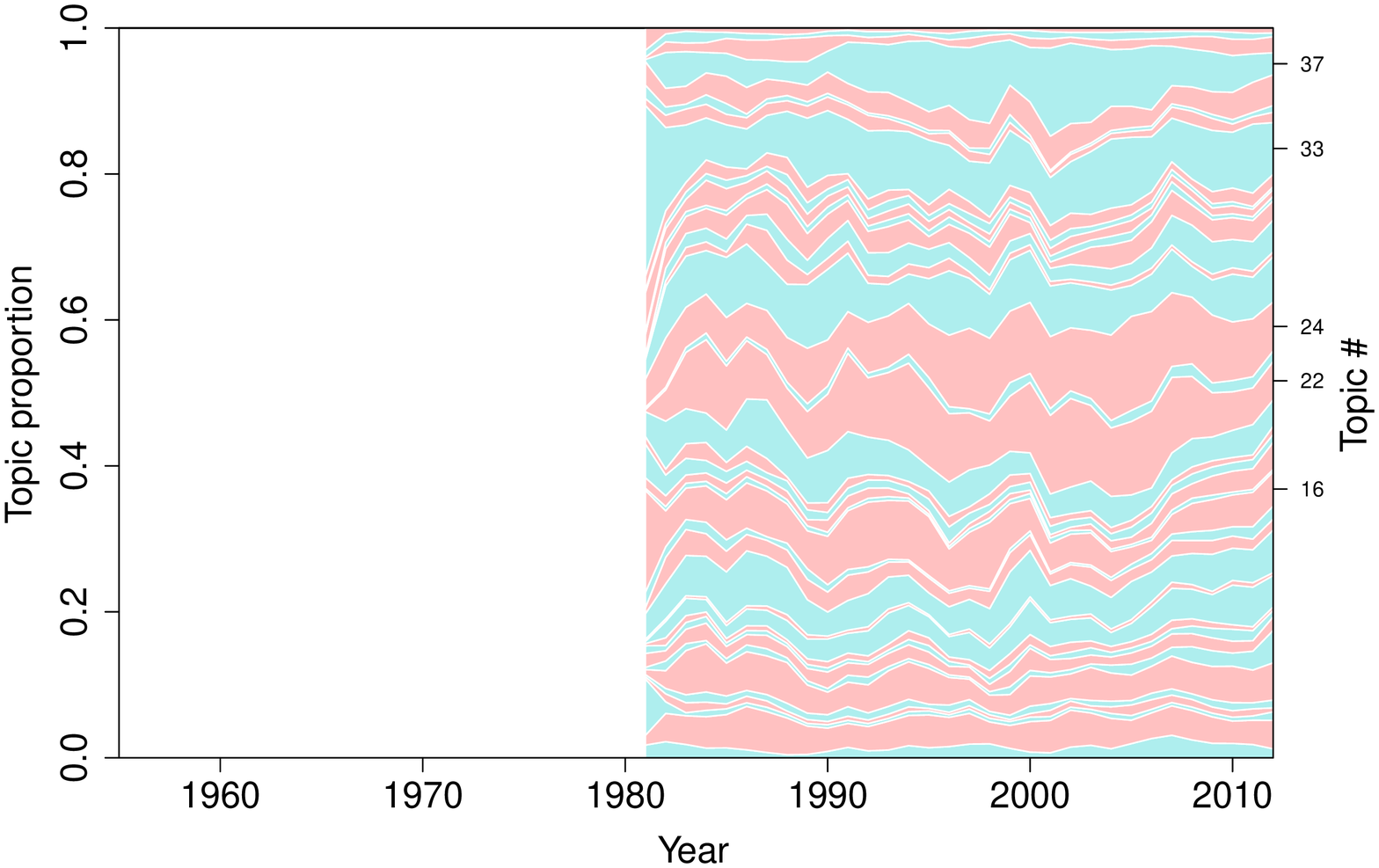} }
\begin{minipage}[b]{5in}
\caption{Journal topic distributions over time.}
\end{minipage}
\label{fig:jtopic_dist_overtime5}
\end{figure}

\begin{figure}[!ht]
\centering
\subfigure[Optimization Letters.]{\includegraphics[width=0.48\textwidth]{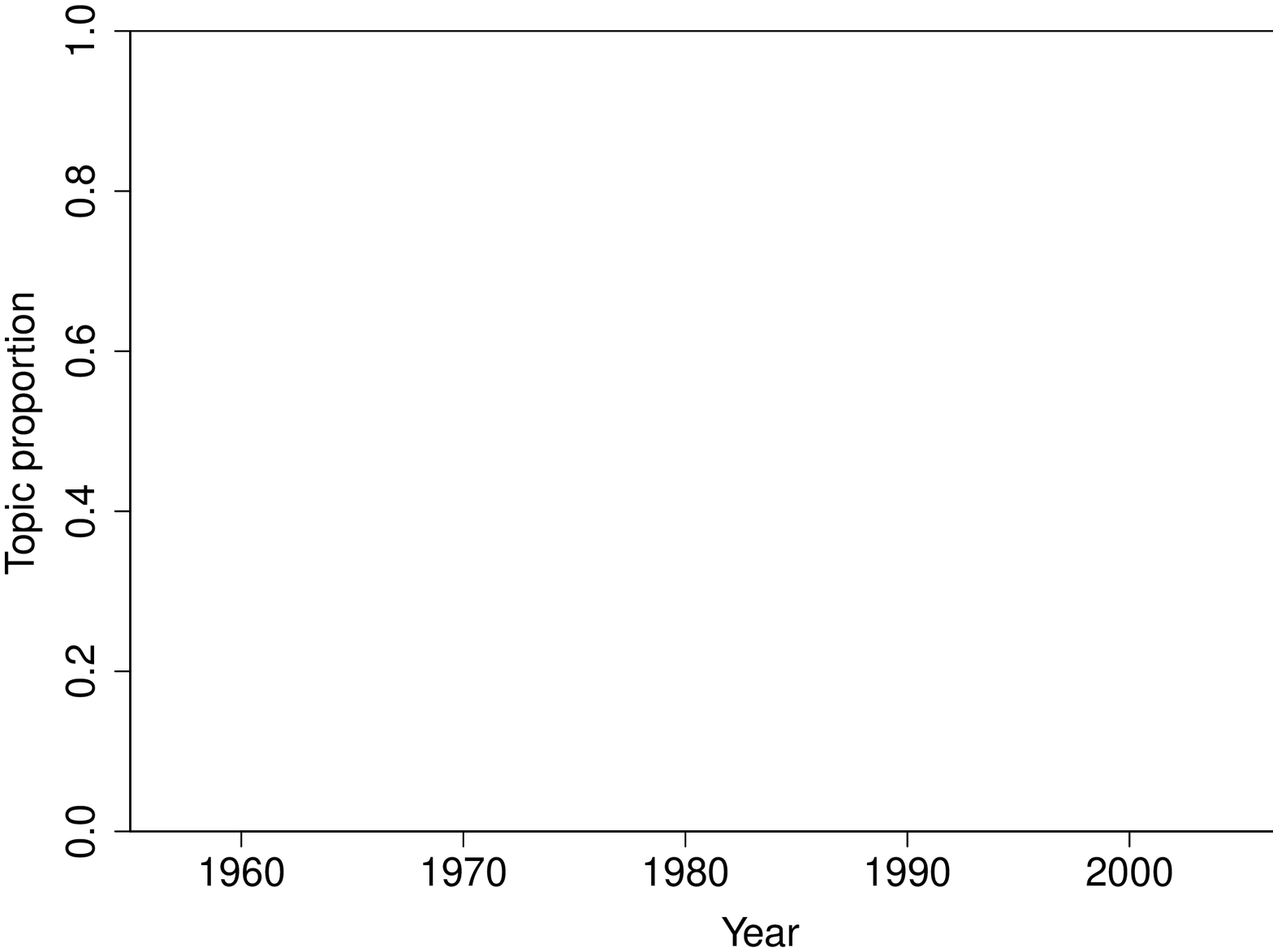} }
\subfigure[OR Spectrum.]{\includegraphics[width=0.48\textwidth]{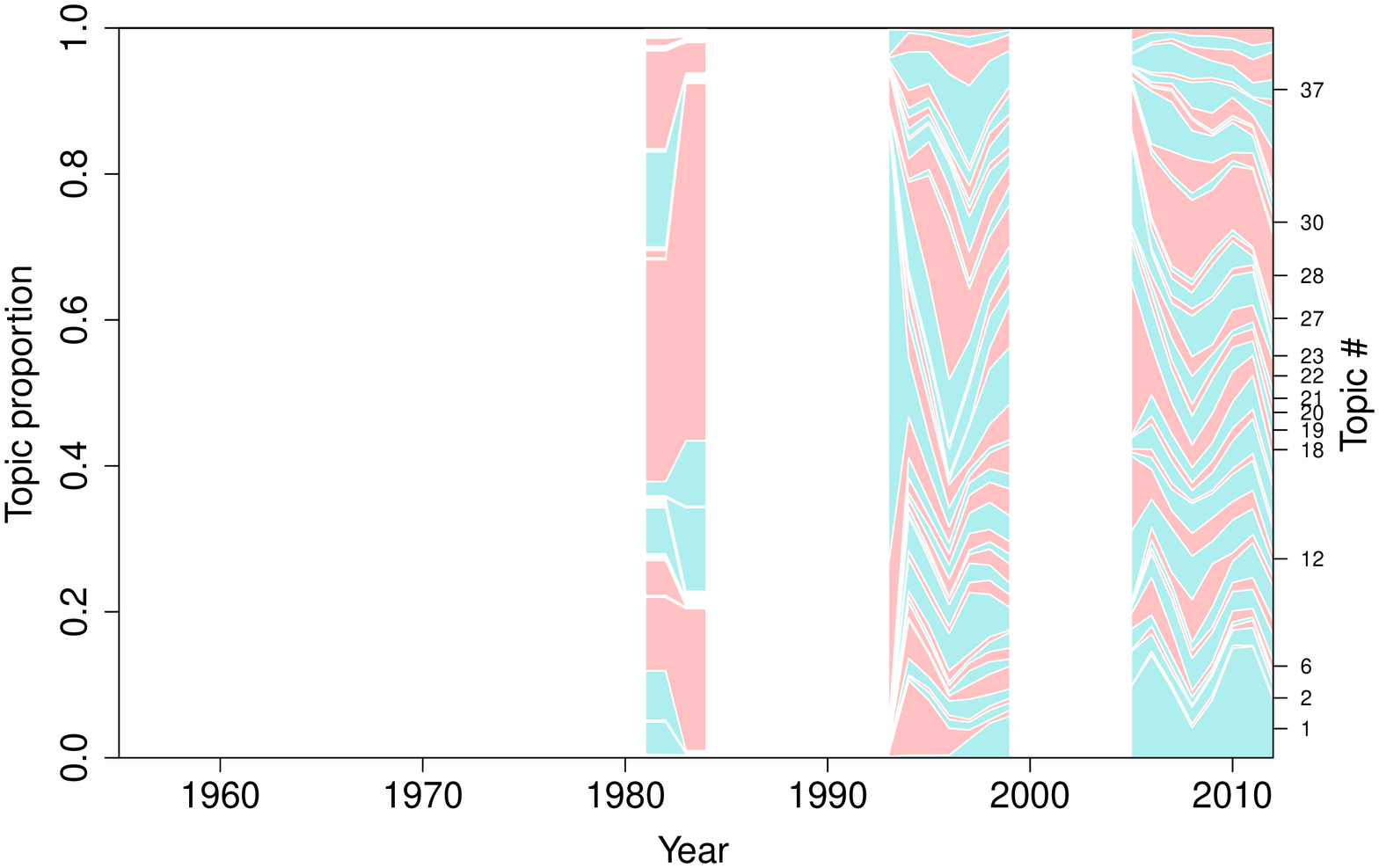} }
\subfigure[Production Planning \& Control.]{\includegraphics[width=0.48\textwidth]{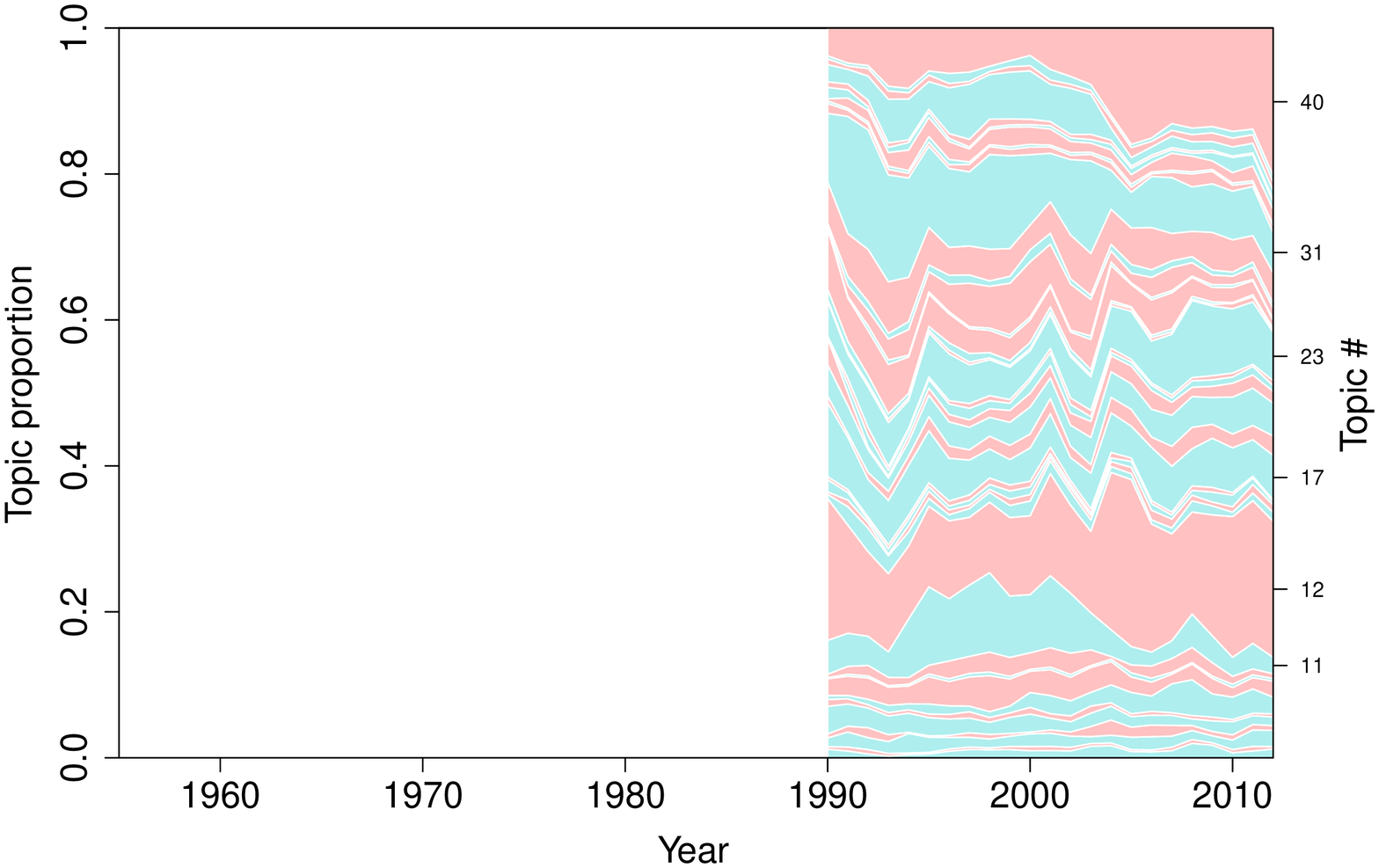} }
\subfigure[Queueing Systems.]{\includegraphics[width=0.48\textwidth]{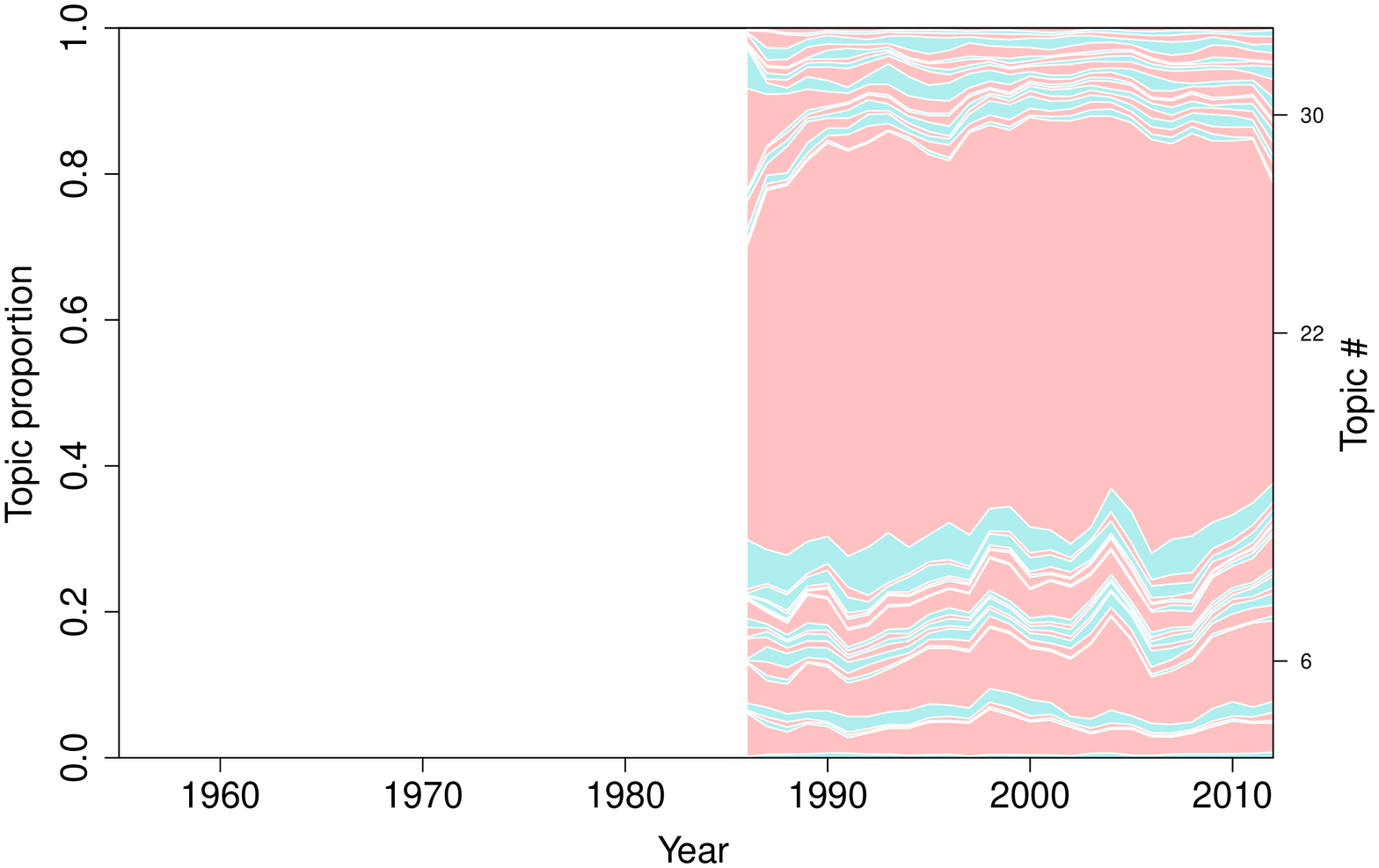} }
\subfigure[Surveys in Operations Research and Management Science.]{\includegraphics[width=0.48\textwidth]{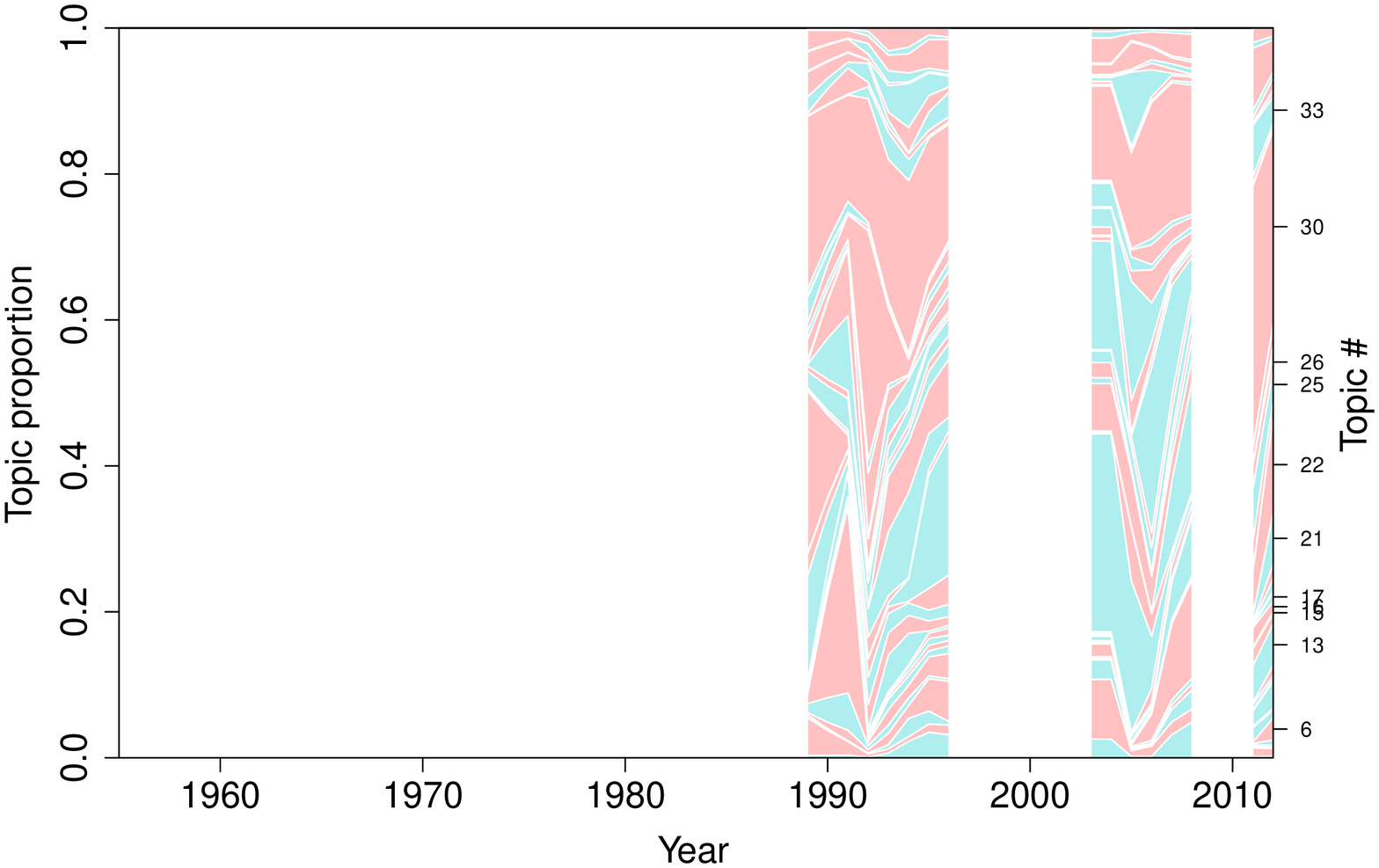} }
\subfigure[Transportation Research Part B.]{\includegraphics[width=0.48\textwidth]{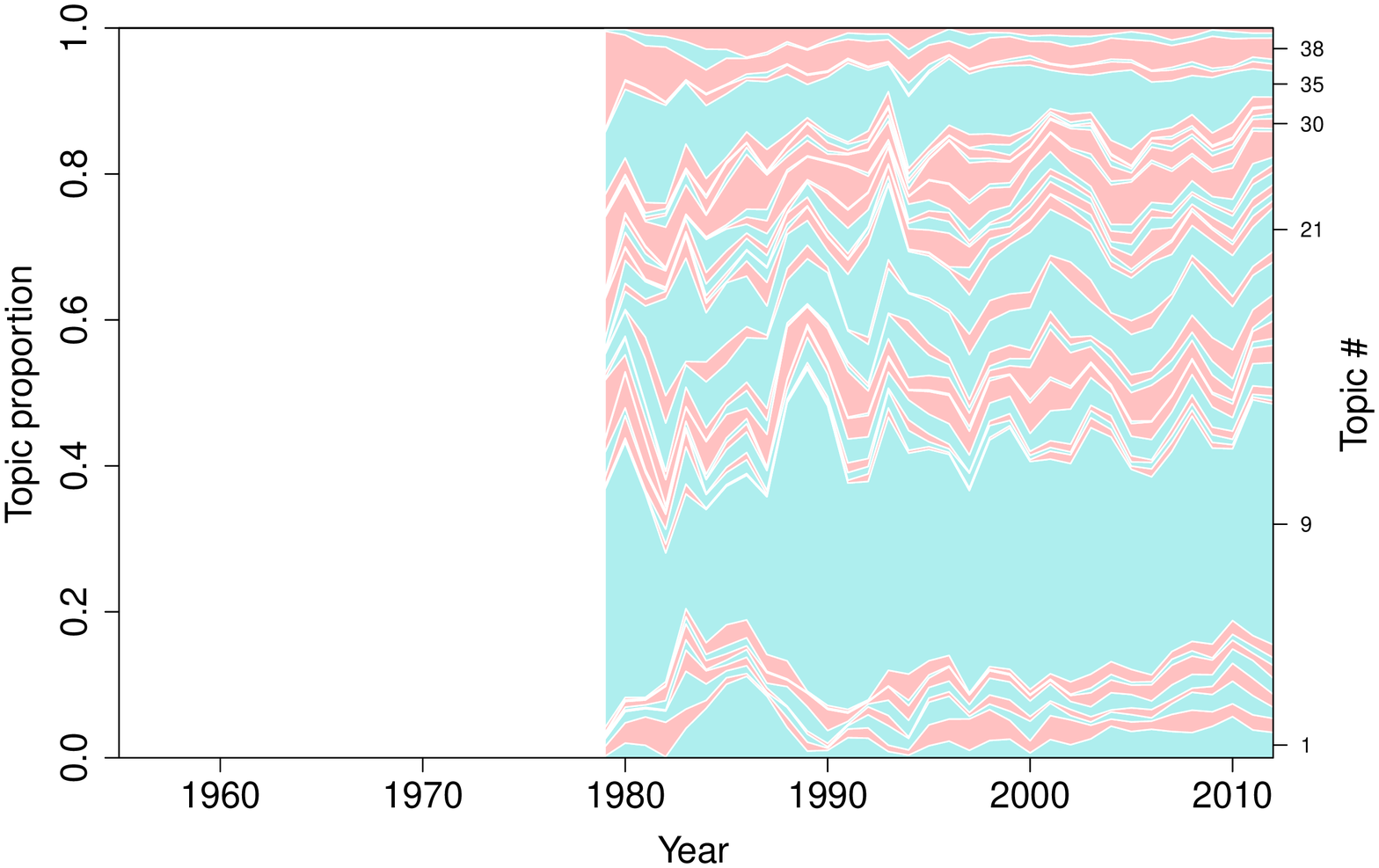} }
\begin{minipage}[b]{5in}
\caption{Journal topic distributions over time.}
\end{minipage}
\label{fig:jtopic_dist_overtime6}
\end{figure}

\begin{figure}[!ht]
\centering
\subfigure[Transportation Research Part E.]{\includegraphics[width=0.48\textwidth]{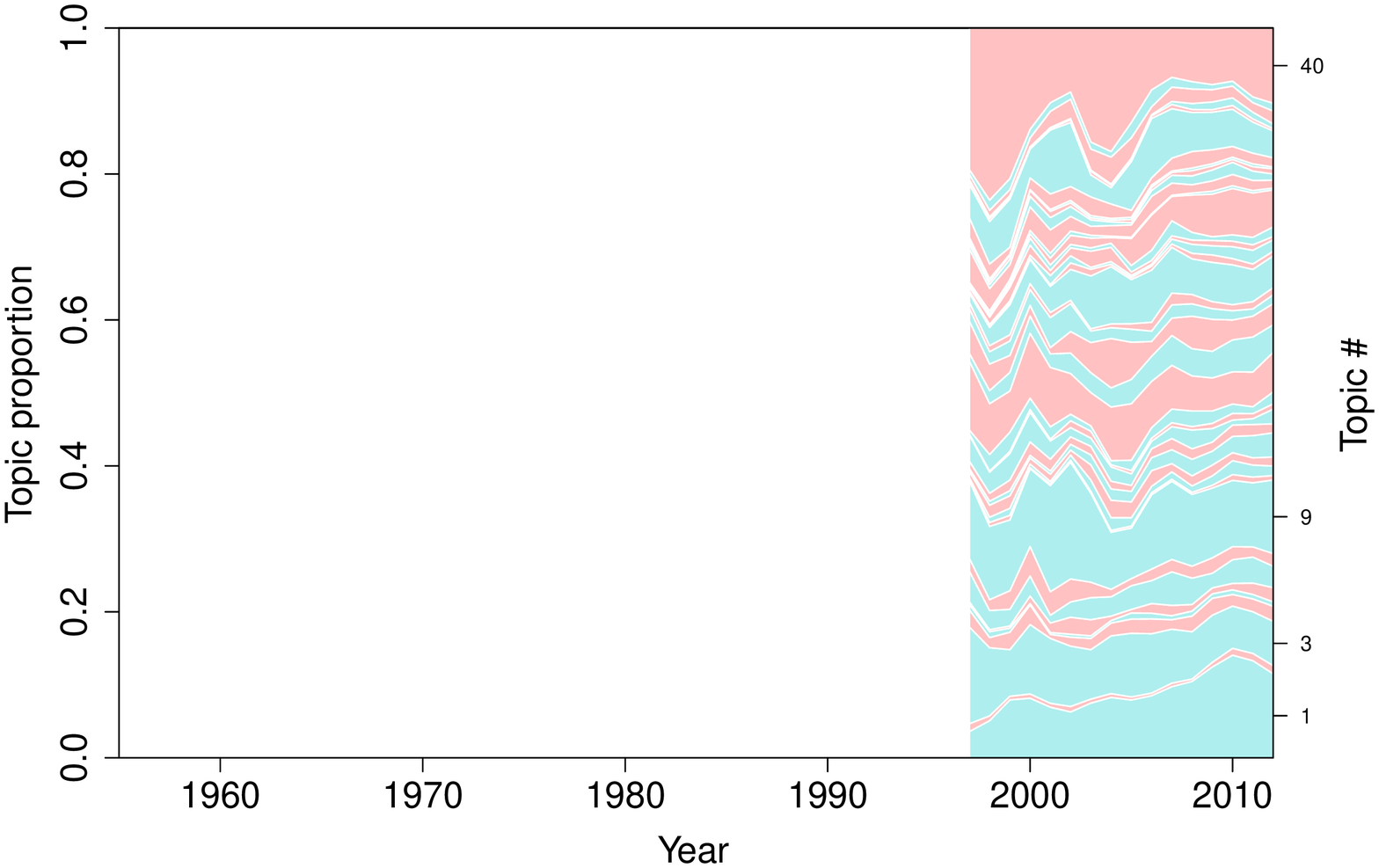} }
\begin{minipage}[b]{5in}
\caption{Journal topic distributions over time.}
\end{minipage}
\label{fig:jtopic_dist_overtime7}
\end{figure}

\end{document}